\documentclass{article}
\usepackage{iclr2027_conference,times}

\usepackage{amsmath,amsfonts,bm}

\def\eqref#1{equation~\ref{#1}}

\def\1{\bm{1}}

\DeclareMathAlphabet{\mathsfit}{\encodingdefault}{\sfdefault}{m}{sl}
\SetMathAlphabet{\mathsfit}{bold}{\encodingdefault}{\sfdefault}{bx}{n}

\usepackage[dvipsnames]{xcolor}
\usepackage{algorithm}
\usepackage{algpseudocode}
\usepackage{multirow}
\usepackage{booktabs}
\usepackage{graphicx}
\usepackage{float}
\usepackage{adjustbox}
\usepackage{comment}
\usepackage{listings}
\usepackage{tcolorbox}
\tcbuselibrary{listings,breakable,skins}
\usepackage{hyperref}
\usepackage{url}
\usepackage{xurl}

\newcommand{\cocoevolve}{CoCoEvolve}

\newtcblisting{promptbox}[1][]{%
  enhanced,
  listing only,
  breakable,
  colback=black!6,
  colframe=black!40,
  boxrule=0.4pt,
  leftrule=1.4pt,
  arc=0pt,
  left=5pt,
  right=5pt,
  top=3pt,
  bottom=3pt,
  before skip=5pt,
  after skip=7pt,
  listing options={%
    basicstyle=\footnotesize\ttfamily,
    breaklines=true,
    columns=fullflexible,
    keepspaces=true,
    showstringspaces=false,
    literate=
      {—}{{\textemdash}}1
      {–}{{\textendash}}1
      {→}{{$\rightarrow$}}1
      {≤}{{$\leq$}}1
      {─}{{-}}1
      {═}{{=}}1
      {…}{{...}}3,
    moredelim={**[is][\color{blue}]{@@}{@@}},
  },
  IfBlankTF={#1}{}{%
    title={#1},
    coltitle=black,
    fonttitle=\small\itshape,
    colbacktitle=black!12,
    toptitle=0.5mm,
    bottomtitle=0.5mm,
    attach boxed title to top left={yshift=-0.1mm},
    boxed title style={
      colframe=black!40,
      colback=black!12,
      boxrule=0.4pt,
      arc=0pt,
      left=3pt,
      right=3pt,
      top=1pt,
      bottom=1pt,
    },
  },
}

\newcommand{\programlisting}[2]{%
  \tcbinputlisting{
    enhanced,
    listing engine=listings,
    listing only,
    listing file={#2},
    breakable,
    colback=black!6,
    colframe=black!40,
    boxrule=0.4pt,
    leftrule=1.4pt,
    arc=0pt,
    left=5pt,
    right=5pt,
    top=3pt,
    bottom=3pt,
    before skip=5pt,
    after skip=7pt,
    listing options={
      language=Python,
      basicstyle=\scriptsize\ttfamily,
      breaklines=true,
      breakatwhitespace=true,
      columns=fullflexible,
      keepspaces=true,
      showstringspaces=false,
    },
    title={#1},
    coltitle=black,
    fonttitle=\small\itshape,
    colbacktitle=black!12,
    toptitle=0.5mm,
    bottomtitle=0.5mm,
  }%
}
\newcommand{\championlisting}[2]{%
  \tcbinputlisting{
    enhanced,
    listing engine=listings,
    listing only,
    listing file={#2},
    breakable,
    colback=black!6,
    colframe=black!40,
    boxrule=0.4pt,
    leftrule=1.4pt,
    arc=0pt,
    left=11pt,
    right=3pt,
    top=1pt,
    bottom=1pt,
    before skip=5pt,
    after skip=7pt,
    listing options={
      language=Python,
      basicstyle=\ttfamily\fontsize{6}{7}\selectfont,
      numbers=left,
      numberstyle=\ttfamily\fontsize{5}{6}\selectfont\color{black!45},
      numbersep=3pt,
      xleftmargin=1.7em,
      literate={~}{{\textasciitilde}}1,
      breaklines=true,
      breakatwhitespace=true,
      columns=fullflexible,
      keepspaces=true,
      showstringspaces=false,
    },
    title={#1},
    coltitle=black,
    fonttitle=\small\itshape,
    colbacktitle=black!12,
    toptitle=0.5mm,
    bottomtitle=0.5mm,
  }%
}

\newcommand{\ProgSpace}{\mathcal{P}}
\newcommand{\prog}{p}
\newcommand{\Task}{\mathcal{T}}
\newcommand{\TaskDist}{\mathcal{D}}
\newcommand{\task}{\tau}
\newcommand{\ObsSpace}{\mathcal{O}}
\newcommand{\obs}{o}
\newcommand{\Obs}{O}
\newcommand{\score}{s}
\newcommand{\Score}{S}
\newcommand{\Fitness}{\Phi}
\newcommand{\Budget}{T}
\newcommand{\Archive}{\mathcal{A}}
\newcommand{\History}{\mathcal{H}}
\newcommand{\CtxSpace}{\mathcal{C}}
\newcommand{\ctxbundle}{c}
\newcommand{\Curate}{\mathrm{Curate}}
\newcommand{\Augment}{\mathrm{Augment}}
\newcommand{\MetaReflect}{\mathrm{MetaReflect}}
\newcommand{\Env}{\mathcal{E}}

\newcommand{\TaskPool}{V}
\newcommand{\Sim}{\kappa}
\newcommand{\ExpScore}{\widehat{\score}}
\newcommand{\InfoGain}{\mathrm{gain}}
\newcommand{\lp}{\ell^p}

\title{\texorpdfstring{\raggedright}{}{} Meta-Reflective Evolution Effectively Improves Production AI Systems}

\author{
Seung Won Wilson Yoo\textsuperscript{1}, Chang Zeng\textsuperscript{2,}\thanks{Work done during internship at Snowflake.}, Aditi Somayajula\textsuperscript{3,$\ast$}, Sahithi Ankireddy\textsuperscript{3,$\ast$} \\
\textbf{Rohan Kosalge\textsuperscript{4,$\ast$}, Avishree Khare\textsuperscript{5,$\ast$}, John Mitchell\textsuperscript{1,3}, Aditya G. Parameswaran\textsuperscript{1,4}} \\
\textbf{Rajhans Samdani\textsuperscript{1}, Anupam Datta\textsuperscript{1}} \\[2pt]
\normalfont\small
\textsuperscript{1}Snowflake \quad
\textsuperscript{2}UMass Amherst \quad
\textsuperscript{3}Stanford \quad
\textsuperscript{4}UC Berkeley \quad
\textsuperscript{5}UPenn
}

\iclrfinalcopy

\begin{document}
\maketitle
\lhead{}

\begin{abstract}
Large Language Model (LLM)-guided evolutionary optimization has successfully improved prompts, programs, and even entire AI systems.
Most optimization methods select a candidate from the search frontier, then mutate it in one LLM pass based only on its failures.
This focus on the frontier can stall on production AI systems, where isolated failures reveal little about complex, hidden interactions.
We introduce \cocoevolve{} (\textbf{Co}rtex \textbf{Co}de \textbf{Evolve}), an evolutionary optimizer built around what we call \emph{meta-reflection}, wherein an agentic mutator (Cortex Code) selectively retrieves evidence from the entire evolutionary history and probes the live system before verifying and proposing mutations.
We apply \cocoevolve{} to optimize three classes of production AI systems: data agents, LLM-powered data pipelines, and data engineering workflows.
On DABStep, \cocoevolve{} tops the public leaderboard, reaching 92.8\% held-out hard accuracy compared to 78.4\% for Meta-Harness and 30.4\% for GEPA under matched conditions.
Across six LLM data pipeline workloads, it outperforms every evaluated baseline.
On data-eng-bench, \cocoevolve{} lifts test-set solve rate from 59.1\% to 77.4\%, past the best baseline of 73.6\%.
It also has 11pp higher accuracy than GEPA on HotpotQA while remaining competitive with specialized optimizers on established mathematical and systems benchmarks.
Matched ablations confirm that each component of \cocoevolve{} contributes to these gains, demonstrating the value of meta-reflection for optimizing production AI systems.

\end{abstract}

\section{Introduction}
\label{sec:introduction}
Large language models (LLMs) are increasingly used as semantic mutation operators: given a candidate prompt or program and evaluation feedback for that candidate, an LLM proposes a revised candidate.
This approach, exemplified by evolutionary optimizers like AlphaEvolve and OpenEvolve, has led to breakthrough improvements in prompts, algorithms, and other text-based artifacts because LLMs can make semantically meaningful changes that conventional random mutation cannot \citep{novikov2025alphaevolve,sharma2025openevolve}.
However, most LLM-guided evolutionary optimizers select a single parent, use its failures to propose a new candidate in one LLM call, and only then evaluate the result.

This parent-local, one-shot interface is effective when the failure makes both the defect and its repair apparent.
It falls short for production AI systems, where isolated failures can reveal little about interacting causes and relevant evidence may be distributed across the search history.
First, the selected parent's failure may not reveal the root cause, and a locally beneficial edit may regress capabilities elsewhere in the search history.
Second, the mutator cannot interact with the target environment, test hypotheses, or revise its candidate in response to intermediate results.
Search can consequently trade one solved task for another, revisit exhausted strategies, or end up plateauing \citep{zhou2026aichilles}.

We introduce \cocoevolve{} (\textbf{Co}rtex \textbf{Co}de \textbf{Evolve}), an evolutionary optimizer built around \textit{meta-reflection}: (i) selectively retrieving evidence from the accumulated search history beyond the selected parent's failure and (ii) actively probing the target environment while constructing and validating a mutation.
To support meta-reflection, \cocoevolve{} uses Cortex Code as its coding-agent mutator \citep{snowflake2026cortexcode}.
The mutator can inspect the parent, run focused experiments, revise its implementation, and validate the result before committing a child.
It accesses evolutionary history through both a compact summary and structured queries, allowing it to retrieve prior programs that solved a specific task, inspect the mutation that restored a lost capability, or identify related tasks that have historically regressed together.

\begin{figure}[t]
    \centering
    \includegraphics[width=\linewidth]{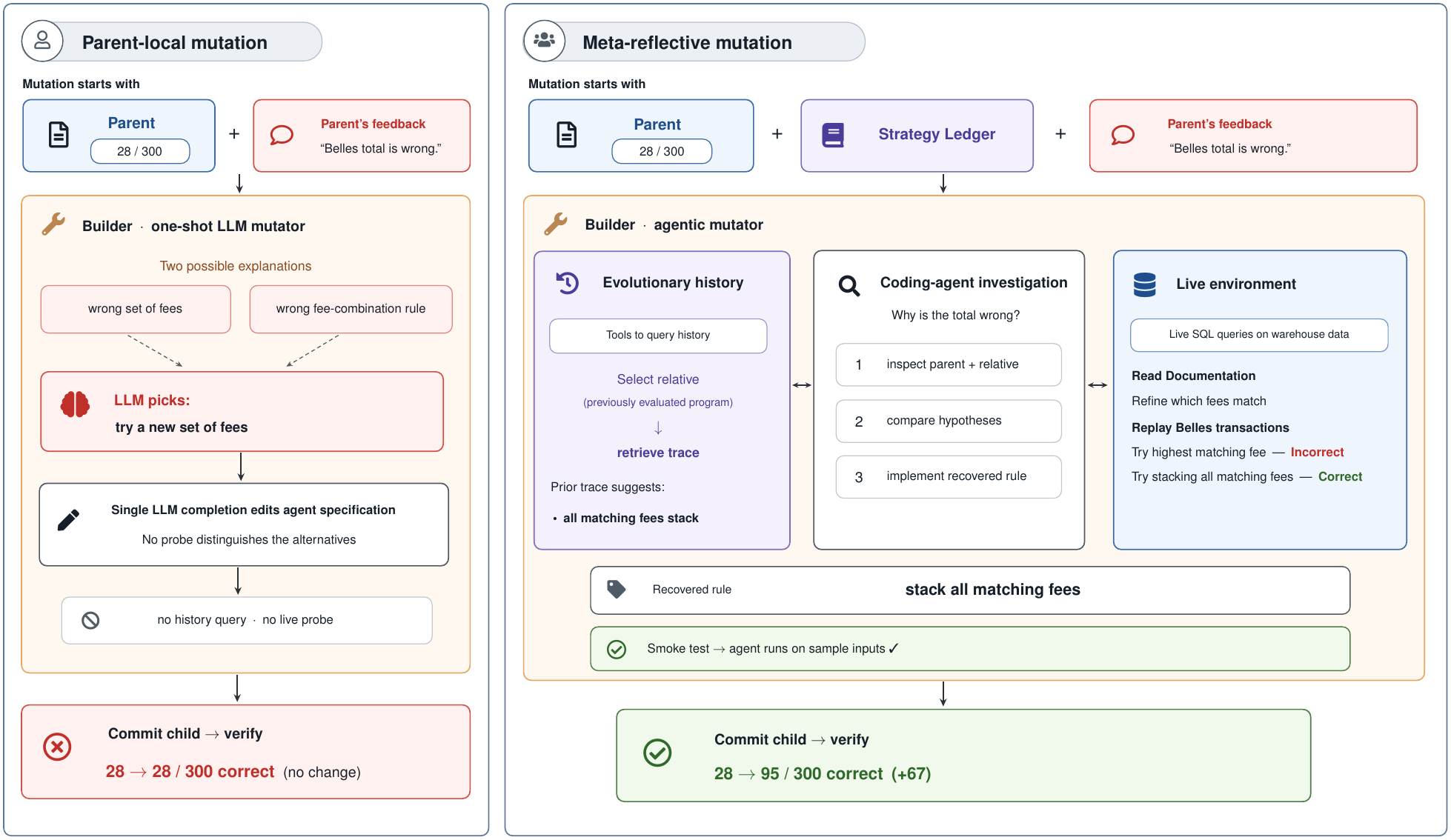}
    \caption{\textbf{Left:} given under-determined feedback, the parent-local mutator guesses a repair in one call and makes no progress. \textbf{Right:} a meta-reflective mutator queries the evolutionary history and reads the trace of a previously evaluated program to determine that \emph{all matching fees stack}. It also reads documentation to refine which fees match, and synthesizes the two observations. It tests the resulting hypothesis before commit and raises training accuracy from $28/300$ to $95/300$.
    }
    \label{fig:meta-reflection-example}
\end{figure}

Figure~\ref{fig:meta-reflection-example} illustrates meta-reflection for a DABStep data agent \citep{egg2025dabstep}, where payment-processing rules must be inferred across documentation and live data. Here, a one-shot, parent-local mutation doesn't make progress, whereas \cocoevolve{}'s meta-reflective mutation retrieves the missing evidence, validates the corresponding repair, and succeeds.
The same behavior repeats with \emph{Lockpicking with Interference}, a synthetic stress test where every leading baseline plateaus below perfect accuracy within 50 iterations, while \cocoevolve{} reaches 100\% in 6 iterations (Appendix~\ref{sec:app-lockpicking}).

We evaluate \cocoevolve{} primarily on production AI systems, where correctness depends on interactions with live systems rather than prompt text alone.
These include agent specifications coupled to tools and semantic data models, SQL pipelines that invoke LLMs to process textual data \citep{liskowski2025cortex}, and data-engineering workflows \citep{dbtlabs2026docs}.
We additionally evaluate standard prompt, mathematical, and systems optimization benchmarks from prior work to test whether the method remains competitive when interaction with the environment is limited or unnecessary.

\cocoevolve{} optimizes an agent specification on DABStep \citep{egg2025dabstep}, reaching 92.8\% held-out hard accuracy, versus 78.4\% for an interactive Meta-Harness baseline and 30.4\% for GEPA \citep{lee2025metaharness,agrawal2026gepa}; it also 
tops the validated public leaderboard.\footnote{At submission, this score has not been externally validated; we have contacted the DABStep maintainers but have not received a response.}
\cocoevolve{} also improves over the strongest evaluated baseline on six LLM-powered data processing pipeline workloads by up to 10 pp.
On held-out HotpotQA, it gains 11 pp over GEPA and 43 pp over the seed \citep{yang2018hotpotqa,agrawal2026gepa}. On data engineering, it achieves the strongest overall performance among the evaluated evolutionary methods and leads on held-out data-eng-bench \citep{snowflake2026dataengbench}, solving 77.4\% of tasks. Across mathematical and systems optimization objectives, it is competitive with leading evolutionary methods such as AdaEvolve \citep{cemri2025adaevolve}, with the largest improvements occurring on objectives that exhibit extended search plateaus.
Matched ablations confirm that each component of \cocoevolve{} contributes to these gains.

\section{\cocoevolve{}}
\label{sec:cocoevolve-context-curation}
We introduce \cocoevolve{}, an evolutionary optimizer built around \emph{meta-reflection} (Section~\ref{sec:meta-reflection}).
Figure~\ref{fig:cocoevolve-overview} summarizes its evolutionary loop and Algorithm~\ref{alg:cocoevolve} in Appendix~\ref{sec:app-method} specifies it in full.
Our implementation instantiates the coding-agent mutator with Cortex Code.

\begin{figure}[t]
    \centering
    \includegraphics[width=0.85\linewidth]{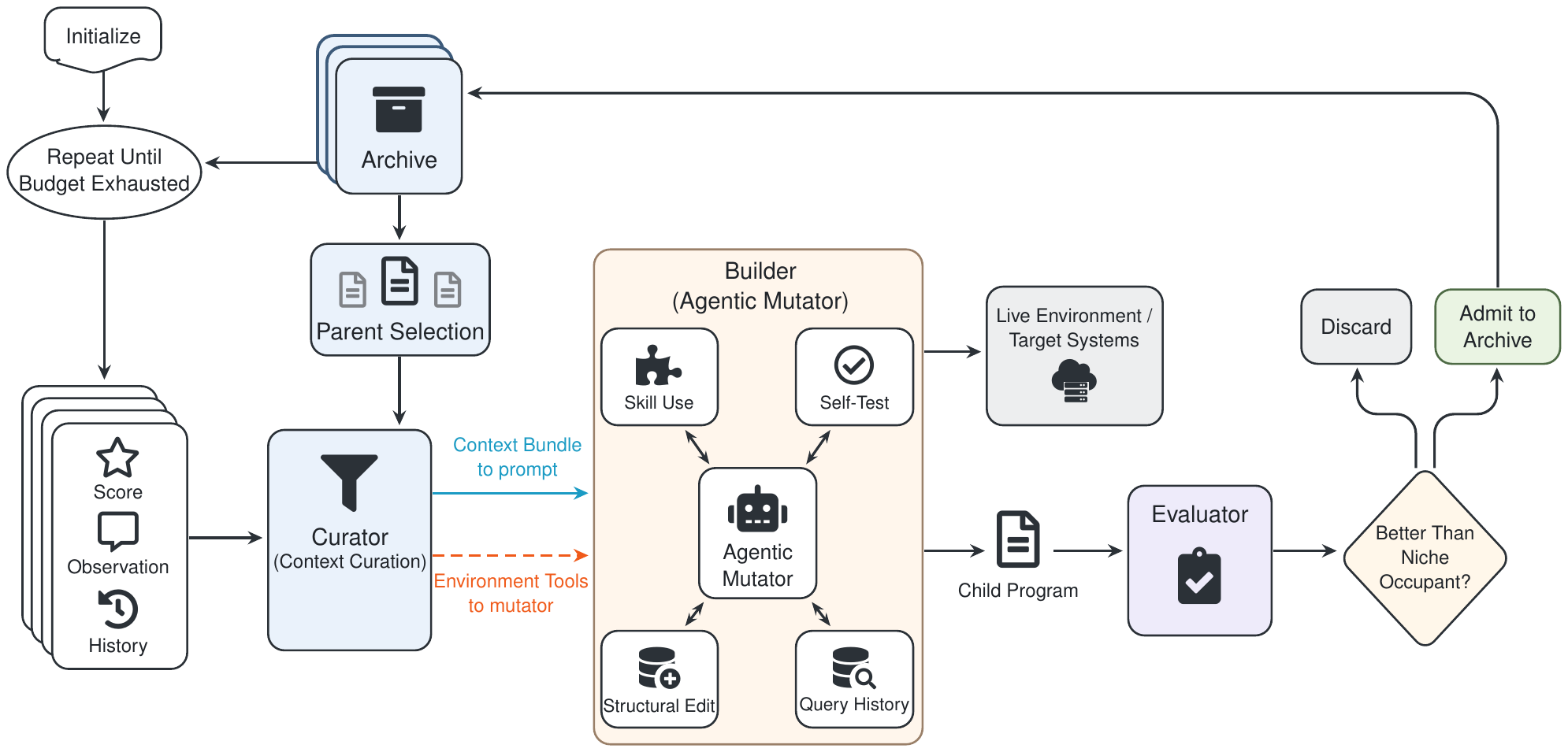}
    \caption{\cocoevolve{} proposes a new candidate in each iteration by selecting a parent from the archive, curating evidence from the search history to include in the prompt, and equipping a coding-agent mutator with on-demand tools for querying prior lineages and the live target environment. A coding agent investigates, edits, and self-tests a child; the evaluator scores it, records the resulting observations in the history, and, under the default MAP-Elites policy, either discards it or admits it to the archive if it improves its niche.
    This loop combines compact prompt-visible evidence with environment feedback while preserving diverse programs for future selection.}
    \label{fig:cocoevolve-overview}
\end{figure}

\subsection{Problem Formulation}

We treat the object of optimization as a \emph{program} $\prog$ --- code, a prompt, a configuration, an agent specification, or any other text-representable artifact --- drawn from a program space $\ProgSpace$.
Let $\Task$ denote the task space and $\TaskDist$ a distribution over it.
An evaluator
\[
f : \ProgSpace \times \Task \to \mathbb{R} \times \ObsSpace, \qquad f(\prog, \task) = \big(\score(\prog,\task),\ \obs(\prog,\task)\big)
\]
returns both a scalar score $\score(\prog,\task) \in \mathbb{R}$ and an \emph{observation} $\obs(\prog,\task) \in \ObsSpace$: the full record produced while computing that score, including tool calls, intermediate outputs, executed queries, and error messages.
The optimization goal is to find $\prog^\ast = \arg\max_{\prog \in \ProgSpace} \Fitness(\prog)$ within $\Budget$ iterations, where $\Fitness(\prog) = \mathbb{E}_{\task \sim \TaskDist}[\score(\prog,\task)]$ represents the aggregate fitness of a program.

\cocoevolve{} pursues this goal via an iterative evolutionary loop.
After iteration $t-1$, it maintains an archive $\Archive_{t-1}$ of previously evaluated candidate programs.
At iteration $t$, the harness selects a parent program $\prog'_t \in \Archive_{t-1}$ and produces a child $\prog_t$.
Most prior methods condition mutation only on the selected parent $\prog'_t$ and its observations; \cocoevolve{} instead uses the \textit{full search history} accumulated so far, denoted by
\[
\History_{t-1} = \{(\prog_i, \Task_i, \Score_i, \Obs_i)\}_{i < t},
\]
Here, $\Task_i \subseteq \Task$ is the subset selected to evaluate $\prog_i$, with $\Score_i := \{\score(\prog_i,\task) \mid \task \in \Task_i\}$ and $\Obs_i := \{\obs(\prog_i,\task) \mid \task \in \Task_i\}$ denoting the resulting sets of scores and observations.
This history supports the mutation policy developed next: context curation exposes selected evidence from $\History_{t-1}$, environment augmentation provides tools for retrieving additional evidence on demand, and the coding-agent mutator uses both to construct the child.

\subsection{Meta-Reflection}
\label{sec:meta-reflection}
We define \emph{meta-reflection} as reasoning beyond one parent's observations by selectively retrieving historical evidence from the accumulated search history and actively acquiring feedback from the target environment.

Formally, a meta-reflective mutator receives two complementary inputs.
First, a compact context bundle $\ctxbundle_t$, curated from the search history, is injected directly into the mutator's prompt.
Second, an environment $\Env_t$ provides tools that the mutator invokes on demand.
It augments a base environment $\Env_0\subseteq\Env$ of execution, validation, and target-system tools with selective access to the search history.
\[
\ctxbundle_t = \Curate(\History_{t-1}) \in \CtxSpace,
\qquad
\Env_t = \Augment(\Env_0;\prog'_t,\History_{t-1}) \subseteq \Env.
\]
A coding-agent mutator then performs meta-reflection by reasoning over $\ctxbundle_t$ and gathering additional evidence from $\Env_t$ as needed while constructing and validating the child $\prog_t$ before committing it.
\[
\prog_t \sim \MetaReflect(\prog'_t,\ctxbundle_t;\Env_t).
\]

\textbf{Relation to prior reflective mutation.}
This interface recovers prior reflective mutation as the special case in which $\ctxbundle_t=\Obs_t'$,and $\Env_t=\Env_0=\emptyset$: each revision conditions only on that trace or \textit{side information} \citep{agrawal2026optimizeanything}.
Additionally, these reflective methods implement mutation with a single model call over this preassembled evidence.

In \cocoevolve{}, these interface components take the following concrete forms.

\textbf{Ledger to gauge stagnation.}
\cocoevolve{}'s $\Curate$ operator constructs a compact ledger of recent strategies and their outcomes (Appendix~\ref{sec:app-loop}).
The agent uses this ledger to assess stagnation and choose whether to continue the current strategy, adopt a newly proposed one, or revert to undirected mutation.
AdaEvolve makes a related search-level decision through a fixed trigger, invoking tactic generation when improvement falls below a threshold \citep{cemri2025adaevolve}.
\cocoevolve{} instead lets the same class of model that performs mutation reason over concurrently tracked strategies and make this decision from the search trajectory, without a hand-tuned stagnation trigger.

\textbf{Live environment interaction.}
The base environment $\Env_0$ provides domain-specific tools for execution, testing, and target-system queries.
For example, in mathematical and systems optimization, $\Env_0$ exposes scratch runners that reveal syntax and execution errors.
In DABStep agent optimization, it provides tools for SQL execution, scratch-agent deployment, target-question tests, and deployment checks.
This feedback lets the agent diagnose and repair failures before committing a child to outer-loop evaluation.
Appendix~\ref{sec:app-tools} specifies the interfaces, and Appendix~\ref{sec:app-ablat-mutation} compares static, edit-only agentic, and self-validating mutation.

\textbf{Tools to query interlinked history.}
Beyond these domain capabilities, $\Augment$ forms $\Env_t$ by adding on-demand tools that selectively query the interlinked program archive as evidence is needed.
These tools expose per-task lineage: which program first solved a task, whether a later mutation lost that capability, and which tasks tend to co-regress.
The agent can therefore retrieve the exact diff that last restored a lost capability or vet a candidate child against known co-regressions before committing.
Appendix~\ref{sec:app-tools} and Table~\ref{tab:app-tools} specify the complete query interface, while Appendix~\ref{sec:app-ablat-history} isolates the contributions of the injected ledger and history-query tools.

After validation, we return the generated child $\prog_t$ to the outer search loop, where $f$ scores it, $\History_t$ records it, and the search policy governs archive admission and subsequent parent selection.

\subsection{Search Policy}
\label{sec:search-policy}

By default, \cocoevolve{} uses standard MAP-Elites parent selection \citep{sharma2025openevolve,mouret2015mapelites}.
For objectives scored over a pool of questions or test cases, we introduce a novel budgeted $\lp$ roster that prioritizes complementary coverage; Appendix~\ref{sec:app-lp} defines it, and Appendix~\ref{sec:app-ablat-parent} compares it with MAP-Elites, best-only selection, and pareto-frontier sampling.
 
Beyond parent selection, an optional similarity-informed task-selection policy uses proxy scores to reduce evaluation cost and focus mutation on high-information tasks.

\textbf{Similarity-guided sparse evaluation and observation curation.}
Prior methods typically expose the mutator to observations from a random task minibatch.
When the task pool $\TaskPool$ is large or individual evaluations are costly, however, exhaustively scoring every candidate is impractical.
On DABStep, for example, each agent query takes roughly 3--10 minutes, so scoring the full 300-question search pool requires 15--50 agent-hours before parallelism.
At the same time, tasks need not be independent: a repair that equips a candidate to solve one task may transfer to other tasks requiring the same capability.
This motivates sparse, targeted evaluation: score a bounded subset, infer likely performance on the remaining tasks, and direct the next mutation toward an unresolved target with broad expected transfer.

\cocoevolve{} realizes this idea with a semantic task graph.
Let $\Sim:\TaskPool\times\TaskPool\to[0,1]$ measure task similarity.
Using only evaluated tasks, it forms a Nadaraya--Watson kernel estimate of the candidate's score on $\task$ \citep{nadaraya1964estimating,watson1964smooth}:
\[
\ExpScore(\prog,\task)=\frac{\sum_{\task'\in\mathrm{tested}(\prog)}\Sim(\task,\task')\score(\prog,\task')}{\sum_{\task'\in\mathrm{tested}(\prog)}\Sim(\task,\task')},
\]
so evaluated neighbors provide a surrogate expected score for tasks not yet tested. These surrogate scores also determine where mutation effort is likely to be most informative. Each task not solved by the parent receives an information-gain score:
\[
\InfoGain(\task)=\bigl(1-\ExpScore(\prog,\task)\bigr)\delta(\task)\!\sum_{\task'\in\TaskPool}\!\Sim(\task,\task').
\]
Here, $\delta(\task)$ downweights repeatedly unsuccessful targets; sampling by $\InfoGain$ then focuses $\ctxbundle_t$ on likely failures with broad semantic reach.
This policy is a task-level counterpart to AdaEvolve's island-level Dynamic Exploration Intensity \citep{cemri2025adaevolve}: it directs scarce evaluation and mutation effort toward failures with the greatest estimated transfer.
Appendix~\ref{sec:app-infogain} details the policy, Table~\ref{tab:app-hparams} records where it is active, and matched DABStep ablations showcase its value (Appendix~\ref{sec:app-ablat-manifold}).

\section{Experiments}
\label{sec:experiments}

Where applicable, we follow a standard train/validation/test split.
\cocoevolve{} has full access to training examples during optimization.
Validation scores may be used to guide search, but the content of validation instances is never exposed to the optimizer directly.
Final results are reported on the held-out test split.
We evaluate \cocoevolve{} on four task categories: agent optimization, AI Functions, data engineering, and mathematical/system/prompt optimization.
Parent-selection variants of the archive rule are specified in Appendix~\ref{sec:app-method} and compared in Appendix~\ref{sec:app-ablat-parent}; the remaining mechanism-ablation protocol is Appendix~\ref{sec:app-ablat-mutation}--\ref{sec:app-ablat-manifold}.
Evaluator details for the math and systems suite appear in Appendix~\ref{sec:app-math}.

\subsection{Agent Optimization}

Major enterprise platforms increasingly deploy fully managed \emph{agentic frameworks} inside the data perimeter \citep
{snowflake2026cortexagents,databricks2024mosaicagent,aws2023bedrockagents}.
These agents iteratively decompose requests and use intermediate results to orchestrate semantic data access, unstructured retrieval, sandboxed code execution, and custom APIs.
Although the platform hides infrastructure and state management, reliability depends on the instructions, tool interfaces, and data models that guide this reasoning, making each a target for optimization.

\textbf{Setup.}
We formulate agent optimization as searching over \emph{agent specifications} $\prog \in \ProgSpace$ encompassing system prompts, tool schemas, and planning instructions to maximize end-to-end success across diverse user queries on tables and unstructured documents containing domain knowledge.
We evaluate on DABStep, a benchmark of 450 payment-processing questions requiring multi-step reasoning over more than nine join dimensions and wildcard joins.
The setting is challenging: the original authors report only 14.55\% accuracy for state-of-the-art agents.
Starting from the same agent specification, we run \cocoevolve{}, GEPA, and Meta-Harness for 100 iterations and compare performance on both the held-out set and the validated public leaderboard.
Because evaluations are costly, \cocoevolve{} uses information-gain guided observation curation (Section~\ref{sec:search-policy}) to direct its budget toward likely failures with broad neighborhood reach.
Appendix~\ref{sec:app-agent-method} details this instantiation.

\textbf{Results.}
Table~\ref{tab:dabstep_test} reports the train/test split for \cocoevolve{}, Meta-Harness, GEPA, and the two baselines (Snowflake Cortex Code Agent, Stock Cortex Agent).
On the official 450-question suite, \cocoevolve{} reaches 95.24\% hard accuracy, topping the validated public leaderboard; Table~\ref{tab:dabstep_combined} in Appendix~\ref{sec:app-agent} gives the full comparison.
Appendix~\ref{sec:app-agent-method} specifies the scoring protocol.

\begin{table}[htbp]
    \centering
    \setlength{\tabcolsep}{4pt}
    \begin{tabular}{l ccc ccc}
    \toprule
    & \multicolumn{3}{c}{Train (300)} & \multicolumn{3}{c}{Test (150)} \\
    \cmidrule(lr){2-4}\cmidrule(lr){5-7}
    Method & Easy & Hard & Total & Easy & Hard & Total \\
    \midrule
    Stock Cortex Agent (baseline) & 57.4 & 13.8 & 20.7 & 56.0 & 8.0 & 16.0 \\
    Snowflake Cortex Code Agent & 72.3 & 26.5 & 33.7 & 80.0 & 21.6 & 31.3 \\
    GEPA & 91.5 & 39.1 & 47.3 & 84.0 & 30.4 & 39.3 \\
    Meta-Harness & 87.2 & 76.7 & 78.3 & 80.0 & 78.4 & 78.7 \\
    \cocoevolve{} & \textbf{89.4} & \textbf{96.4} & \textbf{95.3} & \textbf{84.0} & \textbf{92.8} & \textbf{91.3} \\
    \bottomrule
    \end{tabular}
    \caption{DABStep accuracy (\%) on split (train: 47 easy / 253 hard; test: 25 easy / 125 hard).}
    \label{tab:dabstep_test}
\end{table}

\textbf{Analysis.}
Under matched splits and budgets, \cocoevolve{} outperforms GEPA and Meta-Harness even when the latter is equipped with self-verification and self-testing tools that provide a comparable pre-commit validation surface (Appendix~\ref{sec:app-agent-method}); three mechanisms explain this gap.

\textit{Live environment interaction grounds mutation in execution feedback.}
DABStep intentionally tests domain rules whose downstream implications must be inferred or composed across documentation and data, a failure mode emphasized by the benchmark authors \citep{egg2025dabstep}.
The live environment turns this ambiguity from guesswork into testable hypotheses.
The Cortex Code mutator repeatedly formed fee interpretations, tested them through live SQL against a known merchant total, and revised its diagnosis.
This live feedback exposed two hidden logic errors and enabled a verified vectorized fee engine, producing the run's largest single gain, from 28 to 95 correct answers (Appendix~\ref{sec:app-agent-programs}).

\textit{Curated environment tools compress otherwise lengthy investigations.}
The base environment exposes scripts, managed skills, and developer tools that encode reusable investigation and validation workflows, allowing the mutator to invoke proven procedures instead of reconstructing them across mutation cycles.
During the fee-engine revision, the mutator first checked its raw SQL, then used an existing testing skill to deploy the draft as a live scratch agent and exercise multiple question families as self-validation before commit.
Each passed in a single tool call, whereas the prior specification failed after roughly 40 tool calls, allowing the mutator to invest its budget in structural reasoning rather than verification plumbing (Appendix~\ref{sec:app-tools}).

\emph{Structural changes are more robust than prompt changes.}
Reflective prompt evolution is effective \citep{agrawal2026gepa}, but a single model call over preassembled evidence offers no path to explore and validate changes to the underlying tool architecture.
In a matched DABStep run, GEPA revised the instructions but left the default single-row fee calculator tool unchanged.
Its evolved agent therefore aggregates 26 transaction fees through 26 tool calls and asks the LLM to sum the outputs in context, producing arithmetic drift such as 19.51 instead of 19.36.
Such remedies require live environment interaction: inspecting schemas, executing draft SQL, and validating outputs against known totals before commit.
By testing structural edits in situ, \cocoevolve{} discovered two independently verified remedies: a vectorized fee-engine stored procedure and a dynamic table that materializes per-transaction fees for downstream SQL aggregation.
These solutions shift computation from fallible in-context arithmetic into deterministic system components, illustrating how coding-agent mutation can advance where prompt-only search plateaus (Appendix~\ref{sec:app-agent-programs}).

\subsection{AI Functions}
\label{sec:ai-functions}

Modern cloud data platforms expose \emph{AI functions}, declarative SQL operators that embed LLM inference inside query engines for row-wise tasks such as generation, classification, and extraction \citep{liskowski2025cortex, databricks2023aifunctions, microsoft2024azureai}.
Industry workloads compose these operators with common table expressions (CTEs) and joins into multi-stage pipelines, so reliability depends jointly on prompts, output schemas, and pipeline topology.
For instance, a pipeline may first use a generation function to identify personally identifiable information (PII) spans and define a JSON schema, then pass this plan to a downstream function that performs contextual redaction.
We therefore search over executable SQL programs $\prog \in \ProgSpace$, allowing \cocoevolve{} to restructure both LLM prompts and operator composition for diverse row distributions, missing data, and edge cases.

\textbf{Setup.}
Each task starts from a minimal SQL program $\prog_0$ containing one generative AI-function call, which the mutator may rewrite into a multi-stage, multi-operator pipeline.
\emph{Lockpicking with Interference} (Appendix~\ref{sec:app-lockpicking}) scores exact match on 50 labeled six-digit strings governed by a hidden ruleset.
Beyond this synthetic diagnostic, we evaluate \textit{classification} on Reddit comments (\texttt{go\_emotions}) and lengthy legal documents (\texttt{lex\_glue}) \citep{demszky2020goemotions, chalkidis2022lexglue}.
We also study \textit{financial numerical reasoning}: \texttt{finqa} requires multi-step arithmetic over tables and prose, while \texttt{convfinqa} distributes such reasoning across a dialogue \citep{chen2021finqa, chen2022convfinqa}.
Finally, \texttt{multilexnorm} requires normalizing noisy, nonstandard social-media tokens into canonical forms \citep{vandergoot2021multilexnorm}.

Because evaluation is inexpensive, this setting omits similarity-informed targeting.
The mutator can instead execute candidate pipelines and repair syntax, schema, or engine-specific failures before committing, which is especially valuable for brittle SQL programs.
\textbf{Results.}
Table~\ref{tab:ai_function_results} shows that \cocoevolve{} outperforms the strongest evaluated baseline on all six tasks across classification, numerical reasoning, text normalization, and rule discovery.
The largest gains are a 10 pp improvement that exactly recovers Lockpicking's hidden rules and a 7.7 pp gain on \texttt{multilexnorm}, whose intermingled, context-dependent normalization patterns echo the latent-rule challenge isolated by Lockpicking.

\begin{table}[htbp]
    \centering
    {\setlength{\aboverulesep}{0pt}\setlength{\belowrulesep}{0pt}%
    \adjustbox{max width=\textwidth}{%
    \begin{tabular}{l cccccc}
    \toprule
     & \multicolumn{6}{c}{\texttt{claude-sonnet-5}} \\
    \cmidrule{2-7}
    Method & \texttt{lockpicking\_6} & \texttt{go\_emotions} & \texttt{finqa} & \texttt{convfinqa} & \texttt{lex\_glue} & \texttt{multilexnorm} \\
    \midrule
    GEPA          & 0.820 & 0.425 & 0.800 & 0.920 & 0.735 & 0.510 \\
    AdaEvolve     & 0.900 & 0.366 & 0.740 & 0.885 & 0.695 & 0.470 \\
    OpenEvolve    & 0.540 & 0.375 & 0.770 & 0.900 & 0.645 & 0.470 \\
    \cocoevolve{} & \textbf{1.000} & \textbf{0.444} & \textbf{0.813} & \textbf{0.953} & \textbf{0.737} & \textbf{0.587} \\
    \bottomrule
    \end{tabular}}}
    \caption{AI-function benchmark performance.}
    \label{tab:ai_function_results}
\end{table}

\textbf{Analysis.}
We next examine two mechanisms behind these gains.

\emph{On-demand history queries help escape local minima.}
When a highly coupled objective plateaus, reasoning only over the current parent's failures may not expose a route out.
AdaEvolve's adaptive scheduler illustrates this limitation on Lockpicking \citep{cemri2025adaevolve}: it first reaches 90\% at iteration 34; over the remaining 16 iterations, two stagnation-triggered paradigm generations propose targeted worked examples and exhaustive condition matrices for the five residual failures, yet its best score never moves.
\cocoevolve{} reaches exact recovery at iteration 6, before AdaEvolve's first paradigm generation at iteration 10, by using on-demand history tools to query evidence across lineages rather than repeatedly reframing the same residual errors (Appendix~\ref{sec:app-lockpicking}).

\emph{Self-validation prevents wasted mutation cycles.}
Without a cheap validator, mutation can waste substantial search budget on SQL programs that never execute.
On \texttt{go\_emotions}, GEPA repeatedly placed \texttt{LATERAL FLATTEN} inside a UDF; because each compiler error collapsed to a zero score without diagnostic feedback, broken variants consumed up to 71\% of the budget.
Alternatives such as OpenEvolve mitigate evaluation cost with cascading evaluators that stop after compilation failure, but each rejected candidate still consumes an evolutionary iteration \citep{sharma2025openevolve}.
\cocoevolve{} instead executes drafts on a cheap sample and repairs syntax failures within the same multi-turn mutation before commit.
Self-validation thus converts compiler errors from terminal scores into actionable context, reserving outer-loop evaluation for executable programs.

\subsection{Data Engineering}
Data engineering is the process of transforming raw warehouse data into
trusted, queryable models. Agents often fail to construct or repair
transformations due to challenges in managing the relevant context, such as
data warehouse details, implicit relationships, and domain-specific business
rules. We optimize over context-curation pipeline specifications
$p \in \mathcal{P}$ that govern the creation of a context-grounded knowledge
graph (KG) and a suite of executable adherence tests derived from it, both of which
are supplied to a downstream agent prior to executing a transformation. We
evaluate on data-eng-bench \citep{snowflake2026dataengbench}, a data-engineering benchmark comprising 103 tasks over an enterprise
warehouse with 579 tables, 19 business schemas, and more than 8,000 columns.

\textbf{Setup.}
The seed pipeline $\prog_0$ is a single agent with a minimal prompt and the scaffolding for three skills---lean human priors for authoring KGs and the adherence tests. The mutator may refine the prompt, introduce new skills, contribute to existing ones, or reorganize the pipeline into a multi-agent protocol. $\Fitness$ is evaluated through a two-stage process: the mutator evolves the upstream pipeline $\prog$ while the downstream agent remains fixed; a score is produced only after the agent consumes the generated KG, iterates over the corresponding adherence tests, and implements the transformation that is scored by the benchmark's verifier. We defer further details to Appendix~\ref{sec:app-de}.

\textbf{Results.}
As shown in Table~\ref{tab:data-eng}, \cocoevolve{} raises the solve rate on the test split from 59.1\% at baseline to 77.4\%, surpassing GEPA at 73.6\% and Meta-Harness and AdaEvolve at 71.7\%. On the dev split, \cocoevolve{} reaches 68.0\%, tying Meta-Harness for the highest solve rate and exceeding the baseline, AdaEvolve, and GEPA by 14.0, 8.0, and 6.0 pp, respectively.

\begin{table*}[htbp]
\centering
\caption{Solve rate (\%) on the dev and test splits of \textbf{data-eng-bench}.}
\label{tab:de-results}
\setlength{\tabcolsep}{5pt}
\adjustbox{max width=\textwidth}{%
\begin{tabular}{l cc cc cc cc cc}
\toprule
& \multicolumn{10}{c}{\texttt{claude-sonnet-5}} \\
\cmidrule(lr){2-11}
Dataset & \multicolumn{2}{c}{Baseline} & \multicolumn{2}{c}{Meta-Harness} & \multicolumn{2}{c}{GEPA} & \multicolumn{2}{c}{AdaEvolve} & \multicolumn{2}{c}{\cocoevolve{}} \\
\cmidrule(lr){2-3}\cmidrule(lr){4-5}\cmidrule(lr){6-7}\cmidrule(lr){8-9}\cmidrule(lr){10-11}
& Dev & Test & Dev & Test & Dev & Test & Dev & Test & Dev & Test \\
\midrule
\textbf{data-eng-bench} & 54.0 & 59.1 & 68.0 & 71.7 & 62.0 & 73.6 & 60.0 & 71.7 & 68.0 & \textbf{77.4} \\
\bottomrule
\end{tabular}}
\label{tab:data-eng}
\end{table*}

\textbf{Analysis.} \textit{Meta-reflection preserves globally relevant skills.} During evolution, an earlier lineage introduced a skill to resolve a recurring dbt configuration error. A later hillclimb on an unrelated target dropped the skill. When the failure re-appeared,  CoCoEvolve's mutator retrieved the archived mutation plan via the ledger that first named the fix and re-implemented the skill, rather than re-deriving it from scratch. We observe that the mutator matches returning failures against past resolutions in the evolution ledger, linking current errors to the earlier mutations that fixed them. Meta-reflection thus allows for capabilities the search has already learned to remain visible to the mutator even when not directly present in the parent's trace.

\subsection{Mathematical, Systems, and Prompt Optimization}
\label{sec:math_system}
\label{sec:prompt-optimization}

These experiments ask whether \cocoevolve{} can match specialized optimizers on established mathematical and systems benchmarks and whether its benefits persist on HotpotQA, AIME, and IFBench from GEPA's benchmark suite when live interaction is unavailable.
Within each task, all arms share the task definition, initial artifact, backbone, evaluator, and data split where applicable, enabling matched end-to-end comparisons across executable programs and prompt-only systems.

\textbf{Setup.}
On math and systems tasks, each evaluator returns one scalar for the complete program. \cocoevolve{} uses a lightweight reflector to detect stagnation and decide whether to continue, request a new direction, or revert to undirected mutation.
When a new direction is needed, a distinct proposer queries techniques from any lineage while preserving the mutator's turn for implementation.
Appendix~\ref{sec:app-prompts} specifies the complete reflector-proposer protocol and prompts.
AIME and IFBench use GPT-5.6-luna because Sonnet-5 saturates AIME at the seed, leaving no optimization signal.
The main table omits standard deviations for readability; bold and underlining mark the best and second-best results within each backbone, with ties sharing marks.
Appendix~\ref{sec:app-math} reports standard deviations, while Appendix~\ref{sec:app-prompt-opt-method} details the prompt protocol.

\textbf{Results.}
Table~\ref{tab:optimization_results} reports the mathematical and ADRS systems benchmarks.
\cocoevolve{} is best or tied best in 25 of 44 Avg/Best cells, including 13 of 24 math and 12 of 20 ADRS cells, more than any other method.
On Sonnet-5, the largest gains occur on plateau-prone Signal Processing, TXN, and Heilbronn Convex; on Signal Processing, \cocoevolve{} reaches $0.766$ versus AdaEvolve's $0.747$.
Gaps largely vanish on saturated objectives, while mixed GPT-5.6-terra results show that the advantage depends on the backbone.
Appendix~\ref{sec:app-math-programs} analyzes plateau escape, structural family switching, and saturation controls task by task.
Table~\ref{tab:prompt_opt} reports prompt optimization, where \cocoevolve{} leads GEPA and the seed on every HotpotQA and IFBench split, while AIME is inconsistent under repeated evaluation.

\begin{table*}[t]
    \centering
    \scriptsize
    \setlength{\tabcolsep}{2.5pt}
    \adjustbox{max width=\textwidth}{%
    \begin{tabular}{l cc cc cc cc cc cc cc cc cc cc cc}
    \toprule
     & \multicolumn{12}{c}{Mathematical Optimization} & \multicolumn{10}{c}{ADRS Systems} \\
    \cmidrule(lr){2-13} \cmidrule(lr){14-23}
     & \multicolumn{2}{c}{Circle Packing (Square)} & \multicolumn{2}{c}{Circle Packing (Rect)} & \multicolumn{2}{c}{Heilbronn (Triangle)} & \multicolumn{2}{c}{Heilbronn (Convex)} & \multicolumn{2}{c}{MinMax Distance} & \multicolumn{2}{c}{Signal Processing} & \multicolumn{2}{c}{Cloudcast ($\downarrow$)} & \multicolumn{2}{c}{EPLB} & \multicolumn{2}{c}{Prism} & \multicolumn{2}{c}{LLM-SQL} & \multicolumn{2}{c}{TXN} \\
     & \multicolumn{2}{c}{($N=26$)} & \multicolumn{2}{c}{($N=21$)} & \multicolumn{2}{c}{($N=11$)} & \multicolumn{2}{c}{($N=13$)} & \multicolumn{2}{c}{($d=3, N=14$)} & \multicolumn{2}{c}{} & \multicolumn{2}{c}{} & \multicolumn{2}{c}{} & \multicolumn{2}{c}{} & \multicolumn{2}{c}{} & \multicolumn{2}{c}{} \\
    \cmidrule(lr){2-3} \cmidrule(lr){4-5} \cmidrule(lr){6-7} \cmidrule(lr){8-9} \cmidrule(lr){10-11} \cmidrule(lr){12-13} \cmidrule(lr){14-15} \cmidrule(lr){16-17} \cmidrule(lr){18-19} \cmidrule(lr){20-21} \cmidrule(lr){22-23}
    Strategy & Avg & Best & Avg & Best & Avg & Best & Avg & Best & Avg & Best & Avg & Best & Avg & Best & Avg & Best & Avg & Best & Avg & Best & Avg & Best \\
    \midrule
    Human/SOTA    & -- & 2.636 & -- & 2.366 & -- & 0.037 & -- & 0.031 & -- & 0.240 & -- & 0.718 & -- & 626.200 & -- & 0.127 & -- & 21.890 & -- & 0.692 & -- & 2{,}725 \\
    \midrule
    \multicolumn{23}{l}{\textit{Backbone: Sonnet-5}} \\
    OpenEvolve       & \textbf{2.636} & \textbf{2.636} & \textbf{2.366} & \textbf{2.366} & 0.032 & \underline{0.035} & 0.028 & 0.029 & \underline{0.238} & \underline{0.240} & 0.650 & 0.673 & 660.693 & 660.689 & 0.127 & 0.127 & 26.236 & \underline{26.256} & 0.713 & 0.726 & 3{,}945 & 4{,}016 \\
    GEPA             & 2.602 & \textbf{2.636} & \underline{2.365} & \textbf{2.366} & \textbf{0.036} & \textbf{0.037} & \underline{0.030} & \textbf{0.031} & \textbf{0.240} & \textbf{0.240} & 0.603 & 0.631 & 621.090 & \underline{618.080} & 0.133 & \underline{0.145} & \underline{26.256} & \underline{26.256} & 0.524 & \textbf{0.752} & 3{,}874 & 4{,}000 \\
    AdaEvolve        & 2.633 & \textbf{2.636} & 2.364 & \textbf{2.366} & \underline{0.035} & \textbf{0.037} & 0.029 & \underline{0.031} & \textbf{0.240} & \textbf{0.240} & \underline{0.747} & \underline{0.766} & \underline{618.097} & 618.088 & \underline{0.145} & 0.145 & \textbf{26.391} & \textbf{26.660} & \underline{0.724} & \underline{0.733} & \underline{4{,}115} & \underline{4{,}132} \\
    \cocoevolve{}    & \underline{2.635} & \textbf{2.636} & 2.363 & \textbf{2.366} & 0.035 & \textbf{0.037} & \textbf{0.031} & \textbf{0.031} & \textbf{0.240} & \textbf{0.240} & \textbf{0.766} & \textbf{0.784} & \textbf{618.063} & \textbf{618.060} & \textbf{0.145} & \textbf{0.146} & \underline{26.256} & \underline{26.256} & \textbf{0.728} & 0.732 & \textbf{4{,}292} & \textbf{4{,}348} \\
    \midrule
    \multicolumn{23}{l}{\textit{Backbone: GPT-5.6-terra}} \\
    OpenEvolve       & \textbf{2.636} & \textbf{2.636} & \underline{2.365} & \underline{2.366} & 0.030 & 0.031 & 0.025 & 0.027 & 0.240 & 0.240 & 0.565 & 0.608 & 618.096 & 618.093 & 0.127 & 0.127 & 26.076 & 26.152 & \underline{0.724} & \underline{0.730} & 4{,}107 & 4{,}184 \\
    GEPA             & \underline{2.634} & \textbf{2.636} & 2.364 & \underline{2.366} & \textbf{0.035} & \underline{0.035} & \textbf{0.026} & \textbf{0.028} & \textbf{0.240} & \textbf{0.240} & 0.633 & 0.661 & \textbf{618.081} & \textbf{618.070} & 0.127 & 0.127 & \underline{26.259} & \underline{26.264} & 0.719 & 0.724 & \textbf{4{,}268} & \textbf{4{,}348} \\
    AdaEvolve        & 2.633 & \underline{2.634} & \textbf{2.366} & \textbf{2.366} & \underline{0.035} & 0.035 & 0.024 & 0.024 & \underline{0.240} & \underline{0.240} & \textbf{0.792} & \textbf{0.813} & 618.085 & \textbf{618.070} & \underline{0.142} & \textbf{0.145} & 25.674 & 25.734 & 0.717 & 0.724 & 4{,}144 & 4{,}167 \\
    \cocoevolve{}    & 2.634 & \textbf{2.636} & 2.364 & \underline{2.366} & 0.034 & \textbf{0.037} & \underline{0.025} & \underline{0.027} & \textbf{0.240} & \textbf{0.240} & \underline{0.756} & \underline{0.805} & \underline{618.082} & \underline{618.071} & \textbf{0.145} & \underline{0.145} & \textbf{26.338} & \textbf{26.501} & \textbf{0.728} & \textbf{0.730} & \underline{4{,}250} & \underline{4{,}329} \\
    \bottomrule
    \end{tabular}}
    \caption{Mathematical and ADRS systems benchmarks.
    Higher is better except for Cloudcast.
    Human/SOTA is the maximum published human or automated result reported by AlphaEvolve and AdaEvolve.
    Full-precision means, standard deviations, and best values appear in Tables~\ref{tab:math_results_full} and~\ref{tab:systems_results_full}.}
    \label{tab:optimization_results}
    \label{tab:math_results}
    \label{tab:systems_results}
\end{table*}

\begin{table}[htbp]
    \centering
    \adjustbox{max width=\textwidth}{%
    \begin{tabular}{l ccc ccc ccc}
    \toprule
    & \multicolumn{3}{c}{HotpotQA (Sonnet-5)} & \multicolumn{3}{c}{AIME (luna)} & \multicolumn{3}{c}{IFBench (luna)} \\
    \cmidrule(lr){2-4}\cmidrule(lr){5-7}\cmidrule(lr){8-10}
    Method & Train & Val & Test & Train & Val & Test & Train & Val & Test \\
    \midrule
    Baseline (unmodified seed) & 0.393 & 0.313 & 0.363 & $\mathbf{0.919 \pm 0.056}$ & $\mathbf{0.882 \pm 0.026}$ & $0.833 \pm 0.000^\dagger$ & $0.902 \pm 0.003$ & $0.864 \pm 0.003$ & $0.663 \pm 0.027$ \\
    GEPA & 0.713 & 0.700 & 0.683 & $\mathbf{0.919 \pm 0.078}$ & $0.830 \pm 0.056$ & $0.833 \pm 0.058$ & $0.891 \pm 0.018$ & $0.886 \pm 0.009$ & $0.683 \pm 0.005$ \\
    \cocoevolve{} & \textbf{0.800} & \textbf{0.783} & \textbf{0.793} & $\mathbf{0.919 \pm 0.026}$ & $\mathbf{0.882 \pm 0.046}$ & $\mathbf{0.878 \pm 0.051}$ & $\mathbf{0.927 \pm 0.010}$ & $\mathbf{0.921 \pm 0.009}$ & $\mathbf{0.705 \pm 0.024}$ \\
    \bottomrule
    \end{tabular}}
    \caption{Prompt optimization on HotpotQA, AIME, and IFBench.
    HotpotQA reports a single Sonnet-5 run; AIME and IFBench report mean $\pm$ standard deviation over three evaluations of each final candidate using \texttt{openai-gpt-5.6-luna}; these repeats measure evaluation noise, not search variance.
    Bold marks column-best results that are robust to measurement noise; $^\dagger$the observed zero deviation is coincidental.}
    \label{tab:prompt_opt}
\end{table}

\textbf{Analysis. }
Across the mathematical and systems benchmarks, \cocoevolve{} recovers the gains of AdaEvolve's specialized three-level adaptive controller through meta-reflection over the search history, offering a more flexible route to stagnation-aware search.

\textit{The strategy ledger turns plateaus into new search directions.}
On Signal Processing, the global best remained at 0.7534 for 26 iterations even though the current parent and champion represented distinct Savitzky-Golay and Butterworth filter families.
Reasoning over the strategy ledger, the lightweight reflector recognized a population-wide plateau and requested a new proposal rather than another local mutation or crossover.
Using on-demand history queries to identify what the search had not tried, the separate proposer found an orthogonal family: an L1 trend filter formulated as a linear program.
The mutator implemented this paradigm and raised the score to 0.7745 in one iteration.
This trace illustrates how the strategy ledger and on-demand history tools separate stagnation detection, paradigm selection, and implementation: the search history reveals not only which ideas succeeded, but which structural alternatives remain unexplored.

\textit{On-demand history queries improve prompt optimization even without live environment interaction.}
On HotpotQA, \cocoevolve{} reaches $0.793$ held-out accuracy versus GEPA's $0.683$ and the seed's $0.363$, an 11 pp margin in GEPA's own setting.
IFBench preserves the ordering \cocoevolve{} $>$ GEPA $>$ seed on every repeat and split, although the held-out margin is narrower.
AIME provides no reliable separation: all arms lie in the same $0.83$--$0.88$ test band, while $17$--$27\%$ of individual answers change correctness across repeated evaluations of an unchanged prompt.
Together, these results isolate the value of on-demand history access for text-only artifacts, while AIME illustrates how stochastic evaluation can erase apparent search gains.

\section{Related Work}
\label{sec:related-works}
\textbf{Evolutionary Harnesses.}
LLM-guided evolutionary optimization shifts improvement from training-time parameter updates toward inference-time search, using LLMs as semantic mutation operators over prompts, algorithms, and programs.
Building on quality-diversity search \citep{mouret2015mapelites}, systems such as FunSearch, AlphaEvolve, and OpenEvolve apply this paradigm to scientific and algorithmic discovery \citep{romera2024funsearch, novikov2025alphaevolve, sharma2025openevolve}.
GEPA enriches mutation with language-valued execution traces that expose reasoning, tool calls, outputs, and errors beyond scalar rewards \citep{agrawal2026gepa}.
Subsequent work broadens the optimized artifact space through unified declarative APIs \citep{agrawal2026optimizeanything}, allocates search adaptively \citep{cemri2025adaevolve, lange2025shinkaevolve}, and standardizes outer harnesses \citep{ursekar2025vero}.
For prompts and LM pipelines, EvoPrompt and Promptbreeder evolve prompts with LLM-driven operators \citep{guo2025evoprompt, fernando2024promptbreeder}, MIPROv2 jointly searches instructions and demonstrations for multi-stage programs \citep{opsahl2024mipro}, and TextGrad backpropagates natural-language feedback through compound systems \citep{yuksekgonul2025textgrad}.
These methods generally retain a parent-local, one-shot mutation interface in which the harness preselects evidence and evaluates the proposed edit only after generation.
Promptbreeder also evolves its mutation prompts, whereas \cocoevolve{} changes the mutation interface itself through meta-reflection over the search history and target environment.
Meta-Harness \citep{lee2025metaharness} similarly uses a coding-agent proposer with filesystem access to prior traces; \cocoevolve{} combines compact curated context with structured, on-demand queries over evolutionary history and a live environment.

\textbf{Context Curation Approaches.}
The efficacy of LLM agents relies heavily on rich, domain-specific context, such as organizational schemas \citep{dohmen2026contextlayer} or dynamically constructed failure taxonomies \citep{cojocaru2026taxonomies}.
Recent frameworks automate context engineering by treating contexts as evolving playbooks \citep{zhang2026ace} or co-evolving curation skills alongside artifacts \citep{ye2026mce}.
\cocoevolve{} applies this principle to the optimization trajectory itself, distilling accumulated search history into a compact ledger.

\textbf{Limitations of Evolutionary Approaches.}
AI-driven evolutionary frameworks exhibit systematic vulnerabilities, including hidden regressions \citep{zhou2026aichilles}, overfitting to the initial task set under continual learning \citep{wang2026compound}, and context costs that grow with the accumulated archive \citep{lee2025metaharness}.
\cocoevolve{} addresses these failure modes with bounded, queryable access to historical co-regressions, helping the optimizer avoid cycling through brittle solutions that break previously solved cases.

\section{Conclusion}
\label{sec:conclusion}
We present \cocoevolve{}, an evolutionary optimizer for production AI systems.
By both probing the environment and querying an evolutionary ledger and prior lineages, \cocoevolve{}'s mutator extends parent-local reflection into \emph{meta-reflection} over the search process, enabling it to diagnose failures, reconstruct hidden mechanisms, preserve complementary discoveries, and verify structural revisions before committing.
Across domains, our results suggest that the combination of curating selected historical evidence, providing on-demand tools for probing the live environment and evolutionary history, and using a coding-agent mutator to reason over both offers an effective path to improving the prompts, code, tools, data models, and pipelines of real-world AI systems.

\subsubsection*{Acknowledgments}
We thank Danmei Xu, Garett Tok, Zihan Li, Varich Boonsanong, Kyle Schmaus, and Boxin Jiang for insightful discussions, feedback and suggesting relevant datasets.

\bibliography{bibliography}
\bibliographystyle{iclr2027_conference}

\appendix

\section{Table of Contents}
\label{sec:app-toc}

This appendix is organized as follows.
Appendix~\ref{sec:app-method} specifies the evolutionary loop (Appendix~\ref{sec:app-loop}), the history-query and workspace tools (Appendix~\ref{sec:app-tools}), the similarity-graph InfoGain branch used by agent evolution (Appendix~\ref{sec:app-infogain}), the experimental hyperparameters (Appendix~\ref{sec:app-hparams}), and the budgeted $\lp$ roster (Appendix~\ref{sec:app-lp}).
Appendix~\ref{sec:app-ablations} reports parent selection, mutation and self-validation, history access, and similarity-manifold targeting ablations. The history-access study includes Signal Processing and DABStep. Each ablation subsection reports its trial count and score provenance.
Appendix~\ref{sec:app-prompts} records the mutator, reflector, and proposer prompts.
Appendix~\ref{sec:app-agent} gives the DABStep protocol, analyzes Meta-Harness's post-improvement churn, and presents an evolved agent instruction.
Appendix~\ref{sec:app-aifn} presents the Lockpicking synthetic diagnostic (Appendix~\ref{sec:app-lockpicking}) and describes the AI-function experiments and their evolved artifacts.
Appendix~\ref{sec:app-de} describes the data-engineering pipeline, its evaluation, and an evolved artifact.
Appendix~\ref{sec:app-math} records math and ADRS evaluator settings and the detailed result tables (Appendix~\ref{sec:app-math-full-precision}).
Appendix~\ref{sec:app-math-programs} gives the per-task analysis: which methods lead each cell, then Heilbronn Convex, Circle Packing, Cloudcast, and TXN traces (Figures~\ref{fig:heilbronn-layouts}--\ref{fig:circle-packing-search}).
Appendix~\ref{sec:app-prompt-opt} gives the methodology for all three prompt-optimization tasks (Appendix~\ref{sec:app-prompt-opt-method}) and a sim-graph-targeted, cascaded-evaluation efficiency variant of HotpotQA (Appendix~\ref{sec:app-prompt-opt-simgraph}).
Appendix~\ref{sec:app-program-listings}, placed last, includes the complete base and final programs for Heilbronn Convex, Cloudcast, and TXN.

\section{Use of Large Language Models (LLMs) and other AI Tools}
\label{sec:app-llmusage}

The authors used large language models (LLMs) to assist in literature discovery and polishing the manuscript's prose. The authors reviewed and verified all suggested references, claims, and revisions, and assume full responsibility for the final content. Coding assistants, including Snowflake Cortex Code, Anthropic Claude, and Cursor, helped implement the authors' ideas. The methodology, scientific contributions, and core results are the authors' original work.

\section{Algorithm and Method}
\label{sec:app-method}

This appendix instantiates the operators of Section~\ref{sec:cocoevolve-context-curation}.
Section~\ref{sec:app-loop} gives the evolutionary loop; Section~\ref{sec:app-tools} lists the history tools added through $\Augment$ and the domain tools supplied by $\Env_0$; Section~\ref{sec:app-infogain} records the kernel, decay, and softmax used for curated observations; Section~\ref{sec:app-hparams} lists the hyperparameters used in the experiments of Section~\ref{sec:experiments}; and Section~\ref{sec:app-lp} specifies the budgeted $\lp$ roster used as a population objective and as a parent-selection policy.

\subsection{Evolutionary loop}
\label{sec:app-loop}

\begin{algorithm}[ht]
\caption{\cocoevolve{} evolutionary loop.}
\label{alg:cocoevolve}
\begin{algorithmic}[1]
\Require seed $\prog_0$, budget $\Budget$, evaluator $f$, parent policy $\pi$, base environment $\Env_0 \subseteq \Env$
\State $\Archive_0 \gets \{\prog_0\}$, $\History_0 \gets \emptyset$
\For{$t = 1,\ldots,\Budget$}
    \State $\prog'_t \sim \pi(\Archive_{t-1})$ \Comment{Parent selection}
    \State $\ctxbundle_t \gets \Curate(\History_{t-1})$ \Comment{Context curation}
    \State $\Env_t \gets \Augment(\Env_0;\prog'_t,\History_{t-1})$ \Comment{Environment augmentation}
    \State $\prog_t \sim \MetaReflect(\prog'_t,\ctxbundle_t;\Env_t)$ \Comment{Agentic investigation}
    \State select $\Task_t \subseteq \Task$; evaluate $\prog_t$ with $f$ to obtain $\Score_t$ and $\Obs_t$
    \State $\History_t \gets \History_{t-1}\cup\{(\prog_t,\Task_t,\Score_t,\Obs_t)\}$
    \State update $\Archive_{t-1}$ with $\prog_t$ to form $\Archive_t$
\EndFor
\State \Return $\arg\max_{\prog\in\Archive_{\Budget}} \Fitness(\prog)$
\end{algorithmic}
\end{algorithm}

The loop selects parent $\prog'_t$ from $\Archive_{t-1}$ using $\pi$; the parent policy uses either MAP-Elites or the $\lp$ roster of Section~\ref{sec:app-lp}.
Context curation prepares $\ctxbundle_t$, while environment augmentation adds on-demand history-query tools to $\Env_0$ to form $\Env_t$.
Depending on the task, $\Curate$ reflects on the strategy ledger and selects a target instance using the similarity graph of Section~\ref{sec:app-infogain}.
Because operational requirements vary by domain, these preparation mechanisms are layered according to the search objective.
We detail the exact workflow applied for each experiment in its task subsection.

In general, the injected bundle $\ctxbundle_t$ contains the approach ledger (recent strategies and their score deltas), the active paradigm text if one is in force, and island elites.
When an instance pool is defined, $\ctxbundle_t$ also includes the curated target observation and any donor program that has already solved it.
The augmented environment $\Env_t$ combines the domain tools in $\Env_0$ with the opt-in history tools of Section~\ref{sec:app-tools}.
The Cortex Code coding-agent mutator then performs $\MetaReflect$ by investigating, implementing, and validating the child, after which admission is typically a MAP-Elites update on a complexity--diversity grid.

\subsection{On-demand history and environment tools}
\label{sec:app-tools}

Table~\ref{tab:app-tools} lists the on-demand tools provisioned in $\Env_t$.
The mutator can call these tools from its task workspace while editing and validating a candidate. The ledger is injected every iteration; tool results enter the context only when requested, so the full archive need not appear in the prompt.
For archive access, the mutator calls \texttt{python3 evo.py} to query prior programs, lineages, scores, and traces.
For local execution, the constrained \texttt{evo\_sweep.py} runner executes the candidate without granting unrestricted command execution.
The math, ADRS, Signal Processing, and DABStep experiments all expose lineage and trajectory queries.
Per-task queries are enabled for DABStep and Signal Processing, whose evaluators expose instance-level or per-signal outcomes, and disabled for the remaining math and ADRS tasks, which return only a scalar score.
DABStep additionally overlays warehouse tools that issue SQL, deploy a scratch agent, and pre-screen a draft against a target question before committing the child.

\begin{table}[htbp]
\centering
\adjustbox{max width=\textwidth}{%
\begin{tabular}{lll}
\toprule
Channel & Tool & Returns \\
\midrule
\multirow{5}{*}{Lineage}
  & \texttt{get\_program} & code body of a program \\
  & \texttt{get\_parent} / \texttt{get\_lineage} & parent id; children with score deltas \\
  & \texttt{get\_best\_programs} & top-$N$ programs by primary score \\
  & \texttt{get\_diverse\_programs} & best program per occupied MAP-Elites cell \\
  & \texttt{get\_architecture\_history} & per-cell landing history \\
\midrule
\multirow{2}{*}{Trajectory}
  & \texttt{get\_recent\_scores} & recent primary-score trace \\
  & \texttt{get\_thinking\_trace} & tool/thinking events of iteration $I$ \\
\midrule
\multirow{4}{*}{Per-task}
  & \texttt{get\_question\_solvers} & top programs on one instance, with parent \\
  & \texttt{get\_question\_memory} & child$<$parent regressions on one instance \\
  & \texttt{get\_suggested\_questions} & instances ranked worst-first for a program \\
  & \texttt{get\_hard\_questions} & instances by population-best score, ascending \\
\midrule
\multirow{5}{*}{Warehouse (DABStep)}
  & \texttt{query} & read-only SQL against live tables \\
  & \texttt{execute\_sql} & allowed write SQL against the warehouse \\
  & \texttt{test\_agent} / \texttt{test\_draft} & score a scratch agent on question $Q$ \\
  & \texttt{check\_deploy} & compile-check one \texttt{CREATE} DDL \\
  & \texttt{analyst\_query} & text-to-SQL against the semantic view \\
\bottomrule
\end{tabular}}
\caption{On-demand tools available in $\Env_t$. Lineage, trajectory, and per-task tools are added through $\Augment$ to query the evolutionary archive; the DABStep base environment $\Env_0$ additionally provides live-warehouse validation tools.}
\label{tab:app-tools}
\end{table}

\subsection{Curated observations: kernel, decay, and softmax}
\label{sec:app-infogain}

Section~\ref{sec:search-policy} writes a generic similarity kernel $\Sim$.
This section applies to the DABStep agent-evolution setting, which evaluates a program on many instances and maintains a similarity graph over that pool.
Math, ADRS, and Signal Processing have no such pool and skip this channel, even though Signal Processing still exposes per-signal query tools (Table~\ref{tab:app-hparams}).
We precompute an embedding-similarity matrix $\Sim$ on $\TaskPool$, with values in $[0,1]$. DABStep specializes the estimator in Section~\ref{sec:search-policy} by replacing $\Sim(\task,\task')$ with $\Sim(\task,\task')^4$ in both the numerator and denominator, giving more weight to nearby tested instances. After each mutation, a local success or regression update refines the estimate. Estimates for untested coordinates are capped at $0.7$, and coordinates with observed failures are floored at $0.1$.

The next target $\task^\ast$ is drawn by softmax over $\InfoGain$ with temperature $0.1$.
Candidates exclude instances the parent has already solved (score $\geq 0.9$).
The decay in Section~\ref{sec:search-policy} is instantiated as
\[
\delta(\task)
=
\begin{cases}
1 & \text{if any program in $\Archive_t$ has solved $\task$}, \\
(1/2)^{c(\task)} & \text{otherwise},
\end{cases}
\]
where $c(\task)$ is the number of times $\task$ has been selected as the mutation target without success.
A solve anywhere in the archive therefore resets the penalty, so a newly acquired donor can be reused; repeated failures on an unsolved instance geometrically suppress it.

\subsection{Hyperparameters}
\label{sec:app-hparams}

Table~\ref{tab:app-hparams} records the settings used in Section~\ref{sec:experiments}.
The default math, ADRS, and DABStep runs sample parents from MAP-Elites islands using exploration, elite selection, and fitness-weighted draws. Math and ADRS use exploration weight $0.2$ and elite-selection weight $0.65$. The AI-function objective and the parent-selection ablation instead use the $\lp$ roster of Section~\ref{sec:app-lp}.
InfoGain targeting (Section~\ref{sec:app-infogain}) is on for DABStep only.

\begin{table}[htbp]
\centering
\adjustbox{max width=\textwidth}{%
\begin{tabular}{lccc}
\toprule
 & Math / ADRS & Signal Processing & DABStep \\
\midrule
Iteration budget $\Budget$ & 100 & 100 & 100 \\
Islands / population & 2 / 50 & 2 / 80 & 1 / 40 \\
InfoGain targeting & off & off & on \\
Per-task query tools & off & on & on \\
Task workspace + history queries & on & on & on \\
Warehouse / scratch-agent tools & off & off & on \\
Cascade thresholds & task-dependent$^\ast$ & $[0.3, 0.6]$ & $[0.005]$ \\
LLM temperature & $0.7$ (framework default) & $0.6$ & $0.7$ \\
\bottomrule
\end{tabular}}
\caption{Domain hyperparameters and tool availability. All listed domains edit candidates in a task workspace and expose on-demand history queries; DABStep additionally enables InfoGain targeting and warehouse tools. $^\ast$Cascade evaluation is enabled for Circle Packing (Square) with thresholds $[0.5, 0.8]$ and disabled for the remaining math and ADRS tasks.}
\label{tab:app-hparams}
\end{table}

Signal Processing exposes per-task query tools because its evaluator returns per-signal scores, although it has no $\TaskPool$ and does not use InfoGain. AI functions use the $\lp$ roster and full validation evaluation rather than InfoGain targeting; they are omitted from the table because their remaining settings follow Section~\ref{sec:ai-functions}.

\subsection{Budgeted Population \texorpdfstring{$\lp$}{Lp} Objective and Parent Selection}
\label{sec:app-lp}

Let $\Archive_t$ denote the archive after iteration $t$.
Suppose each program $\prog\in\Archive_t$ is evaluated along $N$ dimensions, with per-dimension score $\score_i(\prog)$. A dimension may represent a task instance, objective, or behavioral criterion. We associate each program with the set
\[
\mathcal{I}(\prog)=\{i\in[N]:\score_i(\prog)\geq\theta_i\}
\]
of dimensions on which it is considered successful. The thresholds $\theta_i$ may be absolute or determined relative to the current population. For binary per-dimension scores, $\mathcal{I}(\prog)$ is exactly the set of dimensions solved by the program. For continuous scores, we use either a calibrated absolute threshold or a per-dimension population quantile, allowing each dimension to be compared against its own achievable score range.

\paragraph{Budgeted diversity roster.}
Although $\History_t$ retains the complete evaluation history, only a budgeted active roster $\mathcal{R}_t\subseteq\Archive_t$, $|\mathcal{R}_t|\leq K$, is used to represent the population and select the next parent. Starting with $\mathcal{R}^{(0)}_t=\emptyset$ and $C^{(0)}=\emptyset$, we greedily choose
\[
\prog_j
=
\arg\max_{\prog\in\Archive_t\setminus\mathcal{R}^{(j-1)}_t}
    |\mathcal{I}(\prog)\setminus C^{(j-1)}|,
\]
define its marginal contribution as
\[
m_j = |\mathcal{I}(\prog_j)\setminus C^{(j-1)}|,
\]
and update
\[
\mathcal{R}^{(j)}_t
=
\mathcal{R}^{(j-1)}_t\cup\{\prog_j\},
\qquad
C^{(j)}=C^{(j-1)}\cup \mathcal{I}(\prog_j).
\]
Construction stops after $K$ programs or when no remaining program contributes new coverage. This is the standard greedy maximum-coverage construction: it retains programs because of capabilities not already represented by higher-ranked roster members, rather than solely because of aggregate fitness.

\paragraph{Population $\lp$ score.}
Given the ordered roster marginals $\mathbf{m}=(m_1,\ldots,m_{|\mathcal{R}_t|})$, we define
\[
L_p(\mathcal{R}_t)
=
\frac{\|\mathbf{m}\|_p}{N}
=
\frac{1}{N}
\left(\sum_j m_j^p\right)^{1/p}.
\]
Equivalently, the evaluator may report its $p$-th power
\[
J_p(\mathcal{R}_t)
=
L_p(\mathcal{R}_t)^p
=
\frac{\sum_j m_j^p}{N^p},
\]
which induces the same ordering for fixed $p$. At $p=1$, the score is the fraction of dimensions covered by the roster. For $p>1$, it also rewards consolidation: for the same total coverage, the score is higher when fewer programs carry that coverage. Greedy roster construction retains complementary capabilities, while the objective rewards programs that combine them.

\paragraph{$\lp$ as an optimization objective.}
For tasks in which population quality is itself the target, a newly evaluated child $\prog_t$ is assessed by inserting it into the candidate pool, rebuilding the roster, and measuring
\[
\Delta J_p
=
J_p(\mathcal{R}_t\mid\Archive_t\cup\{\prog_t\})
-
J_p(\mathcal{R}_t\mid\Archive_t).
\]
The evolutionary loop then directly optimizes $J_p$. This is the formulation used in our AI-function setting with $p=2$, where $J_2=\sum_j m_j^2/N^2$ rewards both broad population coverage and the consolidation of specialist capabilities.

\paragraph{$\lp$ as a parent-selection policy.}
When the task retains its original scalar objective, the same roster can instead be used only to choose the next parent. We assign each roster member probability
\[
\Pr(\prog_j\mid\mathcal{R}_t)
=
\frac{m_j^p}{\sum_\ell m_\ell^p}.
\]
This samples programs according to their contribution to the population $\lp$ score without replacing the task's evaluator. At $p=0$, the policy is uniform over roster members; at $p=1$, it is proportional to exclusive marginal coverage; our default $p=2$ emphasizes large complementary contributions quadratically; and as $p\rightarrow\infty$, it concentrates on a maximum-marginal roster member. An agentic variant exposes the same roster, marginals, and population score to an LLM and allows it to choose the parent directly, using the mechanical distribution as a fallback.
Table~\ref{tab:lp_exponent_hotpotqa} reports a HotpotQA sweep of $p\in\{0.5,1,1.5,2\}$ for this parent-selection policy. The \texttt{lp} arm uses $p=2$.

\subsubsection{Rare-capability carrier selection}

Marginal coverage alone can under-allocate mutations to valuable specialists. Suppose most programs solve the same large set of easy dimensions, while different programs solve only one or two distinct hard dimensions. The greedy roster retains these specialists, but for $p>1$ its mechanical distribution can place nearly all probability on the first member, whose marginal includes the common easy dimensions. To remove this dependence on raw instance frequency, let
\[
\rho_i
=
\frac{1}{|\Archive_t|}
\sum_{\prog\in\Archive_t}\mathbb{I}[i\in \mathcal{I}(\prog)],
\qquad
h_i=(1-\rho_i)^\gamma,
\]
where $\rho_i$ is the population solve rate and $\gamma>0$ controls how strongly rare dimensions are emphasized. We define the hard-capability mass carried by a program as
\[
H(\prog)
=
\sum_{i=1}^{N} h_i\,\mathbb{I}[i\in \mathcal{I}(\prog)].
\]
We incorporate rare-capability mass into parent selection by sampling according to
\[
q_{\mathrm{rare}}(\prog)
\propto
\exp\!\left(
\beta\left[
\lambda\,\widetilde{\Fitness}(\prog)
+(1-\lambda)H(\prog)^p
\right]\right),
\]
where $\widetilde{\Fitness}$ is normalized aggregate fitness. The fitness term favors the strongest carrier of a given rare capability, rather than weak programs in general, while $H(\prog)^p$ gives increasing preference to lineages that accumulate multiple hard capabilities. If all specialists carry one equally rare capability and have equal fitness, they remain indistinguishable and are sampled symmetrically; once one lineage combines multiple rare capabilities, its selection probability increases.
Because no selector can exploit a dimension that no current program solves, this policy can be mixed with the baseline selector,
\[
q(\prog)
=
(1-\epsilon)q_{\mathrm{rare}}(\prog)
+\epsilon q_{\mathrm{base}}(\prog).
\]
The rare-capability term is part of the parent-selection mechanism. It changes which existing program is selected for mutation without adding a donor or changing the evaluation pool. All candidates are evaluated on the same instances, so their capabilities remain directly comparable.

\section{Ablation Details}
\label{sec:app-ablations}

This section reports four ablations against the mechanisms of Appendix~\ref{sec:app-method} and Section~\ref{sec:experiments}.
Parent selection (Appendix~\ref{sec:app-ablat-parent}) compares four parent-selection setups and examines the $\lp$ sampling exponent.
Three claim-driven ablations each isolate one further axis of the evolutionary loop: mutation execution and self-validation (Appendix~\ref{sec:app-ablat-mutation}), the history-access interface (Appendix~\ref{sec:app-ablat-history}), and similarity-manifold targeting (Appendix~\ref{sec:app-ablat-manifold}).

\paragraph{Common controls.}
\label{sec:app-ablat-controls}
Each claim-driven ablation below estimates the effect of one named treatment while holding the rest of the loop fixed; a comparison that changes two axes at once is not reported as a causal ablation.
Unless stated as the named treatment, every arm of a given ablation shares the same backbone model, temperature, and model-facing mutation budget; the same seed program and train/test split; the same parent-selection policy and population configuration; the same editable artifact surface; the same number of outer-loop iterations; and the same evaluator thresholds and per-child evaluation allowance.
Persistent per-program and archive-level solved-question state is retained across arms, so a no-graph arm does not remember only the current minibatch.
The tables below report the final metrics available for each ablation. Each subsection states the number of trials.
These controls are already in force below unless a subsection states an exception.

\subsection{Parent selection}
\label{sec:app-ablat-parent}

We test whether useful parent selection requires both preserving complementary capabilities and directing mutations toward programs that can consolidate them. The four-way comparison examines which programs remain eligible as parents; the HotpotQA exponent sweep probes how mutation effort is allocated among them.

\paragraph{Setup.}
We implement four parent-selection policies in the same OpenEvolve codebase. Within each task, all arms use the same seed, split, backbone, mutation process, island configuration, and 100-iteration budget. Only the parent-selection policy changes. \texttt{select-best} always chooses the current best program. \texttt{map-elites} uses OpenEvolve's exploration/exploitation/fitness draw from the MAP-Elites grid. \texttt{pareto-frontier} samples the per-instance Pareto frontier by win count, as in GEPA. \texttt{lp} builds a greedy $K$-roster from per-instance solved sets and samples member $j$ with probability proportional to $m_j^p$, where $m_j$ is its marginal coverage (Appendix~\ref{sec:app-lp}); here $p=2$. The frontier and roster draw from all islands. Table~\ref{tab:parent_selection_scope_ablation} compares the complete setups on each task.

To probe whether stronger preference for large marginal contributions helps the roster consolidate specialists, we compare HotpotQA configurations with $p\in\{0.5,1,1.5,2\}$ while keeping the roster construction fixed. All four use \texttt{openai-gpt-5.6-luna}, a 3-example mutation minibatch from train-150, val-300 exact match for search and export, and one 100-iteration search per setting. The $p=2$ run reuses the HotpotQA \texttt{lp} result above; $p$ is the only treatment varied in this comparison.

\paragraph{Results.}

Table~\ref{tab:parent_selection_scope_ablation} shows the results for the four parent-selection setups and Table~\ref{tab:lp_exponent_hotpotqa} shows the results for the HotpotQA $\lp$ sweep.

\begin{table}[htbp]
    \centering
    \begin{tabular}{l cc cc}
    \toprule
    & \multicolumn{2}{c}{HotpotQA} & \multicolumn{2}{c}{IFBench} \\
    \cmidrule(lr){2-3}\cmidrule(lr){4-5}
    Parent selection & Test EM & $\Delta$ & Test score & $\Delta$ \\
    \midrule
    \texttt{select-best}     & 0.6400 & $-0.0267$ & 0.6592 & $-0.0360$ \\
    \texttt{map-elites}      & 0.6667 & $0$       & 0.6952 & $0$ \\
    \texttt{pareto-frontier} & 0.6433 & $-0.0234$ & 0.6918 & $-0.0034$ \\
    \texttt{lp} ($p=2$)     & \textbf{0.6933} & $\mathbf{+0.0266}$ & \textbf{0.7003} & $\mathbf{+0.0051}$ \\
    \bottomrule
    \end{tabular}
    \caption{Parent-selection setups on HotpotQA and IFBench. One 100-iteration search per arm; entries are single held-out test evaluations. HotpotQA uses exact match; IFBench uses its per-constraint score. Each $\Delta$ is the score difference from OpenEvolve's default \texttt{map-elites} arm on the same task. The \texttt{lp} row uses $p=2$. Bold marks the highest observed score in each task.}
    \label{tab:parent_selection_scope_ablation}
\end{table}

\begin{table}[htbp]
    \centering
    \begin{tabular}{l ccc}
    \toprule
    Roster exponent & Validation EM & Test EM & $\Delta$ test \\
    \midrule
    $p=0.5$ & 0.6033 & 0.5867 & $-0.0733$ \\
    $p=1$   & 0.7100 & 0.6600 & $0$ \\
    $p=1.5$ & 0.5967 & 0.5700 & $-0.0900$ \\
    $p=2$ & \textbf{0.7300} & \textbf{0.6933} & $\mathbf{+0.0333}$ \\
    \bottomrule
    \end{tabular}
    \caption{HotpotQA validation and held-out test exact match across $\lp$ roster exponents. $\Delta$ is the test-EM difference from $p=1$; bold marks the highest observed score.}
    \label{tab:lp_exponent_hotpotqa}
\end{table}

\paragraph{Analysis.}
These selectors use different evidence to identify useful parents. \texttt{select-best} spends every mutation on the current incumbent, leaving complementary capabilities in other programs unused. \texttt{map-elites} preserves programs in distinct grid cells, but those cells need not correspond to distinct solved instances. \texttt{pareto-frontier} uses per-instance wins, yet win-count weighting does not discount wins already represented by another parent. The greedy $\lp$ roster instead retains a program only while it adds uncovered instances, then favors roster members with larger exclusive contributions. The $\lp$ roster has the highest observed held-out score on both tasks in Table~\ref{tab:parent_selection_scope_ablation}, consistent with allocating mutations by complementary coverage rather than grid diversity alone.

The exponent sweep probes how to allocate mutations after complementary parents have entered the roster. Flatter weighting spends more turns on small-marginal specialists; quadratic weighting concentrates ordinary mutations on parents that already contribute many distinct solves, giving later edits more opportunities to consolidate those capabilities. The $p=2$ row leads on both validation and held-out test, consistent with a balance between consolidation pressure and specialist access. The dip at $p=1.5$ shows that the effect is not simply monotonic in $p$: changing the weighting changes which lineages receive repeated edits, and those search paths can produce different final programs.

\subsection{Mutation execution and self-validation}
\label{sec:app-ablat-mutation}

This ablation separates two benefits of an interactive mutator: editing the current candidate with access to its environment, and testing that candidate before committing it.
The first can support broad instruction-level fixes; the second is especially useful for changes to executable code or dependent objects.

\paragraph{Setup.}
We run this ablation on Heilbronn Convex ($N{=}13$) and DABStep. Within each task, the three arms share the parent policy, population and island settings, feature grid, history-access interface, iteration budget, targeting configuration, and reflector/proposer configuration. They differ in the mutator's action space. Heilbronn Convex has no instance similarity graph; DABStep keeps its similarity graph enabled across all three arms. The reflector/proposer remains on in every arm, matching the published configurations.
A1 (\emph{static mutator}) is one LLM completion over the parent, ledger, and active paradigm; it cannot inspect the workspace or execute anything.
A2 (\emph{agentic, no candidate testing}) can edit the candidate and query history through \texttt{evo.py}, but cannot execute the edited candidate before committing it.
A3 (\emph{full mutator}, published) adds pre-commit execution: a local Python sweep on Heilbronn Convex, or deployment and end-to-end testing of the edited agent on DABStep.
On DABStep, A1 can only diagnose from injected artifacts, A2 can additionally query the database but cannot deploy the edited agent, and A3 has the complete query, deployment, and testing interface.

\paragraph{Results.}
We report three Sonnet-5 trials per Heilbronn Convex arm. One trial stopped after 83 iterations; the others ran for 100. Its score is included in the reported summary, so that arm's mean is not based on three equal-budget trials.
One 100-iteration trial is reported per arm on DABStep.
All DABStep Train/Test values are fresh full-pool re-evaluations of each arm's champion rather than in-run estimates based on limited evaluations; A1's initial estimate used a 25-question bootstrap sample.
A1 never advanced past the first evaluation gate, so its champion is the unmodified seed also reported in Table~\ref{tab:dabstep_test} and as A1 in Table~\ref{tab:manifold_ablation}.
A3 on both tasks reuses the published champion rather than a new run; the DABStep champion is also the C3 arm in Table~\ref{tab:manifold_ablation}.
Each $\Delta$ is the raw-count difference from A1.

\begin{table}[htbp]
    \centering
    \adjustbox{max width=\textwidth}{%
    \begin{tabular}{l cc cccc}
    \toprule
    & \multicolumn{2}{c}{Heilbronn Convex (min triangle area / hull area)} & \multicolumn{4}{c}{DABStep} \\
    \cmidrule(lr){2-3}\cmidrule(lr){4-7}
    Arm & Mean $\pm$ Std & Best & Train (of 300) & $\Delta$ & Test (of 150) & $\Delta$ \\
    \midrule
    A1 static mutator & $0.0260 \pm 0.0003$ & $0.0266$ & $101$ & -- & $47$ & -- \\
    A2 agentic, no candidate test & $0.0270 \pm 0.0002$ & $0.0276$ & $202$ & $+101$ & $98$ & $+51$ \\
    A3 full mutator (\cocoevolve{}) & $0.03066 \pm 0.00036$ & $0.03094$ & $286$ & $+185$ & $137$ & $+90$ \\
    \bottomrule
    \end{tabular}}
    \caption{Mutation-execution ablation on Heilbronn Convex and DABStep. Values summarize three trials per arm for Heilbronn Convex and one full-pool champion evaluation per arm for DABStep; $\Delta$ is relative to A1.}
    \label{tab:mutation_ablation}
\end{table}

\paragraph{Analysis.}
Performance improves monotonically from the static mutator (A1), to agentic editing without pre-commit candidate testing (A2), to the full mutator (A3) on both tasks.
The mean Heilbronn Convex score rises from $0.0260$ to $0.0270$ to $0.03066$, while DABStep test accuracy rises from $47/150$ to $98/150$ to $137/150$.
The two transitions add different capabilities: A1 to A2 adds direct editing and on-demand inspection, while A2 to A3 adds execution of the edited candidate before it is committed.

Heilbronn Convex shows that the A1-to-A2 improvement is not explained by broader exploration: both arms attempt a similar range of strategies.
The difference is how those ideas are incorporated.
A1 frequently replaces large portions of the solver; in one matched trial, roughly one quarter of its scored edits regress and another third have no effect.
A2 generally preserves the existing solver and adds components around it, reducing regressions to about one in eight, although it learns whether an edit worked only after evolutionary evaluation.
A3 can test proposed changes through scratch execution and regresses on roughly $4\%$ of scored edits across the trials.
Agentic editing therefore helps preserve prior functionality, while self-validation turns more of the budget into measured refinement.

DABStep reveals why the relative gains differ across tasks.
A1 sometimes identifies real instruction problems from injected context, but cannot inspect the database or run the revised agent.
A2 can query the database, allowing it to correct fee-rule enumeration; the same accepted edit also tightens the response contract for exact-match answers.
Because both changes affect many questions, they raise train/test accuracy from $33.7\%/31.3\%$ to $67.3\%/65.3\%$.
This closes roughly $55\%$ of the A1-to-A3 gap, compared with about $22\%$ on Heilbronn Convex.
A3 additionally deploys and tests the edited agent, allowing it to validate a structural replacement for the fee-computation procedure and reach $95.3\%/91.3\%$.
Thus, database-grounded editing can capture broad instruction-level improvements, whereas executable-code and dependent-object rewrites benefit most from end-to-end validation.

\subsection{History-access interface}
\label{sec:app-ablat-history}

We test two ways to give the mutator access to earlier search: a short ledger injected into each turn and structured tools for retrieving prior programs and traces. The four arms compare each mechanism alone and both together on Signal Processing and DABStep.

\paragraph{Setup (Signal Processing).}
The selected parent's code and evaluation are available in every arm; this ablation changes access to other programs' history. B1 (\emph{parent-only}) receives no additional history. B2 (\emph{ledger-only}) receives recent approaches, score changes, stagnation information, and the active paradigm in its prompt, but cannot fetch prior programs or traces. B3 (\emph{tools-only}) can request earlier programs, lineage, scores, and thinking traces through \texttt{evo.py}, but receives no ledger. B4 (\emph{full}) has both mechanisms. All four arms use Sonnet-5 for 100 iterations with the same population, islands, evaluator, and model-facing mutation budget. The evolution graph and reflector/proposer are off throughout. B4 therefore uses the published Signal Processing history-access interface without the reflector/proposer.

\paragraph{Results (Signal Processing).}
Table~\ref{tab:history_interface_ablation} reports the best \texttt{overall\_score} reached in one trial per arm.

\begin{table}[htbp]
    \centering
    \begin{tabular}{l cc c}
    \toprule
    Arm & Injected ledger & Structured history tools & \texttt{overall\_score} \\
    \midrule
    B1 parent-only  & off & off & $0.6614$ \\
    B2 ledger-only  & on  & off & $0.7074$ \\
    B3 tools-only   & off & on  & $0.7487$ \\
    B4 full         & on  & on  & $\mathbf{0.7681}$ \\
    \bottomrule
    \end{tabular}
    \caption{History-access interface ablation, Signal Processing. One 100-iteration Sonnet-5 trial per arm, reflector/proposer off throughout. Best \texttt{overall\_score} reached within the 100-iteration budget (iterations 73, 97, 99, and 75 respectively).}
    \label{tab:history_interface_ablation}
\end{table}

\paragraph{Analysis (Signal Processing).}
Scores rise from $0.6614$ for B1 to $0.7074$ with the ledger alone (B2) and $0.7487$ with tools alone (B3). B4, with both mechanisms, reaches $0.7681$. Thus each mechanism helps relative to parent-only in these runs, and the combination scores highest. The B4 score is near the published Sonnet-5 mean of $0.7656$ (best $0.7841$), although the reflector/proposer is disabled in this ablation.

The four final Signal Processing programs use different filtering designs. B1 combines a quadratic ridge-trend filter with a noise-adaptive moving average. B2's champion describes an adaptive Kalman filter. B3 adds Rauch--Tung--Striebel backward smoothing to a Kalman filter. B4 combines Kalman and Savitzky--Golay smoothing with a median prefilter and slope-hysteresis logic.

The inspected winning turns illustrate how the interfaces are used. B1, with no access to other attempts, proposes a larger-window Savitzky--Golay filter and Holt smoothing, tests both, rejects them, and keeps a small parameter change. B3 first queries the best programs, recent scores, and lineage, then verifies a refinement of its Kalman/smoothing filter with local score sweeps. B4 recognizes a plateau of more than thirty iterations, queries high-scoring and diverse programs, and uses a local parameter sweep to find an improvement at a narrow threshold boundary. The B4 turn shows the ledger's broad progress summary and targeted history lookup being used in the same edit.

\paragraph{Setup (DABStep).}
DABStep uses the same four ledger/history-tool settings as Signal Processing. Its evolution graph and reflector/proposer remain enabled in every arm. The graph selects target questions on the controller side, while this ablation changes the history visible to the mutator; turning the graph off would change a second part of the loop. The reflector/proposer also stays fixed across arms. For B1 and B2, the core \texttt{evo.py} history commands and the task-specific \texttt{get\_performance}, \texttt{get\_suggested\_questions}, and \texttt{get\_artifacts} history commands are removed. B3 and B4 retain them. All four arms keep the six commands used to inspect and test the current candidate: \texttt{query}, \texttt{test\_agent}, \texttt{analyst\_query}, \texttt{check\_deploy}, \texttt{execute\_sql}, and \texttt{test\_draft}.

\paragraph{Results (DABStep).}
We report one primary 100-iteration Sonnet-5 trial per arm. Each champion is freshly scored on all 300 training and 150 held-out test questions. B4 reuses the published flagship champion. A second B4 run is discussed below.

\begin{table}[htbp]
    \centering
    \begin{tabular}{l cc cc}
    \toprule
    Arm & Injected ledger & Structured history tools & Train (of 300) & Test (of 150) \\
    \midrule
    B1 parent-only  & off & off & $234$ ($0.780$) & $116$ ($0.773$) \\
    B2 ledger-only  & on  & off & $210$ ($0.700$) & $99$ ($0.660$) \\
    B3 tools-only   & off & on  & $242$ ($0.807$) & $119$ ($0.793$) \\
    B4 full         & on  & on  & $\mathbf{286}$ ($\mathbf{0.953}$) & $\mathbf{137}$ ($\mathbf{0.913}$) \\
    \bottomrule
    \end{tabular}
    \caption{History-access interface ablation on DABStep. One primary 100-iteration Sonnet-5 trial per arm; each champion is freshly scored on the full 300-question training and 150-question test pools. Rows follow B1--B4 arm order.}
    \label{tab:history_interface_dabstep}
\end{table}

\paragraph{Analysis (DABStep).}
The four primary runs show a different pattern from Signal Processing. Ledger-only B2 scores below parent-only B1 on both train ($210$ versus $234$) and test ($99$ versus $116$). Tools-only B3 is only slightly above B1 ($242$ train and $119$ test). Full B4 reaches $286$ on train and $137$ on test, the highest result in this comparison. On DABStep, the large observed gain appears when both history-access mechanisms are enabled; neither mechanism alone shows a comparable gain. This is the central result of the DABStep four-arm table.

\paragraph{A useful fix lost at the evaluation gate.}
The B1 and B2 traces show that discovering a useful fix does not guarantee it survives search. Both arms repeatedly identify a seed instruction that lists only five merchants from an older dataset version. For an unlisted merchant, the seed can answer ``Not Applicable'' without calling a tool, even though the current data contains thirty valid merchants. B1 repairs the instruction, and its assigned target question also improves, so the edit enters the surviving lineage. B2 makes and tests the same kind of repair in its first iteration, but its assigned target is unrelated to the merchant bug; the evaluation gate rejects the edit, and later candidates still inherit the stale instruction.

\paragraph{The ledger helps when the mutator can inspect prior programs.}
The ledger gives the mutator a live-updated record of earlier approaches and projected scores. Its score column is revised as the similarity mechanism estimates full-pool performance, so a ledger value can differ from the program's immediate measured result. In one traced case, a program scored $0.1767$ on its immediate evaluation but later appeared in the ledger at $0.4011$. B1 sees no such history table. B2 receives the changing projections on every turn but cannot retrieve the earlier programs and traces needed to judge what a high-scoring entry actually does. The ledger-only arm is therefore exposed to uncertain historical scores without a way to check the underlying work, which helps explain its weaker result. B3 and B4 can inspect that work through their history tools.

\paragraph{How B3 and B4 use history.}
B3 trails B1 through much of the search and gains most in its final thirty iterations. Before two late improvements, the mutator queries prior programs, lineage, scores, and thinking traces. It then builds a counterfactual simulator for questions about alternate fee schemes and lookup cubes for card-scheme and ACI questions, testing both before acceptance. B1 also attempts substantial late changes, including a case bank for difficult questions; the distinction in these traces is that B3 consults the population's history before deciding where to work, not that B1 is incapable of complex edits.

B4 reaches B1's final score by roughly iteration $40$ and improves further. Around iteration $36$, it queries earlier high-scoring programs and suggested questions, then builds and checks a vectorized aggregation engine for group-by questions. The checks expose an unsupported array/object parameter type, which the mutator works around before the edit is kept. Around iteration $84$, it adds a computation mode for ACI-selection questions and checks it against known cases. These episodes show the full arm using its history tools while carrying out substantial, verified edits.

\paragraph{Variation within the full configuration.}
A second B4 run uses the same configuration and model as the flagship but scores $248/300$ on train and $114/150$ on test, rather than $286/300$ and $137/150$. It ends near B3's primary $242/300$ and $119/150$ result despite having both history-access mechanisms available. The two B4 runs have similar Bash-call volume per iteration ($32.3$ versus $33.7$), but allocate those calls differently. In the flagship, $2092$ of $3229$ Bash calls invoke \texttt{evo.py} and $974$ invoke \texttt{evo\_sweep.py}, about $95\%$ combined. In the second run, $1671$ of $3367$ invoke \texttt{evo.py} and $127$ invoke \texttt{evo\_sweep.py}, about $53\%$ combined. The remaining calls, many of which inspect the edited blueprint directly, are far more common in the second run (about $47\%$ versus $5\%$). The tool-use pattern and outcome vary together in these two runs; tool use was observed rather than assigned as another ablation treatment.

\subsection{Similarity-manifold targeting}
\label{sec:app-ablat-manifold}

We test whether the true question-similarity graph improves DABStep target selection and expansion. The full semantic configuration (C3) scores well above the uniform and shuffled controls in these runs.

\paragraph{Setup.}
All three arms use DABStep's 300-question training pool and keep the mutator, history-access interface, parent policy, first-evaluation gate, regression-sample size, expansion-question count, scout count, and acceleration budget fixed. The evolution graph remains enabled in every arm. The arms differ in how they select a target and expand evaluation to previously untested questions, as described below.

C1 (\emph{uniform matched control}) samples the target uniformly from the same unsolved candidate set, keeps the regression step identical, and evaluates the same number of random previously untested questions, including the same scout count; persistent solved-state, donor registry, and target-attempt counts are retained.
C2 (\emph{shuffled similarity graph}) runs C3's identical information-gain formula, but reads the similarity term off a symmetric index permutation of the real matrix rather than the real values --- same value distribution, wrong question pairing, isolating whether the specific semantic geometry matters rather than merely having \emph{some} structured (non-uniform) sampling bias.
C3 (\emph{full semantic manifold}) selects the target by information gain over questions the parent has not solved, runs the usual regression sample from the parent's cumulative solved set, and expands the search to semantically similar previously untested questions; archive-wide solvers may be retrieved as donors.
All three arms share the same regression-risk sampler: whatever target a given arm's rule selects, that target's most at-risk previously-solved questions are drawn from the true, unpermuted similarity graph. Only target selection and untested-question expansion differ across arms.

C3's champion is the flagship run already reported in Table~\ref{tab:dabstep_test}; we report a fresh full-pool re-evaluation of that same champion here, rather than the in-run bookkeeping trajectory used elsewhere, to match C1/C2's methodology exactly.

\paragraph{Results.}
One 100-iteration Sonnet-5 trial per fresh arm (C1, C2); C3 and A1 reuse the flagship and seed champions already reported elsewhere, each freshly re-evaluated against the same full pools as C1/C2.

\begin{table}[htbp]
    \centering
    \begin{tabular}{l cc cc}
    \toprule
    Arm & Train (of 300) & $\Delta$ & Test (of 150) & $\Delta$ \\
    \midrule
    A1 seed (no evolution) & $101$ & -- & $47$ & -- \\
    C1 uniform (matched)    & $95$ & $-6$ & $42$ & $-5$ \\
    C2 shuffled similarity  & $113$ & $+12$ & $51$ & $+4$ \\
    C3 information-gain (flagship) & $\mathbf{286}$ & $\mathbf{+185}$ & $\mathbf{137}$ & $\mathbf{+90}$ \\
    \bottomrule
    \end{tabular}
    \caption{Similarity-manifold targeting on DABStep. C1 and C2 each use one 100-iteration Sonnet-5 search; C3 reuses the published flagship and A1 the seed. All four champions are freshly scored on the full 300-question training and 150-question test pools. $\Delta$ is relative to A1. The flagship's in-run trace reached $293$ on train, while its independent full-pool re-score here is $286$.}
    \label{tab:manifold_ablation}
\end{table}

\paragraph{Analysis.}
C3 reaches $286/300$ on train and $137/150$ on test, well above C1 ($95/300$, $42/150$) and C2 ($113/300$, $51/150$). C1 scores below the seed ($101/300$, $47/150$), while C2 is only slightly above it. A repeat held-out evaluation of C2's champion scored $49/150$ rather than $51/150$, underscoring the small size of that margin.

The per-question comparison in Table~\ref{tab:manifold_diff} shows both more gains and fewer regressions for C3: it gains $189$ questions and regresses $4$, versus $29$ gained/$35$ regressed for C1 and $28$ gained/$16$ regressed for C2. Most gains in every arm transfer beyond its directly targeted questions (C3: $143/189$; C1: $25/29$; C2: $22/28$). All arms use the true graph for regression-risk sampling, so aligning target selection and expansion with that graph may help protect previously solved neighbors. These runs evaluate those two semantic operations together; they do not separate their contributions.

\begin{table}[htbp]
    \centering
    \adjustbox{max width=\textwidth}{%
    \begin{tabular}{l rrr rr}
    \toprule
    Arm & Gained & Regressed & Net & Gain via target & Gain via transfer \\
    \midrule
    C1 uniform (matched)    & $29$  & $35$ & $-6$   & $4$  & $25$ \\
    C2 shuffled similarity  & $28$  & $16$ & $+12$  & $6$  & $22$ \\
    C3 information-gain (flagship) & $\mathbf{189}$ & $\mathbf{4}$ & $\mathbf{+185}$ & $46$ & $143$ \\
    \bottomrule
    \end{tabular}}
    \caption{Per-question diff against the A1 seed baseline on the training pool (300 questions), by arm. Net is Gained minus Regressed. Gain via target/transfer partitions each arm's Gained count by whether that question was ever one of the arm's own Phase-0 targets across its 100 iterations.}
    \label{tab:manifold_diff}
\end{table}

\section{Prompts Used}
\label{sec:app-prompts}

\cocoevolve{} uses Cortex Code as its mutator, alongside a reflector and proposer. The mutator edits the candidate in its workspace. The reflector chooses \texttt{keep}, \texttt{new}, or \texttt{none} from the approach ledger; \texttt{new} invokes the proposer to name a strategy. The excerpts below show the instructions relevant to these roles.

The math, ADRS, and Signal Processing mutators share the same generic interface: an injected ledger, on-demand \texttt{evo.py} history queries, and a task workspace for editing and execution.
DABStep replaces the generic system message with a Cortex-agent builder prompt that adds warehouse tools. The data engineering prompt defines the scaffolding of the context-curation pipeline that CoCoEvolve improves.

\begin{promptbox}[Mutator user (math / ADRS)]
Richer historical evidence is available on demand through a CLI:
  python3 evo.py get_best_programs -n 5
  python3 evo.py get_lineage --id <pid>
  python3 evo.py get_program --id <pid>
  python3 evo.py get_recent_scores -n 20

The current program is already written to a file in this directory.
Edit that file directly. Do not paste code into the reply.

Before finishing, write the declared approach to APPROACH.txt:
the single word hillclimb or new.
\end{promptbox}

\begin{promptbox}[Mutator system (DABStep workspace)]
You are an expert Snowflake Cortex Agent builder. You evolve agent
blueprints to improve accuracy on payment processing questions.

Diagnosis priority: test before you diagnose. Run the existing tool
the failing question would use (query, test_agent, or analyst_query).
If it returns the right result, the fix is routing; if not, the fix
is the tool itself.

Live investigation tools (python3 evo.py <subcommand>):
  query        -- read-only SQL against live tables
  test_agent   -- run a scratch clone of the current agent
  check_deploy -- compile-check a candidate DDL
  execute_sql  -- CREATE/CALL against the warehouse
  test_draft   -- deploy your edited program.json on question Q
\end{promptbox}

\begin{promptbox}[Mutator system (Data Engineering)]
You evolve the **KG construction pipeline** for data-eng-bench tasks.
    ## What a Knowledge Graph is (and why it matters)

    Each data-eng-bench task is a small analytics engineering challenge: given a
    database and an `instruction.md`, an agent must write dbt models that pass
    pytest. Before that coding step runs, **step 1** produces a **Knowledge Graph**
    — a structured markdown file (`KG.md`, ≤300 lines) that acts as a briefing
    document for the coder. A KG that only restates `instruction.md` is
    useless. A good KG is warehouse-grounded and captures:

    - What the task asks for (outputs, metrics, grain)
    - Which schemas/tables/columns exist and how they join
    - Model layering (staging → intermediate → mart), macros, and business rules
    - Edge cases, validation checks, and pitfalls to avoid

    A separate **Harbor coding agent** (fixed, not under your control) reads
    `KG.md` and writes dbt SQL. Before scoring, it may iterate against
    **KG-derived self-tests** (`test_kg.py` compiled from that KG after step 1).
    Those self-tests are NOT the grader — the hidden verifier is unchanged.
    **Your** job is to design **one** KG factory reused on every data-eng-bench
    task — the pipeline that *creates* that KG (and the lean
    `kg-adherence-test-author` protocol that shapes how tests are derived) via skills and agent orchestration descriptions.
    Better architecture → richer KGs + granular self-tests → higher pytest pass rate.

    ## The pipeline explained

    Your pipeline **explores the task environment and
    produces `KG.md`, which is handed off to a different, fixed agent that solves
    the task.** That is the only fixed contract. Everything else — how many
    agents do the exploring/writing, and **what reusable skills the KG factory
    follows — is yours to design.** Harbor step 2 receives **only** `KG.md`, not
    catalog or runtime skill files. The skill catalog **SHIPS with three
    section-skills** (`kg-agent-usage-guide`, `kg-metrics-and-business-rules`,
    `kg-structure`). A fourth catalog key, **kg-adherence-test-author**, is
    compile-time only (lean protocol for deriving granular test_kg.py from each
    KG — not a KG.md section). Your job
    is to **EDIT or GROW the existing pipeline**: author NEW skills and DIVIDE THE
    LABOR across agents or REFINE/IMPROVE the existing skills. Refine
    kg-adherence-test-author when traces show tests are too generic or Harbor
    ignores the self-test loop. Drop any skill that stops earning its place.

    ## Two axes of architecture improvement

    **Axis 1 — Agent topology** (who runs where, doing what)
    - Edit `agents[]`: an ordered list of sequential agent sessions.
    - The seed is ONE `architect` that explores and writes `KG.md` in the same
      session (query results stay in context). This is a STARTING point.
      Do not strip the warehouse look — a factory that writes from the
      instruction alone produces a useless briefing.
    - Add, split, merge, or reorder agents freely. Split into
      explorer → architect only if `explore_coverage` shows the look was skipped
      and a separate look step would help — a handoff file loses raw tool output.
    - Write each agent's `prompt` (explore the DB, synthesize findings, verify
      facts, render the final KG, etc.).
    - Attach proven skills to specific pipeline steps via `agents[].skills`.
      Skills guide KG exploration and authoring inside the factory; Harbor step 2
      receives **only** the generated KG (not catalog or runtime skill files).
    Agents share one workspace. Prefer look+write in the same session so
    `sql_execute` results stay in context — a handoff file loses raw tool output.

    **Axis 2 — Skills** (what reusable protocols the team needs)
    - Add, edit, or remove keys in `skills{}`: `{description}`. Attach each skill to a pipeline step via `agents[].skills`.
    - The assigned agent writes the full skill this eval →
    `runtime_skills/<name>.md`. You never paste markdown into the spec.

    ## Evidence (how to improve either axis)

    Do not guess topology or skills. Justify an Axis 1 or Axis 2 edit with a
    recurring gap in the parent eval: Harbor traces
    (`agent_trace.txt`), verifier stdout (`verifier_output_raw.txt`),
    and/or `explore_coverage`.

    ## Full system (three platform steps)

    Step 1 — KG pipeline (you design) → `KG.md`
    Step 1b — Platform compiler (fixed) → `test_kg.py` from KG + `kg-adherence-test-author`
    Step 2 — Harbor agent (fixed) → dbt models + self-test iteration → pytest score

    ## Your artifact: `_PIPELINE_SPEC`

    Edit between `# EVOLVE-BLOCK-START` and `# EVOLVE-BLOCK-END`:

      agents — ordered steps: {name, prompt, skills[]}
      skills — catalog of {description} per key (justification optional; assigned agent writes .md)

    ## Constraints

    - Valid Python dict; generic templates only (no task-specific table/column names).
    - Agent prompts keep `{task_name}`, `{db_name}`, `{timeout_sec}` placeholders.
    - Final KG ≤300 lines; no solution SQL in the KG.
    - Never put markdown skill bodies in `skills{}` — metadata only.
    
\end{promptbox}

\begin{promptbox}[Reflector system]
You are a strategy-pool controller for an evolutionary
code-optimization run. One worker slot has just freed up and you
decide ONE thing for it: feed it into an existing active strategy,
spawn a brand-new strategy (a proposer agent will design it), or
run plain hillclimbing with no named strategy. You do NOT design
the new strategy yourself and you do NOT edit code.
\end{promptbox}

\begin{promptbox}[Proposer system]
You are an expert algorithm researcher and strategic advisor.
Suggest breakthrough algorithmic IDEAS in a JSON object with an
'ideas' array. Each idea describes a high-level strategy (not code).
Focus on ideas that are fundamentally different from what has been
tried. You MUST always return at least one idea.
\end{promptbox}

\section{Agent Optimization}
\label{sec:app-agent}

\subsection{Detailed Methodology}
\label{sec:app-agent-method}

DABStep optimization uses the InfoGain branch of Algorithm~\ref{alg:cocoevolve} over a 300-question training pool, with a disjoint held-out test split reported in Table~\ref{tab:dabstep_test}.
The run behind Table~\ref{tab:dabstep_combined} uses a 100-iteration budget, one island, population 40, cascade threshold $[0.005]$, a task workspace, \texttt{evo.py} lineage tools, and the warehouse overlay of Appendix~\ref{sec:app-tools}.
The seed $\prog_0$ is a Cortex Agent blueprint with instructions, \texttt{payment\_analytics} text-to-SQL, \texttt{docs\_search}, and a single-transaction \texttt{fee\_calculator}. The mutator may rewrite the instructions, tool schemas, stored procedures, and dynamic tables.
GEPA receives the same specification but optimizes only its instruction text. A post-hoc full-pool evaluation of its champion scores $142/300$ on train and $59/150$ on test (Table~\ref{tab:dabstep_test}).

Meta-Harness \citep{lee2025metaharness} receives the same seed and 100-iteration budget through a faithful port of the Terminal-Bench-2 loop. The proposer prompt contains only a file pointer and receives no curated history or pre-injected per-question signal, matching the original design. We make two changes to the port. Cortex replaces the Anthropic API as the proposer backend, and the pooled-parallel DABStep evaluator replaces Harbor sandboxes. Each child is a Cortex Agent JSON specification scored on all 300 training questions.

\begin{figure}[ht]
    \centering
    \includegraphics[width=\textwidth]{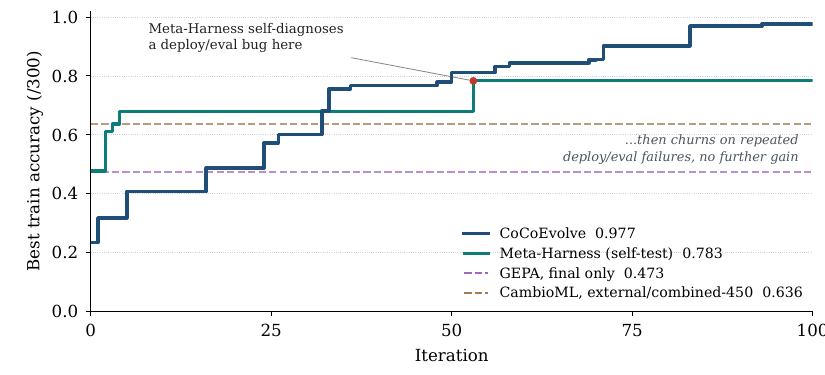}
    \caption{DABStep best-so-far train accuracy ($/300$) versus iteration. \cocoevolve{} and Meta-Harness (self-test) are running-best curves from their evolution logs. GEPA has no per-iteration full-pool trace because its search evaluates small subsamples, sometimes only one question per step, so its final post-hoc score appears as a dashed reference line. The second dashed line shows CambioML energent.ai DS Agent, an external combined-pool leaderboard submission without a matched budget or train-300 trajectory. Its easy and hard accuracies are $94.44\%$ and $57.67\%$, respectively (Table~\ref{tab:dabstep_combined}). Appendix~\ref{sec:app-agent-error-analysis} analyzes Meta-Harness's post-iteration-53 stall.}
    \label{fig:dabstep-search}
\end{figure}

The main-paper Meta-Harness row uses the self-test variant. Its proposer receives \texttt{test\_draft}, \texttt{test\_agent}, and \texttt{check\_deploy}, which provide the same self-validation surface as \cocoevolve{}'s A3 arm (Appendix~\ref{sec:app-ablat-mutation}). A held-out re-evaluation of the champion produces $80.0\%$ easy accuracy ($20/25$), $78.4\%$ hard accuracy ($98/125$), and $78.7\%$ total accuracy ($118/150$), with no evaluator errors. These are the values in Table~\ref{tab:dabstep_test}. Figure~\ref{fig:dabstep-search} plots best-so-far train accuracy against iteration for all three methods.

The remaining two rows are no-search baselines, each scored once on all 450 questions with the same evaluator and scorer as the search methods. The Snowflake Cortex Code Agent row scores the unmodified seed; the Stock Cortex Agent row scores a general-purpose coding agent without the seed's specification.

\textbf{Snowflake Cortex Code Agent} redeploys the unmodified seed $\prog_0$. Its instructions, \texttt{payment\_analytics} text-to-SQL tool, \texttt{docs\_search}, and \texttt{fee\_calculator} remain unchanged. This row measures what the seed already solves before optimization.

\textbf{Stock Cortex Agent} removes the Cortex Agent product entirely. A single Cortex Code CLI call using \texttt{claude-sonnet-5} answers each question in a fresh session without memory or an evolutionary loop. The model receives the DABStep documentation and read-only SQL access to the live tables. It does not receive a semantic view, text-to-SQL tool, fee-calculation procedure, or the seed's domain-specific instructions. Together, these two rows bound performance without search: the first retains the seed specification, while the second retains only a general-purpose coding agent with raw data access.

\subsection{Combined-suite leaderboard}

Table~\ref{tab:dabstep_combined} places our combined-pool evaluations beside public DABStep leaderboard entries \citep{egg2025dabstep} and the context-engineered configuration of \citet{dohmen2026contextlayer}. Our \cocoevolve{}, Meta-Harness \citep{lee2025metaharness}, GEPA \citep{agrawal2026gepa}, and no-search rows are evaluated on the 300 training and 150 test questions together; the Meta-Harness and GEPA rows use the champions from Table~\ref{tab:dabstep_test}. Public leaderboard submissions report combined-suite accuracy without a separate train/test breakdown. The table is ranked by hard-question accuracy.

\begin{table}[htbp]
    \centering
    \begin{tabular}{l cc}
    \toprule
    Method & Easy (\%) & Hard (\%) \\
    \midrule
    \cocoevolve{}                                       & 87.50 & \textbf{95.24} \\
    MotherDuck, full-context + data modeling~\citep{dohmen2026contextlayer} & 93.10 & 93.20 \\
    NVIDIA KGMON (NeMo Agent Toolkit) Data Explorer     & 87.50 & 89.95 \\
    DataPilot                                           & 86.11 & 87.57 \\
    Meta-Harness                                        & 84.72 & 77.25 \\
    gg-agent-gpt5-1104-1                                & 88.89 & 62.96 \\
    CambioML energent.ai DS Agent                       & \textbf{94.44} & 57.67 \\
    GEPA                                                & 88.89 & 36.24 \\
    Snowflake Cortex Code Agent                         & 75.00 & 24.87 \\
    Stock Cortex Agent (baseline)                       & 56.94 & 11.90 \\
    \bottomrule
    \end{tabular}
    \caption{Accuracy on the combined 450-question DABStep suite, ranked by hard-split accuracy.}
    \label{tab:dabstep_combined}
\end{table}

\subsection{Why self-test tools did not prevent Meta-Harness's churn}
\label{sec:app-agent-error-analysis}

Figure~\ref{fig:dabstep-search} shows Meta-Harness's self-test variant landing its one real gain at iteration $53$ and then failing to improve for the remaining $47$ iterations despite holding the same \texttt{test\_draft}/\texttt{test\_agent}/\texttt{check\_deploy} self-validation surface throughout. We inspected the archived per-iteration tool-call transcripts to understand the stall. They show two recurring problems: repeated searches for nonexistent question tables and semantic-view edits that pass the preliminary check but fail at deployment. These traces explain failures in this run; they do not establish that either problem would persist under every use of the self-test tools.

\paragraph{The query tool works; what recurs is a dead end nobody flagged, relearned every time.} Meta-Harness can query the real warehouse schema successfully. At iteration $61$, its proposer runs \texttt{DESCRIBE TABLE DABSTEP\_PAYMENTS} and receives a valid column list. In the same iteration, it reads \texttt{DABSTEP\_FEES} and \texttt{DABSTEP\_TXN\_FEE\_FACT} correctly.

The recurring failure is different. On each occurrence, the proposer spends nine calls looking for \texttt{DABSTEP\_QUESTIONS}, \texttt{DABSTEP\_EVAL\_QUESTIONS}, and \texttt{DABSTEP\_EVAL}. These tables do not exist because the question and answer bank is stored only in \texttt{frontier\_val.json} and \texttt{evolution\_summary.jsonl}. The prompt tells the proposer to read these files for per-question pass rates, but it never states that they are the only source of question-level information.

Both harnesses start from the same \texttt{initial\_program.json}, \texttt{agent\_spec}, and \texttt{referenced\_tables}. The difference is therefore not the tool or seed. It is what each prompt surfaces on every turn and whether a discovered fact persists.

\paragraph{The self-test tool cannot see the specific bug class that recurs.} The DDL failure is structural. A Cortex Analyst semantic view contains two representations that must remain synchronized: an outer SQL \texttt{tables()}/\texttt{relationships()} clause and an embedded Cortex Analyst extension JSON object. The JSON separately lists tables, facts, and dimensions.

At iteration $61$, the proposer edits both representations with ad hoc Python string operations. It inserts a relationship into the outer clause and separately patches the JSON table list instead of generating both from one source of truth. The two representations can therefore drift apart. In this case, a table appears in \texttt{tables()} but not in the JSON.

The resulting text is syntactically well formed, so \texttt{check\_deploy} cannot reject it in advance. That tool checks whether the DDL compiles but does not perform a data-flow check (\texttt{mh\_tools.py}). It can catch syntax failures, as it did for an unescaped quote in an earlier iteration, but it cannot validate cross-references between the two representations. Only the live Snowflake compiler exposes that error, and it returns only a line and position rather than a diff.

The observed failures point to distinct remedies. A prompt that identifies where question-level data lives and carries learned dead ends forward could reduce repeated fruitless queries. Structural validation of the semantic view's two representations could catch the DDL mismatch before deployment. In this run, self-test access alone did not supply either safeguard.

\subsection{Examples of Best Programs}
\label{sec:app-agent-programs}

DABStep's artifact is a Cortex Agent spec \citep{snowflake2026cortexagents}. The champion reached an in-run train estimate of $293/300$ (\texttt{combined\_score} $0.977$). A fresh full-pool re-evaluation of the same artifact scored $286/300$, as reported in Table~\ref{tab:manifold_ablation}. While the mutator discovered a suite of optimizations across the specification (including refined verified queries, precise orchestration instructions for disambiguation, and injected domain context), we highlight the architectural shift we believe contributed most to its performance: demoting the text-to-SQL surface in favor of a deterministic compute kernel. By routing fee, ranking, and glossary questions around the LLM entirely, it avoids freehand SQL errors. The champion agent achieved this by rewriting the custom tool from a 5KB SQL procedure to a 53KB Python engine with 14 modes, backed by NumPy, SciPy, and NetworkX. The agent can therefore answer complex questions with a single tool call instead of summing many single-row fees in context, directly addressing the failure mode discussed in Section~\ref{sec:experiments}.

The deployed agent reads the following abridged specification, which heavily guides it to trust its new modes over its text-to-SQL capabilities. Key routing instructions and tool descriptions added by the mutator are highlighted in blue:

\begin{promptbox}[Evolved specification (champion, abridged)]
{
  "agent_spec": {
    "instructions": {
      "response": "Provide precise, data-driven answers. Always show the exact numbers from query results. <abridged...>",
      "orchestration": "You are a Payment Operations Analyst agent specialized in the Dabstep payment processing dataset. <abridged...>
      @@## Tool Selection Rules
### New fee_calculator modes for residual question families
Route these DIRECTLY to fee_calculator instead of freehand SQL via payment_analytics:
- "For [account type X and/or MCC description: Y...] what would be the average fee..." -> mode='avg_fee_criteria'
- "Which fee ID(s) apply to <criteria>" with no merchant scope -> mode='fee_ids_criteria'
- "How many unique/distinct <column> are there" -> mode='distinct_count'
<abridged...>

## Fee Rule Engine (fee_calculator tool) -- generalized modes
The fee_calculator tool is a generalized vectorized fee-matching engine over a merchant's real transactions (it sums ALL matching fee rules per transaction, then aggregates). Route these question categories to it directly instead of hand-writing SQL: <abridged...>
Trust this tool's numbers over manually derived SQL for these categories -- it already handles monthly fraud/volume bracket computation, capture_delay bucketing, and the correct intracountry definition. <abridged...>@@"
    },
    "tools": [
      {
        "tool_spec": {
          "type": "cortex_analyst_text_to_sql",
          "name": "payment_analytics",
          "description": "Use this tool to query payment transaction data, merchant information, fee rules, and acquirer details. <abridged...>"
        }
      },
      {
        "tool_spec": {
          "type": "cortex_search",
          "name": "docs_search",
          "description": "Use this tool to search payment processing domain documentation including the merchant guide, fee optimization strategies, fraud prevention best practices. <abridged...>"
        }
      },
      {
        "tool_spec": {
          "type": "generic",
          "name": "fee_calculator",
          "description": "@@Generalized vectorized fee-rule engine for a merchant's transactions. Use this tool (NOT freehand SQL) for ANY question about: (1) which card scheme a merchant should steer traffic to for minimum/maximum fees -- mode='steer'; (2) total fees a merchant paid/should pay over a year/month/day -- mode='total'; <abridged...>@@",
          "input_schema": {
            "type": "object",
            "properties": {
              "p_mode": {
                "description": "One of: 'steer' (best/worst card scheme for min/max fees), 'steer_aci' (best/worst ACI to move a merchant's fraudulent transactions towards for min/max fees... <abridged...>",
                "type": "string"
              }
              <abridged...>
            }
          }
        }
      }
    ]
  }
}
\end{promptbox}

A warehouse stored procedure backs the \texttt{fee\_calculator} tool. While we omit the full 53KB Python body, the evolution trajectory reveals several striking mechanistic and semantic discoveries that the agent authored and verified via workspace execution:

\begin{itemize}
    \item \textbf{Vectorized tensor contraction.} The initial tool used SQL to evaluate 9-field conditional fee rules. The champion replaces this with a sparse boolean tensor contraction in Python. It precomputes a boolean ``allowed'' matrix for each field where wildcards become an all-True row, combines them via bitwise \texttt{\&}, and computes the final fees using sparse matrix-vector products (\texttt{scipy.sparse.csr\_matrix}).
    \item \textbf{Semantic bug fixes.} The mutator fixed two silent bugs in the human-authored seed. First, it corrected a whitespace bug in a string label (\texttt{'> 5'} to \texttt{'>5'}) that had silently disabled 92 fee rules for 6 merchants. Second, it redefined ``intracountry'' transactions from \texttt{IP\_COUNTRY = ISSUING\_COUNTRY} to \texttt{ISSUING\_COUNTRY = ACQUIRER\_COUNTRY}, a definitional shift that alters the evaluation of 70\% of the 138,236 rows in the dataset.
    \item \textbf{Emergent domain semantics.} The agent discovered through trial and error that fees \emph{stack}. Its code comments explicitly warn its future self: ``because fees STACK... Do NOT take the MAX of a single matching rule... that produces a different (and wrong) ranking.''
    \item \textbf{Graph-based join resolution.} Rather than letting the orchestrator guess join keys for text-to-SQL, the tool hardcodes the schema's foreign-key graph and uses \texttt{networkx.shortest\_path} to emit deterministic \texttt{JOIN ... ON} fragments, returning an explicit refusal if no path exists.
    \item \textbf{Regex over semantic search.} For exact-definition questions, the agent deliberately replaces vector search with a regex and \texttt{difflib} parser over the documentation, reasoning that embeddings might return a topically-related chunk that omits the exact definition.
\end{itemize}

Beyond the champion tool shown above, other evolutionary lineages explored alternative structural solutions. The most notable alternatives materialized dynamic tables to pre-compute frequently-joined columns and eliminate flaky on-the-fly joins and produced detailed sets of verified queries encoding domain-specific reasoning paths. While these alternative approaches were interesting, they ultimately lost to the champion's centralized Python compute kernel. Across all lineages, the agent also continually refined its orchestration guidelines, encoding lessons from observed failures into tight situational guardrails.
\section{AI Functions}
\label{sec:app-aifn}

\subsection{Synthetic Diagnostic: Lockpicking with Interference}
\label{sec:app-lockpicking}

\paragraph{Construction and protocol.}
This diagnostic isolates recovery of interacting rules from the semantic
uncertainty of the natural-data tasks.  Each example is a six-digit string
$P_1\ldots P_6$ with an integer label in $\{0,\ldots,6\}$.  A position first
passes a position-specific candidate-set check; twelve ordered trigger rules
(24 trigger--target edges) can then invalidate otherwise-valid positions.
The label is the number of positions that remain valid.  For example,
\[
P_1 \in \{1,4,7\}.
\]
Two rules that can subsequently invalidate $P_1$ are
\[
P_2=8 \Longrightarrow P_1,P_3\ \mathrm{invalid},
\qquad
P_3=3 \Longrightarrow P_1,P_2\ \mathrm{invalid}.
\]
The seed provides all six candidate sets but only one of the twelve
interference rules.  Every method receives the same 50 labeled strings and 50
mutation iterations.  There is no held-out partition: train and test are the
same fixed set, because the measured quantity is exact reconstruction of the
latent rule system rather than distributional generalization.

\paragraph{Why aggregate accuracy is underdetermined.}
The evaluator returns only the final count, not which positions were
invalidated.  In the example above, if both triggers fire, omitting either edge
into $P_1$ need not change that row's label.  A trigger can likewise be masked
when its target was already rejected by the candidate-set check or an earlier
rule.  Consequently, many structurally different rule tables agree on most of
the 50 examples.  This property makes the diagnostic useful: a locally
successful patch can raise accuracy without identifying the rule responsible
for the improvement.

\begin{table}[htbp]
    \centering
    \begin{tabular}{l cccc}
    \toprule
    Method & GEPA & OpenEvolve & AdaEvolve & \cocoevolve{} \\
    \midrule
    Correct / 50 & 41 & 27 & 45 & \textbf{50} \\
    Accuracy & 0.82 & 0.54 & 0.90 & \textbf{1.00} \\
    \bottomrule
    \end{tabular}
    \caption{Accuracy on \emph{Lockpicking with Interference} after 50 iterations.}
    \label{tab:lockpicking_motivation}
\end{table}

\paragraph{Recovered search behavior.}
\cocoevolve{} reaches $50/50$ after six mutations by contrasting multiple
failures and compiling all twelve recovered rules into deterministic SQL.
OpenEvolve ends at $27/50$ with unresolved contradictions in its own worked
examples.  AdaEvolve reaches $45/50$ at iteration 34 and then plateaus: only
two of its twelve learned trigger rows have the correct target pair, five
reuse the same \texttt{[P3,P5]} shortcut, and none can invalidate $P_1$.
The gap is therefore structural rather than evaluation noise: the baselines
fit surrogate rules that score well on the fixed set, whereas \cocoevolve{}
recovers the latent rule system.

\subsection{Detailed Methodology}
\label{sec:app-aifn-method}

\paragraph{Program representation.}
Each candidate is an executable Snowflake SQL expression that forms the body of
a scalar AI function.  The seed for every task contains one
\texttt{AI\_COMPLETE} call, a task instruction, a row-dependent user message,
and a structured-output schema.  Evolution may rewrite the instruction and
schema descriptions, introduce CTEs or deterministic SQL post-processing, and
compose multiple model calls, but it must preserve the function signature,
output field and type, and final accessor.  Thus the search space includes both
prompt optimization and pipeline optimization while every candidate remains
directly executable by the target query engine.  The seed is deliberately
minimal: for example, the \texttt{go\_emotions} seed asks for every emotion that
``clearly applies,'' and the \texttt{multilexnorm} seed gives only four
normalization examples.

\paragraph{Tasks and metrics.}
The five live-data tasks use the following row-level objectives:
\texttt{go\_emotions} uses example-level multi-label F1;
\texttt{finqa} and \texttt{convfinqa} use execution accuracy after canonicalizing
the returned numeric answer;
\texttt{lex\_glue} uses exact match over thirteen Supreme Court issue areas; and
\texttt{multilexnorm} uses exact string match, including case, punctuation, and
spacing.  The synthetic \texttt{lockpicking\_6} diagnostic uses exact match on
the integer score.  A malformed SQL body, invalid structured output, or
warehouse execution failure receives score zero together with the compilation
or execution diagnostic.

\paragraph{Data partitions.}
For each live-data task, the search pool is partitioned once with seed 42.
Two thirds form a validation split used only for fitness and acceptance, while
the remaining third supplies row contents and scorer feedback to the mutation
agent.  The disjoint test table is never exposed during evolution; after the
50-iteration run, the exported validation champion is re-executed on that test
table to obtain Table~\ref{tab:ai_function_results}.  This separation is
important because the mutator can inspect concrete training failures.
\texttt{lockpicking\_6} is intentionally different: all examples are visible
and scored because its purpose is to test whether search can reconstruct a
hidden rule system rather than generalize to an unseen natural-data split.

\paragraph{Mutation protocol.}
The AI-function runs instantiate the general loop in
Appendix~\ref{sec:app-loop}, the history interface in
Appendix~\ref{sec:app-tools}, and the $\lp$ roster in
Appendix~\ref{sec:app-lp}.  Their task-specific specialization is a
fresh-session curator--mutator split: the curator inspects the roster, ledger,
and training-only failure history to select a parent, an optional merge
partner, and up to ten representative rows; the mutator receives only that
bounded packet and may verify its SQL on those rows up to three times.  The
subsequent full-validation evaluation is the sole admission gate.  Runs use 50
mutation attempts, one island, population 40, archive capacity 20, and eight
parallel workers; similarity-informed targeting is disabled because every
child is evaluated on the complete validation split.

\paragraph{Population objective.}
AI functions use the continuous $p=2$ instantiation of
Appendix~\ref{sec:app-lp}, with a roster budget of ten.  Marginals are sums of
row-level score improvements rather than binary solved-row counts; all other
roster construction, consolidation, and parent-selection details follow that
section.

\paragraph{Matched baselines and score provenance.}
OpenEvolve and AdaEvolve start from the same task-specific SQL body and receive
50 mutation attempts.  Their validation champions are scored post hoc on the
same held-out test partition.  GEPA likewise optimizes the SQL body against the
validation partition; Table~\ref{tab:ai_function_results} uses the best
completed budget for each task.  For the five live-data tasks, the paper values
are held-out re-evaluations rather than the optimizer's internal archive
fitness.  For lockpicking, all methods are scored on the same fixed visible
set, consistent with the diagnostic protocol above.

\subsection{Examples of Best Programs}
\label{sec:app-aifn-programs}

The evolved programs remain ordinary SQL bodies, but they are substantially
more operational than their seeds.  Across tasks, evolution turns a generic
instruction into an explicit decision procedure: ordered rules for
classification, reference and sign conventions for numerical reasoning, or
token-preservation contracts for normalization.  We give representative
examples below, reproducing the two lockpicking champions in full and
abridging the longer live-task programs.  For the live tasks, the base boxes
show the task-bearing seed instruction; we omit repeated label enumerations,
the \texttt{AI\_COMPLETE} wrapper, and the unchanged structured-output schema.

\paragraph{\texttt{lockpicking\_6}: exact rules versus a surrogate ruleset.}
The two champion artifacts make the structural difference directly
inspectable.  The \cocoevolve{} listing is the complete deterministic scoring
expression produced by its winning child.  In the deployed AI-function body,
the resulting integer is passed to \texttt{AI\_COMPLETE} only for
schema-conforming serialization.

\begin{promptbox}[\texttt{lockpicking\_6} base program (instruction)]
Score a 6-digit program string.

If the input is not exactly six digits, return 0.
Valid digits:
  P1: {1,4,7}  P2: {2,6,8}  P3: {0,3,9}
  P4: {1,4,7}  P5: {0,5,8}  P6: {2,5,9}

Apply interference rules in order:
  1. IF P1 = 4 -> invalidate P2, P6

Return the number of positions still valid after both checks.
\end{promptbox}

\programlisting{\cocoevolve{} champion: deterministic scoring expression}{listings/lockpicking-cocoevolve.sql}

The AdaEvolve listing is its complete $45/50$ champion prompt, including the
ten worked examples accumulated during search.  We normalize typographic
dashes and operators to ASCII for typesetting; the wording and rules are
otherwise unchanged.

\programlisting{AdaEvolve champion: evolved executor prompt}{listings/lockpicking-adaevolve.txt}

\paragraph{\texttt{go\_emotions}: conservative labeling and ordered rules.}
\begin{promptbox}[\texttt{go\_emotions} base program (instruction)]
Identify every emotion expressed in the text. A comment may express
more than one emotion. Return an array drawn from the declared
28-label vocabulary. Include every category that clearly applies;
use ["neutral"] only when no other emotion applies.
\end{promptbox}

The seed's instruction to include every plausible emotion caused systematic
over-labeling.  At iteration 7, the curator queried the seed's full failure
set, inspected representative rows, and then used question memory to
contextualize partial scores.  For ``Happy Cake day!'' the gold label was
\texttt{joy} but the parent received $0.5$; for ``This is my favorite wild
memory'' it received $0.4$.  These scores exposed a common mechanism: the
parent was returning the correct emotion together with one or more unnecessary
labels.  The curator prescribed a conservative single-label default.  The
mutator's first draft passed eight of ten selected rows; after adding rules for
sarcastic positive language and social check-ins, the same live verifier
reported ten of ten.  The committed child increased the search objective from
$0.0022$ to $0.1741$ in one iteration.

The final champion generalizes this idea into a strict-priority rule system:
\begin{promptbox}[\texttt{go\_emotions} champion (abridged)]
Classify emotions using the GoEmotions annotation scheme.

ALLOWED LABELS --- use only the 28 labels in the declared vocabulary.
Apply the following rules in strict priority order. Record which rule
matched, then output only that rule's prescribed emotion list.

RULE A --- GRATITUDE EXPRESSION:
If the text starts or ends with "thanks" or "thank you",
output ["gratitude"] only.

RULE B --- EXPLICIT FIRST-PERSON EMOTION:
Check explicit worry, fear, love, sadness, joy, curiosity, pride,
and relief patterns in order; stop at the first match.

RULE C --- INTERPERSONAL STATE QUESTION:
For a direct question about another person's wellbeing,
output ["caring", "curiosity"].

...

DEFAULT:
Return the single most prominent emotion unless two emotions are
independently and explicitly expressed.
\end{promptbox}
The champion was discovered at iteration 47 and obtains the held-out score
$0.444$ reported in Table~\ref{tab:ai_function_results}.  Although the
artifact is prompt-heavy, its structure is algorithmic: ordered guards,
explicit exceptions, a closed vocabulary, and a forced structured output
replace the seed's unconstrained semantic judgment.

\paragraph{\texttt{finqa} and \texttt{convfinqa}: executable arithmetic
contracts.}
\begin{promptbox}[\texttt{finqa}/\texttt{convfinqa} base programs (instructions)]
FINQA:
Answer a numerical question over a financial report (text + table).
Work out the answer and return the final numeric value (or yes/no).
Use no units, thousands separators, or percent sign.

CONVFINQA:
Answer the current turn of a multi-turn numerical conversation over
a financial report. Use dialogue history to resolve references such
as "the difference" or "this". Return the final plain numeric value
(or yes/no), with no units, commas, or percent sign.
\end{promptbox}

The financial-QA seeds merely ask the model to work out a numeric answer.
Their champions instead define an operational contract for reference
resolution, formula selection, sign direction, units, and answer
serialization.  The \texttt{finqa} program distinguishes percent change,
signed absolute change, relative increase, same-period differences, ratios,
averages, CAGR, and yes/no questions.  The conversational champion adds rules
for resolving ``that,'' ``the difference,'' and ``the last year,'' and for
reusing prior answers without silently substituting a nearby year.

\begin{promptbox}[\texttt{finqa} champion (abridged)]
Work through each question step by step:
1. Identify the relevant numbers from the text or table.
2. Match the question's exact wording to the correct formula.
3. Compute carefully and show intermediate steps.

FORMULA SELECTION:
"percent change"       -> (new - old) / old; preserve the sign
"change in X"          -> new - old in the table's stated unit
"increase/decrease"    -> (new - old) / old as a signed ratio
"how much bigger"      -> X - Y; do not divide
"average"              -> sum / count
"ratio of X to Y"      -> X / Y
"CAGR"                 -> (end / start)^(1 / years) - 1, then x100

Return plain values only: no currency signs, commas, units, or percent
signs. Do not convert away from the table's stated unit. Sign matters.
\end{promptbox}

\begin{promptbox}[\texttt{convfinqa} champion (abridged)]
STEP 1 --- RESOLVE REFERENCES:
Resolve pronouns and shorthand from the dialogue history. If the current
question requests a value already stated for the same quantity and year,
reuse that exact value.

STEP 2 --- IDENTIFY OPERATION AND SIGNS:
"decline from A to B"      -> A - B
"change from A to B"       -> B - A
"combined total"           -> sum the referenced values
"as a percentage of"       -> component / whole

STEP 3 --- EXTRACT VALUES:
List every required value and its source. Match the requested year exactly.

STEP 4 --- COMPUTE:
Show the arithmetic explicitly, then emit a plain numeric answer
with no unit, comma, or percent sign.
\end{promptbox}
These contracts address errors that are difficult to repair with isolated
answer demonstrations: the same numbers can require a ratio, signed
difference, or magnitude depending on a short phrase.  The resulting held-out
scores are $0.813$ for \texttt{finqa} and $0.953$ for
\texttt{convfinqa}.

\paragraph{\texttt{multilexnorm}: conservative token transformation.}
\begin{promptbox}[\texttt{multilexnorm} base program (instruction)]
Normalize noisy social-media text. Rewrite each non-standard token
to its canonical form (e.g., u->you, tmrw->tomorrow,
gonna->going to, dont->don't). Leave standard words, @mentions,
#hashtags, and URLs unchanged. Return the full normalized text.
\end{promptbox}

Exact-match normalization is unusually sensitive to helpful but unauthorized
rewrites.  The evolved champion therefore begins with preservation rules,
then applies a bounded lexicon of high-confidence transformations while
preserving capitalization, spacing, punctuation, mentions, hashtags, URLs,
proper nouns, and uncertain or non-English tokens.
\begin{promptbox}[\texttt{multilexnorm} champion (abridged)]
Rewrite each non-standard English token to its canonical form,
leaving everything else exactly as written.

ALWAYS LEAVE UNCHANGED:
- @mentions, #hashtags, URLs, proper nouns, and non-English words
- already-standard informal forms such as wanna, kinda, lol, and btw
- uncertain tokens; when in doubt, leave the token unchanged
- emotional sound effects whose repeated letters encode emphasis

CAPITALIZATION:
Preserve the exact capitalization pattern of the original token.

SPACING AND PUNCTUATION:
Preserve all whitespace and punctuation spacing exactly.

HIGH-CONFIDENCE TRANSFORMS:
dont -> don't;  u -> you;  ur -> your;  pls/plz -> please;
tryn/tryna -> trying to;  thx/thnx -> thanks;  nvm -> never mind.
\end{promptbox}
The final program also records exceptions for contractions, elongated forms,
and truncated \texttt{-ing} words.  This ``transform only when confident''
policy is the opposite of the seed's broad invitation to normalize and
achieves $0.587$ on the held-out set, a $7.7$ pp gain over GEPA.

\paragraph{\texttt{lex\_glue}: hierarchy of legal exceptions.}
\begin{promptbox}[\texttt{lex\_glue} base program (instruction)]
Classify a US Supreme Court opinion by its primary SCDB issue area.
Choose exactly one of:
Criminal Procedure, Civil Rights, First Amendment, Due Process,
Privacy, Attorneys, Unions, Economic Activity, Judicial Power,
Federalism, Interstate Relations, Federal Taxation, Miscellaneous.

Use the label glossary: rights of the accused -> Criminal Procedure;
discrimination and voting -> Civil Rights; speech, press, and religion
-> First Amendment; abortion and personal autonomy -> Privacy;
fees and professional discipline -> Attorneys; labor relations ->
Unions; antitrust, contracts, torts, and business regulation ->
Economic Activity; jurisdiction, standing, mootness, and remedies ->
Judicial Power. Choose the issue area that dominates the holding.
\end{promptbox}

The legal classifier evolves from a short glossary into a hierarchy of
issue-area distinctions.  Its champion begins with a priority scan for
attorney-fee holdings, then separates the constitutional substance of a case
from its procedural posture.  For example, it distinguishes a speaker denied
access to a government forum (\texttt{First Amendment}) from a captive listener
resisting unwanted content (\texttt{Privacy}); federal-court jurisdiction over
an agency order (\texttt{Judicial Power}) from the substantive scope of an
agency's economic authority (\texttt{Economic Activity}); and federal labor
preemption (\texttt{Federalism}) from the underlying labor right
(\texttt{Unions}).  The complete prompt is long, but these repeated
``default--exception--counterexample'' clauses are precisely the structure
that the archive accumulated from observed confusions.  The champion scores
$0.737$ on held-out opinions.

\begin{promptbox}[\texttt{lex\_glue} champion (abridged)]
Choose exactly one SCDB issue area from the declared 13-label vocabulary.

FIRST --- ATTORNEY FEES PRIORITY SCAN:
If the primary holding concerns attorney-fee entitlement, computation,
recovery, or denial, classify as Attorneys immediately; the underlying
substantive dispute does not change the classification.

SPEAKER / LISTENER DISTINCTION:
A speaker denied access to a government forum is First Amendment.
A captive listener resisting compelled content is Privacy.

AGENCY CASES:
Whether a federal court may review an agency order is Judicial Power.
The substantive scope of the agency's regulatory authority is
Economic Activity (or Unions for a substantive labor right).

LABOR PREEMPTION:
Federal labor law displacing state or local regulation is Federalism;
the underlying organizing or bargaining right is Unions.
\end{promptbox}

Together, these artifacts show the same pattern at different scales:
evolution replaces broad task descriptions with explicit precedence,
exception, and serialization rules.

\section{Data Engineering}
\label{sec:app-de}
\subsection{Detailed Methodology}

Data engineering optimization applies Algorithm~\ref{alg:cocoevolve} to data-eng-bench.  We define 50 dev tasks and 53 held-out test tasks, randomly sampled from the full set of 103 tasks. 

\paragraph{Knowledge graphs.}
A primary output of an evolved pipeline is a \emph{knowledge graph} (KG), produced
before the downstream agent attempts solving a task. The KG is a context-grounded markdown briefing, \texttt{KG.md}, limited to 300 lines. Evolved KGs describe schemas, tables, columns, dbt models, outputs, and metrics. They also record joins, cardinalities, \texttt{source()}/\texttt{ref()} lineage, model dependencies, derivations, model grain, business rules, boundary behavior, null handling, fanout risks, validation checks, and task-specific pitfalls.

The KG is not an RDF store and contains no solution SQL. The authoring pipeline verifies physical table locations, column types, categorical values, null encodings, and row counts using \texttt{SHOW}, \texttt{DESCRIBE}, \texttt{INFORMATION\_SCHEMA}, and \texttt{sql\_execute}. This avoids relying on a potentially stale \texttt{\_sources.yml} file.

\paragraph{The evolved program.}
The evolved artifact is the Python dictionary
\texttt{\_PIPELINE\_SPEC} in \texttt{initial\_program.py}. It jointly controls
the KG authoring protocols, agent orchestration, and adherence test authoring
protocol. Executing this specification produces the task-specific
\texttt{KG.md} and \texttt{test\_kg.py} supplied to the downstream coding
agent. The mutator can improve this pipeline through the following three
components.

\begin{enumerate}
    \item \textbf{KG authoring protocols.}
    The \texttt{skills\{\}} catalog describes reusable procedures for
    constructing different parts of the KG, including its usage guide, metrics
    and business rules, and warehouse and model structure. Catalog entries
    contain metadata rather than complete skill bodies. During execution,
    assigned agents expand them into
    \texttt{runtime\_skills/<name>.md}. These skills guide KG production and
    may be shared among agents in the authoring pipeline.

    \item \textbf{Agent orchestration.}
    The \texttt{agents[]} list defines an ordered sequence of agent
    sessions that produce the KG. Each session has an editable prompt and a set of assigned skills. The mutator may rewrite prompts, add or remove agents, split or merge roles, reorder the pipeline, and redistribute skills. The agents share a sanitized workspace. This permits either a single agent or a pipeline with separate exploration, verification, and synthesis roles.

    \item \textbf{Adherence test protocol.}
    A second form of context for the solver agent is
    executable tests derived from the KG. It is a protocol used during
    compilation rather than a section of \texttt{KG.md}. A fixed test authoring
    model consumes the KG together with this evolved protocol and produces
    \texttt{test\_kg.py}. Evolution can therefore improve both the declarative
    guidance given to the solver and the way that guidance is translated into
    executable feedback.
\end{enumerate}

\paragraph{Inference and evaluation.}
At inference time, the best evolved program is fixed and its executor
materializes the evolved protocols and agent configuration. All evaluated agent roles use the same out-of-the-box \texttt{claude-sonnet-5}
backbone and harness. These roles include the ordered agent sessions
that construct the KG and the downstream agent that consumes it
before executing a task. Only \texttt{KG.md} is registered as a skill
for the Harbor agent. The internal \texttt{runtime\_skills/*.md} files are not
forwarded.

The sole model exception is the fixed adherence test author, which
uses \texttt{claude-opus-4-8} with temperature $0.2$ and a completion limit of 50,000
tokens. It translates \texttt{KG.md} and the evolved adherence test authoring
protocol into \texttt{test\_kg.py}. We hard pin this model because \texttt{claude-sonnet-5} frequently produced truncated or syntactically invalid long test
suites, particularly unterminated strings. The generated suite is supplied to
the Harbor agent as self-correction feedback. It is not the hidden grader,
which remains fixed throughout evaluation.

We note that adherence tests were generated for these evaluation runs but were not used for all runs. We therefore treat adherence testing as optional additional context rather than a uniform component of the reported results.

\paragraph{Search methods.}
The reported \cocoevolve{} result uses the best program available after 200
iterations. We configure one island, an active population of 15 candidates, an
archive capacity of 8, 8 parallel evaluations, and the
P0/P1/P2 evolution graph over the dev pool. We compare against faithful
ports of GEPA, Meta-Harness, and AdaEvolve using the same \texttt{claude-sonnet-5}
backbone, downstream agent, equivalent iteration budget, and seed program \texttt{\_PIPELINE\_SPEC}.

\subsection{Examples of Best Programs}
\subsubsection{Evolved Programs}
\label{sec:app-de-programs}

The evolved specification $\prog^\ast$ retains the initial three seed skills and authors an additional four. Two are pre-exploration environment checks:
\texttt{kg-environment-preflight} confirms a usable, authorized warehouse,
and \texttt{kg-dbt-profiles-default-path} pins the
\texttt{\textasciitilde/.dbt/} default so fixes are persisted rather than
session-local. The other two are a column-verification protocol requiring
\texttt{DESCRIBE} evidence for every named column, and a worked-example
protocol for weighted or ratio-based metrics. The
\texttt{kg-adherence-test-author} description also tightens, requiring
each stated business rule to be checked by live-SQL recomputation
against source and banning negative-only assertions. In this example, the initial
\texttt{architect} agent is the sole agent; no other agents were spawned.

\begin{promptbox}[Initial pipeline specification $\prog_0$]
_PIPELINE_SPEC = {
    "skills": {
        "kg-agent-usage-guide": {
            "description": (
                "Author the Agent Usage Guide: what the downstream solver must "
                "produce (object names, grain) and how to know it is done. "
                "Generic guidance only --- no task-specific table or column names."
            ),
        },
        "kg-metrics-and-business-rules": {
            "description": (
                "Author Metrics & Business Rules: record numeric thresholds, "
                "windows, and formulas from the instruction. Do not invent values."
            ),
        },
        "kg-structure": {
            "description": (
                "Author Structure: confirmed sources (live warehouse location), "
                "required models/outputs (name, grain, columns, dependencies), "
                "and joins (keys, cardinality). Verified warehouse facts only --- "
                "do not restate the instruction. Generic guidance only --- no "
                "task-specific table or column names."
            ),
        },
        # Compile-time only (platform emits test_kg.py after KG.md --- not a KG.md section).
        "kg-adherence-test-author": {
            "description": (
                "One precise [KG-ADHERENCE] test per testable KG fact; failure "
                "messages quote the KG. Ban generic smoke tests."
            ),
        },
    },
    # Seed topology: one architect (look + write in the same session so
    # query results stay in context). Split only if explore_coverage shows
    # the look was skipped and a handoff would help.
    "agents": [
        {
            "name": "architect",
            "prompt": (
                "You architect the Knowledge Graph for task {task_name} (database "
                "{db_name}). Budget: {timeout_sec:.0f}s.\n\n"
                "The instruction is a brief, not the warehouse. Do not paraphrase "
                "it into KG.md --- that briefing is useless to the solver. Looking "
                "is not optional. Use `sql_execute` in THIS session before you "
                "write; you are not done until every table you will name has been "
                "DESCRIBE (or SHOW COLUMNS). Do not stop at SHOW DATABASES. "
                "Confirm each source's actual database.schema.table via SHOW / "
                "INFORMATION_SCHEMA. Do not stop because the instruction looks "
                "complete. For categorical / boolean / dirty-typed columns that "
                "feed a filter, join, or metric, GROUP BY / COUNT stored tokens, "
                "null encoding, and row counts. Name only what those queries "
                "confirmed. Verified "
                "facts only; no solution SQL. Then write `KG.md` (<=300 lines, "
                "first line `# Knowledge Graph: {task_name}`).\n\n"
                "Structure KG.md by following EVERY skill listed under \"Skills to "
                "author\" below. Author each as `runtime_skills/<name>.md` before "
                "you finish."
            ),
            "skills": [
                "kg-agent-usage-guide",
                "kg-metrics-and-business-rules",
                "kg-structure",
            ],
        },
    ],
}
\end{promptbox}

\begin{promptbox}[Best evolved pipeline specification $\prog^\ast$]
_PIPELINE_SPEC = {
    "skills": {
        "kg-agent-usage-guide": {
            "description": (
                "Author the Agent Usage Guide: what the downstream solver must "
                "produce (object names, grain) and how to know it is done. "
                "Generic guidance only — no task-specific table or column names."
            ),
        },
        "kg-metrics-and-business-rules": {
            "description": (
                "Author Metrics & Business Rules: record numeric thresholds, "
                "windows, and formulas from the instruction. Do not invent values."
            ),
        },
        "kg-structure": {
            "description": (
                "Author Structure: confirmed sources (live warehouse location), "
                "required models/outputs (name, grain, columns, dependencies), "
                "and joins (keys, cardinality). Verified warehouse facts only — "
                "do not restate the instruction. Generic guidance only — no "
                "task-specific table or column names."
            ),
        },
        # Compile-time only (platform emits test_kg.py after KG.md — not a KG.md section).
        "kg-adherence-test-author": {
            "description": (
                "One precise [KG-ADHERENCE] test per testable KG fact; failure "
                "messages quote the KG. Ban generic smoke tests. For any stated "
                "filter/dedup/exclusion business rule — especially a KG value "
                "explicitly labeled evidence rather than a prescribed target — "
                "the assertion must be a live SQL comparison: recompute the "
                "correct aggregate from source inside the test by embedding the "
                "exact stated predicate/dedup key, then assert the model output "
                "matches that live-recomputed value within a tight numeric "
                "tolerance. Do not assert only 'output != one known-wrong value' "
                "for a business rule — a negative-only check can pass while the "
                "rule is still violated a different way; it must be paired with "
                "a positive, rule-derived comparison."
            ),
        },
        "kg-column-verification-protocol": {
            "description": (
                "Before naming any specific column in Structure or Metrics & "
                "Business Rules, that column must have an associated "
                "DESCRIBE/SHOW COLUMNS (or information_schema.columns) result "
                "from THIS session. If a column was not schema-probed, omit it "
                "or mark it explicitly unverified rather than stating it as "
                "fact. Generic guidance only — no task-specific table or "
                "column names."
            ),
        },
        "kg-environment-preflight": {
            "description": (
                "Before other exploration, verify the actual database session "
                "context the downstream solver will inherit: confirm which "
                "warehouse is active/authorized for the working role (a session "
                "may start with no active warehouse, or a default warehouse name "
                "that does not exist or is not authorized). Record the "
                "confirmed-working warehouse and the exact persistent fix (e.g. "
                "profile/project-level config or an explicit USE WAREHOUSE in the "
                "model-build path) as an Agent Usage Guide fact — not a "
                "session-local env-file edit a later step could silently revert. "
                "Generic guidance only; no task-specific table or column names, "
                "and no assumption about which warehouse name is broken vs "
                "working in a given account."
            ),
        },
        "kg-dbt-profiles-default-path": {
            "description": (
                "Author a profiles-directory fact for the Agent Usage Guide when "
                "the task requires creating a new dbt project (rather than "
                "editing a pre-existing one): dbt's CLI defaults to "
                "`~/.dbt/profiles.yml` whenever `--profiles-dir`/"
                "`DBT_PROFILES_DIR` is not explicitly set on a given invocation. "
                "If the solver only writes `profiles.yml` to a custom path and "
                "relies on exporting `DBT_PROFILES_DIR` for its own manual `dbt "
                "run` calls, any OTHER process that later calls `dbt run`/`dbt "
                "build` in a fresh shell without that env var will fail at the "
                "default location instead. State explicitly: also place (or "
                "symlink) a copy of `profiles.yml` at the OS-default dbt "
                "profiles directory, not only the custom one. Verify by checking "
                "whether `~/.dbt/` already exists and what it contains before "
                "assuming a custom path is sufficient. Generic dbt-CLI behavior "
                "— no task-specific table or column names."
            ),
        },
        "kg-composite-formula-worked-example": {
            "description": (
                "For any metric in Metrics & Business Rules that involves a "
                "weighted composite score, a percentile/rank direction "
                "(ascending vs. inverted), or a ratio-to-median/cap-and-invert "
                "normalization, do not describe the formula in prose alone. "
                "Pick 1-2 real rows from the live warehouse, compute every "
                "intermediate factor by hand following the exact stated formula "
                "(including cap/invert/boundary behavior and NULL-fallback "
                "substitution), and record the final numeric result as a worked "
                "example in the KG. If the instruction's wording is genuinely "
                "ambiguous even after this exercise, state the ambiguity "
                "explicitly as an open question rather than silently picking one "
                "interpretation and presenting it as fact. Never invent "
                "thresholds not present in the instruction. Generic guidance "
                "only — no task-specific table/column names or copied assertion "
                "values."
            ),
        },
    },
    # Seed topology: one architect (look + write in the same session so
    # query results stay in context). Split only if explore_coverage shows
    # the look was skipped and a handoff would help.
    "agents": [
        {
            "name": "architect",
            "prompt": (
                "You architect the Knowledge Graph for task {task_name} (database "
                "{db_name}). Budget: {timeout_sec:.0f}s.\n\n"
                "The instruction is a brief, not the warehouse. Do not paraphrase "
                "it into KG.md — that briefing is useless to the solver. Looking "
                "is not optional. Use `sql_execute` in THIS session before you "
                "write; you are not done until every table you will name has been "
                "DESCRIBE (or SHOW COLUMNS). Do not stop at SHOW DATABASES. "
                "Confirm each source's actual database.schema.table via SHOW / "
                "INFORMATION_SCHEMA. Do not stop because the instruction looks "
                "complete. For categorical / boolean / dirty-typed columns that "
                "feed a filter, join, or metric, GROUP BY / COUNT stored tokens, "
                "null encoding, and row counts. Name only what those queries "
                "confirmed. Verified "
                "facts only; no solution SQL. Then write `KG.md` (≤300 lines, "
                "first line `# Knowledge Graph: {task_name}`).\n\n"
                "Structure KG.md by following EVERY skill listed under \"Skills to "
                "author\" below. Author each as `runtime_skills/<name>.md` before "
                "you finish.\n\n"
                "Before any schema exploration, run the environment-preflight "
                "check: confirm the session actually has a usable, authorized "
                "warehouse (do not assume the default env/session warehouse "
                "works — it may not exist or may be unauthorized). Also confirm "
                "the dbt profiles-directory resolution fact from "
                "kg-dbt-profiles-default-path when the task builds a new dbt "
                "project. Fold both the confirmed-working warehouse/persistent "
                "fix and the profiles-directory fact into the Agent Usage Guide "
                "section of KG.md; do not add new top-level sections for them "
                "and do not exceed the 300-line budget.\n\n"
                "For any weighted composite score or percentile/ratio-based "
                "metric in Metrics & Business Rules, follow "
                "kg-composite-formula-worked-example: compute at least one "
                "worked numeric example from live rows before writing the "
                "formula down, rather than describing it in prose alone."
            ),
            "skills": [
                "kg-agent-usage-guide",
                "kg-metrics-and-business-rules",
                "kg-structure",
                "kg-environment-preflight",
                "kg-column-verification-protocol",
                "kg-dbt-profiles-default-path",
                "kg-composite-formula-worked-example",
            ],
        },
    ],
}
\end{promptbox}

\subsubsection{Final Pipeline Specification Artifacts}

Below is \texttt{KG.md} for task
\texttt{dbt-fix-marketing-attribution}, produced by $\prog^\ast$. It records verified tables, columns, data values, and environment settings for the downstream solver agent. It also flags differences between
the task instruction and the live warehouse instead of making
unsupported assumptions.

\begin{promptbox}[Final \texttt{KG.md} for \texttt{dbt-fix-marketing-attribution} produced by pipeline specification $\prog^\ast$]
# Knowledge Graph: dbt-fix-marketing-attribution

## Structure

**Confirmed sources** (database `DBT_BENCH_RETAIL`, schema `MAIN`; all confirmed via `SHOW TABLES`/`SHOW VIEWS`/`INFORMATION_SCHEMA.TABLES` + `DESCRIBE VIEW`):

- `DBT_BENCH_RETAIL.MAIN.STG_GA__SESSIONS` (VIEW, 5053 rows, 5053 distinct `SESSION_ID`, 0 NULL `SESSION_ID`/`SESSION_START`)
  Columns: `SESSION_ID`, `VISITOR_ID`, `CUSTOMER_ID`, `CHANNEL_ID`, `SESSION_START` (VARCHAR — timestamp-like string, NOT native date/timestamp), `SESSION_END` (VARCHAR), `DURATION_SECONDS` (NUMBER), `PAGE_VIEWS` (NUMBER), `LANDING_PAGE`, `EXIT_PAGE`, `REFERRER`, `UTM_SOURCE`, `UTM_MEDIUM`, `UTM_CAMPAIGN`, `DEVICE_TYPE`, `BROWSER`, `OS`, `IP_ADDRESS`, `COUNTRY`, `IS_CONVERTED` (VARCHAR), `ORDER_ID` (VARCHAR, always NULL in this env — 0/5053 populated), `CREATED_AT`, plus metadata cols `_LOADED_AT`, `_SOURCE_SYSTEM`, `_BATCH_ID`, `_ROW_NUMBER`, `_ROW_HASH`.

- `DBT_BENCH_RETAIL.MAIN.STG_GA__EVENTS` (VIEW, 15047 rows)
  Columns: `EVENT_ID`, `SESSION_ID` (1131 rows NULL), `EVENT_TYPE`, `EVENT_NAME`, `EVENT_TIMESTAMP` (VARCHAR; 0 literal NULLs but 3801/15047 rows fail `TRY_CAST(... AS TIMESTAMP)` — dirty/malformed timestamp strings), `PAGE_URL`, `ELEMENT_ID`, `ELEMENT_CLASS`, `PRODUCT_ID`, `EVENT_VALUE` (VARCHAR storing numeric-looking decimal strings, e.g. `86.52592478731349`), `EVENT_DATA`, `CREATED_AT`, plus same metadata cols as sessions.

- `DBT_BENCH_RETAIL.MAIN.INT_SESSIONS_EVENTS_JOINED` (VIEW, LEFT JOIN of the two above; 15775 rows vs 5053 distinct `SESSION_ID` → confirmed fanout ~3.1x rows per session)
  Columns (15 total, confirmed via `DESCRIBE VIEW`): `SESSION_ID`, `VISITOR_ID`, `CUSTOMER_ID`, `CHANNEL_ID`, `SESSION_START`, `SESSION_END`, `DURATION_SECONDS`, `PAGE_VIEWS`, `T2_EVENT_ID`, `T2_EVENT_TYPE`, `T2_EVENT_NAME`, `T2_EVENT_TIMESTAMP`, `T2_PAGE_URL`, `T2_ELEMENT_ID`, `T2_ELEMENT_CLASS`.
  **`T2_EVENT_VALUE` DOES NOT EXIST in this view in this environment** — confirmed absent from the 15-column `DESCRIBE VIEW` output. This matches the instruction's warned "may omit `t2_event_value` entirely" branch, and in this live warehouse that branch is the actual case: revenue cannot be sourced from `int_sessions_events_joined` here.

Rejected/decoy candidates seen in schema search (do not use — wrong grain or duplicative of the above): `STG_RAW_GA__SESSIONS*`, `STG_SESSIONS*`, `RAW_GA.SESSIONS*`, `STG_RAW_GA__EVENTS*`, `FCT_MULTI_TOUCH_ATTRIBUTION`, `FCT_RECONSTRUCTED_SESSIONS`, `RPT_SESSION_HOURLY_DISTRIBUTION` — these are separate staging-history/rebuild variants and pre-existing marts unrelated to the required `int_sessions_events_joined` / `stg_ga__sessions` / `stg_ga__events` lineage named by the instruction.

**Required output** (per instruction, not re-verified in warehouse since it does not yet exist): dbt model `marts/marketing/rpt_attribution_fixed.sql`, materializing to schema `main` (Snowflake backend) or `analytics` (DuckDB backend). Grain: one row per (`attribution_date`, `channel`, `medium`, `campaign`). Depends on `stg_ga__sessions`, `stg_ga__events`, `int_sessions_events_joined` (all confirmed above).

**Joins:**
- `stg_ga__sessions.SESSION_ID` = `int_sessions_events_joined.SESSION_ID` — this is the join already baked into the `int_sessions_events_joined` view (LEFT JOIN sessions→events); cardinality 1:many, fanout confirmed 5053 sessions → 15775 joined rows.
- `stg_ga__events.SESSION_ID` = `stg_ga__sessions.SESSION_ID` — 1131/15047 event rows have NULL `SESSION_ID` (orphan events, cannot join).

## Metrics & Business Rules
See `runtime_skills/kg-metrics-and-business-rules.md` for the generic capture protocol. Task-specific verified facts:

**Windows:** 90-day trailing window anchored to `MAX(CAST(session_start AS DATE))` computed from `stg_ga__sessions` itself (per instruction) — NOT anchored to current wall-clock date. Confirmed `SESSION_START` values in `stg_ga__sessions` range from `1765-02-04` to `2026-01-03`; 46/5053 rows are pre-2020 outlier/garbage dates, 5007/5053 fall in a normal recent range. The MAX used for the 90-day anchor will be pulled from this same distribution (max observed = `2026-01-03 19:00:08`).

**Formula:** `conversion_rate = conversions / sessions`, divide-by-zero must be handled (per instruction, `NULLIF`-style guard), rounded to 4 decimals.

**Default/mapping rules (verified against actual NULL counts):**
- `UTM_SOURCE` NULL → `'direct'`: 387/5053 sessions have NULL `UTM_SOURCE`.
- `UTM_MEDIUM` NULL → `'none'`: 402/5053 sessions have NULL `UTM_MEDIUM`.
- `UTM_CAMPAIGN` NULL → `'none'`: 3663/5053 sessions have NULL `UTM_CAMPAIGN` (majority of sessions are uncampaigned).

**Fallback/precedence rule (verified, not assumed):** Instruction says prefer `t2_event_value` from `int_sessions_events_joined`, else use `event_value` from `stg_ga__events`. In this environment, `t2_event_value` is confirmed absent (see Structure section) — the fallback branch to `stg_ga__events.EVENT_VALUE` is the one that applies here.

**Dedup/grain rule:** Instruction requires summing revenue per session first, then aggregating to the final (date, channel, medium, campaign) grain, to avoid inflation from the confirmed 1:~3.1 session→event-row fanout in `int_sessions_events_joined`.

**Verified discrepancies (facts only, not resolved here):**
- Instruction's conversion rule states `is_converted = 1`. Actual stored tokens in `STG_GA__SESSIONS.IS_CONVERTED` (VARCHAR column, confirmed via `GROUP BY`): `'0'`=1532, `'1'`=47, `'N'`=997, `'Y'`=35, `'false'`=2364, `'true'`=78 (total 5053). Truthy-token candidates (`'1'`,`'Y'`,`'true'`) sum to 160 sessions (~3.2%); falsy-token candidates (`'0'`,`'N'`,`'false'`) sum to 4893. A literal `= 1` comparison against this VARCHAR column will not match the `'Y'`/`'true'` rows.
- Instruction's revenue/conversion-event rule states `event_name = 'purchase'` (and `t2_event_name = 'purchase'`). A case-insensitive scan (`ILIKE '%purchas%'`) across all 15047 rows of `STG_GA__EVENTS.EVENT_NAME` returned **0 matching rows**. Distinct `EVENT_NAME` values actually present (51 distinct, 1216 rows literal NULL): `'View Product'`(1420), `'Search'`(1418), `'Filter'`(1410), `'Video Play'`(1406), `'Click'`(1391), `'Form Submit'`(1384), `'Add To Cart'`(1357), `'Sort'`(1344), `'Scroll'`(1333), `'Remove From Cart'`(1321), plus ~40 additional low-frequency dirty tokens of the form `'event_N.0'` (each count 1–2, ~46 rows total). The same absence of `'purchase'` holds in `INT_SESSIONS_EVENTS_JOINED.T2_EVENT_NAME` (11 distinct groups observed, same 10 named categories plus NULL, no `'purchase'`).
- `STG_GA__EVENTS.EVENT_VALUE` is stored as VARCHAR containing decimal numeric strings (e.g. `'86.52592478731349'`) — requires explicit cast to a numeric type before summing.
- `SESSION_START`, `SESSION_END`, `EVENT_TIMESTAMP` are all stored as VARCHAR, not native DATE/TIMESTAMP, in this environment — confirmed via `DESCRIBE VIEW`. `TRY_CAST(SESSION_START AS TIMESTAMP)` succeeded on sampled rows; `TRY_CAST(EVENT_TIMESTAMP AS TIMESTAMP)` failed on 3801/15047 rows (dirty timestamp strings in the events table specifically, not the sessions table).
- `STG_GA__SESSIONS.CUSTOMER_ID` is NULL on 3632/5053 rows (guest/anonymous sessions) and `ORDER_ID` is NULL on all 5053 rows in this environment.

## Environment & Execution Notes

**Fresh-session, zero-override check:** `SELECT CURRENT_WAREHOUSE(), CURRENT_ROLE(), CURRENT_DATABASE(), CURRENT_SCHEMA(), CURRENT_USER(), CURRENT_ACCOUNT()` run in a brand-new `sql_execute` session with no prior `USE ...` statements issued in that session and no override of the architect's or any prior agent's context → literal result: `WH=KG_AUTHOR_WH, ROLE=PUBLIC, DB=DBT_BENCH_RETAIL, SCHEMA=MAIN, USER=DEX_EVAL_USER, ACCT=UOB68132` (query_id `01c72a54-030b-8aae-0032-d0c3016224b6`). The default connection resolved cleanly — no NULL/none, no failure to record or fix.

**Compute verification:** `SELECT COUNT(*) FROM DBT_BENCH_RETAIL.MAIN.STG_GA__SESSIONS` on that same resolved context → `5053` rows returned successfully (query_id `01c72a54-030b-8aae-0032-d0c3016224e6`). This confirms `KG_AUTHOR_WH` is not just a name returned by `CURRENT_WAREHOUSE()` — it is actually authorized and usable by role `PUBLIC` for real compute in this environment.

**Schema alignment (checked, not assumed):** The task instruction requires the Snowflake output model to land in schema `main` via a `generate_schema_name` macro override. The live session's `CURRENT_SCHEMA()` already resolves to `MAIN` with zero overrides. These agree in this environment — there is no live discrepancy between the instruction's hardcoded schema literal and the session's default schema to flag here.

**Instruction to downstream solver:** No environment fix was needed to produce this KG — the default warehouse/role/database/schema all resolved and worked with zero manual overrides. If, in your own session, you find you must override any of these (e.g. to get `dbt run` or your own self-tests to pass), leave that override in the **persisted** profile/environment used for final scoring (e.g. `profiles.yml`, environment variables, or whatever config file supplies these values at grading time) — do NOT revert it back afterward under the belief that reverting is "cleanup." The independent grading connection opens its own fresh session and inherits only the persisted defaults, not anything you fixed transiently in your live session. Immediately before finishing, verify with a second, genuinely fresh connection/session (not the one you just patched) that the build still succeeds under the exact environment state that will be graded. Do not blanket-override the schema/database literal in `profiles.yml`/macros based solely on this session's env var reading — the alignment check above already confirms they agree in this environment, so no override is warranted here without new evidence.

## Known Ambiguities & Risk Flags

- **`is_converted` truthy definition** (real COUNT queries against `DBT_BENCH_RETAIL.MAIN.STG_GA__SESSIONS`, 5053 total rows): Instruction states a session is converted if `is_converted = 1`. Interpretation A — literal `IS_CONVERTED = '1'` → **47** sessions. Interpretation B — truthy-token set `IS_CONVERTED IN ('1','Y','true')` → **160** sessions (`'1'`=47, `'Y'`=35, `'true'`=78). Complementary falsy-token set `IS_CONVERTED IN ('0','N','false')` → **4893** sessions (`'0'`=1532, `'N'`=997, `'false'`=2364). Disagreement: 47 vs 160 — a ~3.4x difference in how many sessions count as converted. Affects: `conversions` numerator, `conversion_rate`, and which sessions' revenue is eligible to be summed. Not resolved here — solver must choose an interpretation and be prepared to justify it; both counts are real and verified, not modeled.

- **`event_name = 'purchase'` literal match** (real COUNT query against `DBT_BENCH_RETAIL.MAIN.STG_GA__EVENTS`, 15047 total rows, 1216 NULL `EVENT_NAME`): Instruction states conversion/revenue events are identified by `event_name = 'purchase'` (or `t2_event_name = 'purchase'` in the joined view). Literal match `EVENT_NAME = 'purchase'` → **0** rows. Case-insensitive fuzzy match `EVENT_NAME ILIKE '%purchas%'` → **0** rows. Distinct-value scan confirms 51 distinct non-NULL `EVENT_NAME` values actually present, none resembling purchase (top values: `'View Product'`=1420, `'Search'`=1418, `'Filter'`=1410, `'Video Play'`=1406, `'Click'`=1391, `'Form Submit'`=1384, `'Add To Cart'`=1357, `'Sort'`=1344, `'Scroll'`=1333, `'Remove From Cart'`=1321, plus ~40 low-frequency dirty `'event_N.0'` tokens). Risk: under the literal instruction rule, `attributed_revenue` will be **0 for every row** of the final output, because the named conversion-event literal does not exist anywhere in this environment's data under any casing tried. This is a structural risk, not a two-way count disagreement — flagging it rather than substituting a different event-name literal on the solver's behalf.

Runtime skill files authored per assignment (full protocols, generic/reusable, no task-specific names inside them):
- `runtime_skills/kg-agent-usage-guide.md`
- `runtime_skills/kg-metrics-and-business-rules.md`
- `runtime_skills/kg-structure.md`
- `runtime_skills/kg-ambiguity-guardrails.md`
- `runtime_skills/kg-environment-handoff.md`
\end{promptbox}

The suite below turns KG statements into executable checks. Some tests
verify recorded facts against the live data warehouse, while others
recompute business rules from source data and compare them with the
solver's output. Their failure messages point the solver to the relevant
KG statement, helping it locate and fix mismatches.

\begin{promptbox}[Adherence test suite \texttt{test\_kg.py} authored by \texttt{claude-opus-4-8}]
"""Self-correction test suite for task: dbt-fix-marketing-attribution.

[KG-ADHERENCE] tests encode specific facts stated in the Knowledge Graph (KG):
required output model, schema, grain, columns, metric formulas, filter predicates,
UTM default mappings, dedup-before-aggregate rules, and the confirmed numeric
anchors (160 truthy is_converted sessions, 1216 NULL event_name rows, 5053 sessions).

[UNIVERSAL] tests encode output-quality invariants the KG's Output requirements
imply (no NULL required columns, conversion_rate in [0,1], conversions <= sessions)
and cross-source reconciliation against the confirmed sources.

These tests are derived EXCLUSIVELY from the KG — not from instruction.md, not from
any solution SQL, and not from the hidden grader. Passing them is strong evidence the
agent followed the KG. A failing test points to a specific KG sentence to re-read
(quoted in each failure message).
"""

import os
import sys

_here = os.path.dirname(os.path.abspath(__file__))
for _p in [_here, os.path.normpath(os.path.join(_here, "..", ".."))]:
    if os.path.exists(os.path.join(_p, "test_backbone.py")):
        sys.path.insert(0, _p)
        break
from test_backbone import execute_query, execute_scalar, make_conn_fixture, require

# ── Task constants ────────────────────────────────────────────────────────────
AGENT_PROJECT_DIR = "/app/dbt_project"

# Output schema: main (Snowflake) or analytics (DuckDB). We resolve per db_type.
SCHEMA_DUCKDB = "analytics"
SCHEMA_SNOWFLAKE = "main"

MODEL = "rpt_attribution_fixed"
MODEL_FILE = f"{AGENT_PROJECT_DIR}/models/marts/marketing/rpt_attribution_fixed.sql"

# Confirmed sources (database DBT_BENCH_RETAIL, schema MAIN).
SRC_SCHEMA = "main"
STG_SESSIONS = "stg_ga__sessions"
STG_EVENTS = "stg_ga__events"
INT_JOINED = "int_sessions_events_joined"

# Expected grain columns per KG ("one row per (attribution_date, channel, medium, campaign)").
GRAIN_COLS = ["attribution_date", "channel", "medium", "campaign"]

conn = make_conn_fixture(check_paths=[MODEL_FILE])


def _out_schema(db_type):
    return SCHEMA_DUCKDB if db_type == "duckdb" else SCHEMA_SNOWFLAKE


def _fqn(schema, table):
    return schema + "." + table


def _columns(c, db_type, schema, table):
    if db_type == "duckdb":
        rows = execute_query(
            c, db_type,
            "SELECT name FROM pragma_table_info('" + schema + "." + table + "')",
        )
        return set(str(r[0]).lower() for r in rows)
    rows = execute_query(
        c, db_type,
        "SELECT column_name FROM information_schema.columns "
        "WHERE lower(table_schema) = lower('" + schema + "') "
        "AND lower(table_name) = lower('" + table + "')",
    )
    return set(str(r[0]).lower() for r in rows)


# ── Prerequisite ────────────────────────────────────────────────────────────
def test_dbt_run_prerequisite(conn):
    c, db_type = conn
    schema = _out_schema(db_type)
    cnt = execute_scalar(
        c, db_type,
        "SELECT COUNT(*) FROM " + _fqn(schema, MODEL),
    )
    require(
        cnt is not None and cnt > 0,
        "[KG-ADHERENCE] Required output dbt model 'marts/marketing/"
        "rpt_attribution_fixed.sql' must be built and populated — run dbt first.",
    )


# ── Source sanity anchors (confirmed in KG) ─────────────────────────────────
def test_source_sessions_row_count(conn):
    c, db_type = conn
    cnt = execute_scalar(
        c, db_type,
        "SELECT COUNT(*) FROM " + _fqn(SRC_SCHEMA, STG_SESSIONS),
    )
    require(
        cnt == 5053,
        "[UNIVERSAL] KG: 'STG_GA__SESSIONS (VIEW, 5053 rows)'.",
    )


def test_source_sessions_distinct_session_id(conn):
    c, db_type = conn
    cnt = execute_scalar(
        c, db_type,
        "SELECT COUNT(DISTINCT session_id) FROM " + _fqn(SRC_SCHEMA, STG_SESSIONS),
    )
    require(
        cnt == 5053,
        "[UNIVERSAL] KG: '5053 distinct SESSION_ID, 0 NULL SESSION_ID'.",
    )


def test_source_events_row_count(conn):
    c, db_type = conn
    cnt = execute_scalar(
        c, db_type,
        "SELECT COUNT(*) FROM " + _fqn(SRC_SCHEMA, STG_EVENTS),
    )
    require(
        cnt == 15047,
        "[UNIVERSAL] KG: 'STG_GA__EVENTS (VIEW, 15047 rows)'.",
    )


def test_source_int_joined_fanout(conn):
    c, db_type = conn
    total = execute_scalar(
        c, db_type,
        "SELECT COUNT(*) FROM " + _fqn(SRC_SCHEMA, INT_JOINED),
    )
    distinct = execute_scalar(
        c, db_type,
        "SELECT COUNT(DISTINCT session_id) FROM " + _fqn(SRC_SCHEMA, INT_JOINED),
    )
    require(
        total == 15775 and distinct == 5053,
        "[KG-ADHERENCE] KG: 'INT_SESSIONS_EVENTS_JOINED ... 15775 rows vs 5053 "
        "distinct SESSION_ID -> fanout ~3.1x rows per session'. This fanout is "
        "why revenue must be summed per session BEFORE final aggregation.",
    )


def test_int_joined_lacks_t2_event_value(conn):
    c, db_type = conn
    cols = _columns(c, db_type, SRC_SCHEMA, INT_JOINED)
    require(
        "t2_event_value" not in cols,
        "[KG-ADHERENCE] KG: 'T2_EVENT_VALUE DOES NOT EXIST in this view in this "
        "environment' — the fallback to stg_ga__events.EVENT_VALUE applies here.",
    )


# ── is_converted truthy anchor (160) ─────────────────────────────────────────
def test_is_converted_truthy_anchor_160(conn):
    c, db_type = conn
    cnt = execute_scalar(
        c, db_type,
        "SELECT COUNT(*) FROM " + _fqn(SRC_SCHEMA, STG_SESSIONS) +
        " WHERE is_converted IN ('1','Y','true')",
    )
    require(
        cnt == 160,
        "[KG-ADHERENCE] KG: truthy-token set IS_CONVERTED IN ('1','Y','true') "
        "-> 160 sessions ('1'=47,'Y'=35,'true'=78).",
    )


def test_is_converted_literal_one_anchor_47(conn):
    c, db_type = conn
    cnt = execute_scalar(
        c, db_type,
        "SELECT COUNT(*) FROM " + _fqn(SRC_SCHEMA, STG_SESSIONS) +
        " WHERE is_converted = '1'",
    )
    require(
        cnt == 47,
        "[KG-ADHERENCE] KG: literal IS_CONVERTED = '1' -> 47 sessions "
        "(Interpretation A of the is_converted ambiguity).",
    )


def test_is_converted_falsy_anchor_4893(conn):
    c, db_type = conn
    cnt = execute_scalar(
        c, db_type,
        "SELECT COUNT(*) FROM " + _fqn(SRC_SCHEMA, STG_SESSIONS) +
        " WHERE is_converted IN ('0','N','false')",
    )
    require(
        cnt == 4893,
        "[KG-ADHERENCE] KG: falsy-token set IS_CONVERTED IN ('0','N','false') "
        "-> 4893 sessions.",
    )


# ── event_name = 'purchase' structural risk ──────────────────────────────────
def test_event_name_null_anchor_1216(conn):
    c, db_type = conn
    cnt = execute_scalar(
        c, db_type,
        "SELECT COUNT(*) FROM " + _fqn(SRC_SCHEMA, STG_EVENTS) +
        " WHERE event_name IS NULL",
    )
    require(
        cnt == 1216,
        "[KG-ADHERENCE] KG: 'STG_GA__EVENTS ... 1216 NULL EVENT_NAME'.",
    )


def test_event_name_purchase_absent(conn):
    c, db_type = conn
    cnt = execute_scalar(
        c, db_type,
        "SELECT COUNT(*) FROM " + _fqn(SRC_SCHEMA, STG_EVENTS) +
        " WHERE event_name = 'purchase'",
    )
    require(
        cnt == 0,
        "[KG-ADHERENCE] KG: literal EVENT_NAME = 'purchase' -> 0 rows. "
        "Structural risk: attributed_revenue may be 0 for every output row.",
    )


# ── UTM default-mapping NULL anchors ─────────────────────────────────────────
def test_utm_source_null_anchor_387(conn):
    c, db_type = conn
    cnt = execute_scalar(
        c, db_type,
        "SELECT COUNT(*) FROM " + _fqn(SRC_SCHEMA, STG_SESSIONS) +
        " WHERE utm_source IS NULL",
    )
    require(
        cnt == 387,
        "[KG-ADHERENCE] KG: 'UTM_SOURCE NULL -> direct: 387/5053 sessions'.",
    )


def test_utm_medium_null_anchor_402(conn):
    c, db_type = conn
    cnt = execute_scalar(
        c, db_type,
        "SELECT COUNT(*) FROM " + _fqn(SRC_SCHEMA, STG_SESSIONS) +
        " WHERE utm_medium IS NULL",
    )
    require(
        cnt == 402,
        "[KG-ADHERENCE] KG: 'UTM_MEDIUM NULL -> none: 402/5053 sessions'.",
    )


def test_utm_campaign_null_anchor_3663(conn):
    c, db_type = conn
    cnt = execute_scalar(
        c, db_type,
        "SELECT COUNT(*) FROM " + _fqn(SRC_SCHEMA, STG_SESSIONS) +
        " WHERE utm_campaign IS NULL",
    )
    require(
        cnt == 3663,
        "[KG-ADHERENCE] KG: 'UTM_CAMPAIGN NULL -> none: 3663/5053 sessions'.",
    )


# ── Output model: existence, columns, grain, uniqueness ─────────────────────
def test_model_exists_as_table(conn):
    c, db_type = conn
    schema = _out_schema(db_type)
    if db_type == "duckdb":
        rows = execute_query(
            c, db_type,
            "SELECT table_name FROM information_schema.tables "
            "WHERE lower(table_schema)=lower('" + schema + "') "
            "AND lower(table_name)=lower('" + MODEL + "')",
        )
    else:
        rows = execute_query(
            c, db_type,
            "SELECT table_name FROM information_schema.tables "
            "WHERE lower(table_schema)=lower('" + schema + "') "
            "AND lower(table_name)=lower('" + MODEL + "')",
        )
    require(
        len(rows) >= 1,
        "[KG-ADHERENCE] KG: Required output dbt model 'marts/marketing/"
        "rpt_attribution_fixed.sql' materializing to schema '" + schema + "'.",
    )


def test_model_has_grain_columns(conn):
    c, db_type = conn
    schema = _out_schema(db_type)
    cols = _columns(c, db_type, schema, MODEL)
    missing = [g for g in GRAIN_COLS if g not in cols]
    require(
        len(missing) == 0,
        "[KG-ADHERENCE] KG grain: 'one row per (attribution_date, channel, medium, "
        "campaign)'. Missing grain columns: " + str(missing),
    )


def test_model_has_metric_columns(conn):
    c, db_type = conn
    schema = _out_schema(db_type)
    cols = _columns(c, db_type, schema, MODEL)
    needed = ["sessions", "conversions", "conversion_rate"]
    missing = [n for n in needed if n not in cols]
    require(
        len(missing) == 0,
        "[KG-ADHERENCE] KG metrics: sessions, conversions, conversion_rate "
        "(= conversions / sessions). Missing: " + str(missing),
    )


def test_model_has_revenue_column(conn):
    c, db_type = conn
    schema = _out_schema(db_type)
    cols = _columns(c, db_type, schema, MODEL)
    require(
        "attributed_revenue" in cols,
        "[KG-ADHERENCE] KG: attributed_revenue derived from t2_event_value else "
        "stg_ga__events.EVENT_VALUE, summed per session first.",
    )


def test_model_grain_is_unique(conn):
    c, db_type = conn
    schema = _out_schema(db_type)
    dup = execute_query(
        c, db_type,
        "SELECT attribution_date, channel, medium, campaign, COUNT(*) AS n "
        "FROM " + _fqn(schema, MODEL) +
        " GROUP BY attribution_date, channel, medium, campaign HAVING COUNT(*) > 1",
    )
    require(
        len(dup) == 0,
        "[KG-ADHERENCE] KG grain: 'one row per (attribution_date, channel, medium, "
        "campaign)' — the grain key must be unique (no duplicate combinations).",
    )


def test_model_row_count_positive(conn):
    c, db_type = conn
    schema = _out_schema(db_type)
    cnt = execute_scalar(
        c, db_type,
        "SELECT COUNT(*) FROM " + _fqn(schema, MODEL),
    )
    require(
        cnt is not None and cnt > 0,
        "[UNIVERSAL] Output model must produce at least one grain row.",
    )


# ── Data quality invariants (Output requirements) ───────────────────────────
def test_no_null_required_columns(conn):
    c, db_type = conn
    schema = _out_schema(db_type)
    rows = execute_query(
        c, db_type,
        "SELECT * FROM " + _fqn(schema, MODEL) +
        " WHERE attribution_date IS NULL OR channel IS NULL OR medium IS NULL "
        "OR campaign IS NULL OR sessions IS NULL OR conversions IS NULL "
        "OR conversion_rate IS NULL",
    )
    require(
        len(rows) == 0,
        "[UNIVERSAL] Output requirements forbid NULL in required columns "
        "(grain keys, sessions, conversions, conversion_rate).",
    )


def test_conversion_rate_in_range(conn):
    c, db_type = conn
    schema = _out_schema(db_type)
    rows = execute_query(
        c, db_type,
        "SELECT conversion_rate FROM " + _fqn(schema, MODEL) +
        " WHERE conversion_rate < 0.0 OR conversion_rate > 1.0",
    )
    require(
        len(rows) == 0,
        "[UNIVERSAL] KG: 'conversion_rate = conversions / sessions' must be in "
        "[0.0, 1.0].",
    )


def test_conversions_le_sessions(conn):
    c, db_type = conn
    schema = _out_schema(db_type)
    rows = execute_query(
        c, db_type,
        "SELECT sessions, conversions FROM " + _fqn(schema, MODEL) +
        " WHERE conversions > sessions",
    )
    require(
        len(rows) == 0,
        "[KG-ADHERENCE] KG: converted sessions are a subset of all sessions — "
        "conversions must never exceed sessions.",
    )


def test_sessions_nonnegative(conn):
    c, db_type = conn
    schema = _out_schema(db_type)
    rows = execute_query(
        c, db_type,
        "SELECT sessions FROM " + _fqn(schema, MODEL) +
        " WHERE sessions IS NULL OR sessions < 0",
    )
    require(
        len(rows) == 0,
        "[UNIVERSAL] sessions must be a non-negative count of distinct sessions.",
    )


def test_conversions_nonnegative(conn):
    c, db_type = conn
    schema = _out_schema(db_type)
    rows = execute_query(
        c, db_type,
        "SELECT conversions FROM " + _fqn(schema, MODEL) +
        " WHERE conversions IS NULL OR conversions < 0",
    )
    require(
        len(rows) == 0,
        "[UNIVERSAL] conversions must be a non-negative count.",
    )


# ── conversion_rate formula recomputation ────────────────────────────────────
def test_conversion_rate_formula(conn):
    c, db_type = conn
    schema = _out_schema(db_type)
    # divide-by-zero guard: rate must be 0 (or NULL-guarded) when sessions=0.
    rows = execute_query(
        c, db_type,
        "SELECT sessions, conversions, conversion_rate FROM " + _fqn(schema, MODEL) +
        " WHERE sessions > 0 AND "
        "ABS(conversion_rate - ROUND(CAST(conversions AS DOUBLE)/sessions, 4)) > 0.0001",
    )
    require(
        len(rows) == 0,
        "[KG-ADHERENCE] KG: 'conversion_rate = conversions / sessions ... rounded "
        "to 4 decimals' with divide-by-zero handled.",
    )


def test_conversion_rate_zero_sessions_guard(conn):
    c, db_type = conn
    schema = _out_schema(db_type)
    rows = execute_query(
        c, db_type,
        "SELECT * FROM " + _fqn(schema, MODEL) +
        " WHERE sessions = 0 AND conversion_rate IS NULL",
    )
    require(
        len(rows) == 0,
        "[KG-ADHERENCE] KG: 'divide-by-zero must be handled (NULLIF-style guard)' — "
        "conversion_rate should not be NULL/undefined when sessions = 0.",
    )


# ── Cross-model reconciliation against confirmed sources ─────────────────────
def test_total_sessions_reconciles_to_90day_window(conn):
    c, db_type = conn
    schema = _out_schema(db_type)
    # KG: 90-day trailing window anchored to MAX(CAST(session_start AS DATE))
    # from stg_ga__sessions itself, counting distinct session_id.
    src_total = execute_scalar(
        c, db_type,
        "WITH anchored AS ("
        " SELECT session_id, CAST(session_start AS DATE) AS sd,"
        " (SELECT MAX(CAST(session_start AS DATE)) FROM " +
        _fqn(SRC_SCHEMA, STG_SESSIONS) + ") AS max_sd"
        " FROM " + _fqn(SRC_SCHEMA, STG_SESSIONS) +
        ") SELECT COUNT(DISTINCT session_id) FROM anchored"
        " WHERE sd > max_sd - 90",
    )
    model_total = execute_scalar(
        c, db_type,
        "SELECT SUM(sessions) FROM " + _fqn(schema, MODEL),
    )
    require(
        src_total is not None and model_total is not None
        and abs(int(model_total) - int(src_total)) <= 0,
        "[UNIVERSAL] KG: '90-day trailing window anchored to "
        "MAX(CAST(session_start AS DATE))' with distinct session counting — "
        "SUM(sessions) in the mart must equal distinct sessions in that window "
        "(model=" + str(model_total) + ", source=" + str(src_total) + ").",
    )


def test_attribution_date_within_90day_window(conn):
    c, db_type = conn
    schema = _out_schema(db_type)
    max_sd = execute_scalar(
        c, db_type,
        "SELECT MAX(CAST(session_start AS DATE)) FROM " +
        _fqn(SRC_SCHEMA, STG_SESSIONS),
    )
    rows = execute_query(
        c, db_type,
        "SELECT attribution_date FROM " + _fqn(schema, MODEL) +
        " WHERE CAST(attribution_date AS DATE) <= "
        "(SELECT MAX(CAST(session_start AS DATE)) - 90 FROM " +
        _fqn(SRC_SCHEMA, STG_SESSIONS) + ")",
    )
    require(
        len(rows) == 0,
        "[KG-ADHERENCE] KG: 90-day trailing window anchored to "
        "MAX(CAST(session_start AS DATE)) (max observed 2026-01-03). "
        "All attribution_date values must fall inside that window (max_sd=" +
        str(max_sd) + ").",
    )


def test_channel_defaults_direct_present_or_valid(conn):
    c, db_type = conn
    schema = _out_schema(db_type)
    # KG: UTM_SOURCE NULL -> 'direct' default. No NULL/empty channel should remain.
    rows = execute_query(
        c, db_type,
        "SELECT channel FROM " + _fqn(schema, MODEL) +
        " WHERE channel IS NULL",
    )
    require(
        len(rows) == 0,
        "[KG-ADHERENCE] KG: 'UTM_SOURCE NULL -> direct' — channel must never be "
        "NULL in the output.",
    )


def test_medium_defaults_none_present_or_valid(conn):
    c, db_type = conn
    schema = _out_schema(db_type)
    rows = execute_query(
        c, db_type,
        "SELECT medium FROM " + _fqn(schema, MODEL) +
        " WHERE medium IS NULL",
    )
    require(
        len(rows) == 0,
        "[KG-ADHERENCE] KG: 'UTM_MEDIUM NULL -> none' — medium must never be NULL.",
    )


def test_campaign_defaults_none_present_or_valid(conn):
    c, db_type = conn
    schema = _out_schema(db_type)
    rows = execute_query(
        c, db_type,
        "SELECT campaign FROM " + _fqn(schema, MODEL) +
        " WHERE campaign IS NULL",
    )
    require(
        len(rows) == 0,
        "[KG-ADHERENCE] KG: 'UTM_CAMPAIGN NULL -> none' — campaign must never be "
        "NULL.",
    )


# ── Revenue: dedup-before-aggregate & non-negativity ─────────────────────────
def test_attributed_revenue_nonnegative(conn):
    c, db_type = conn
    schema = _out_schema(db_type)
    rows = execute_query(
        c, db_type,
        "SELECT attributed_revenue FROM " + _fqn(schema, MODEL) +
        " WHERE attributed_revenue < 0",
    )
    require(
        len(rows) == 0,
        "[UNIVERSAL] attributed_revenue must be non-negative.",
    )


def test_revenue_not_inflated_by_fanout(conn):
    c, db_type = conn
    schema = _out_schema(db_type)
    # KG: sum revenue per session first, then aggregate — to avoid 1:~3.1 fanout
    # inflation. Total attributed_revenue must not exceed the total per-session
    # summed EVENT_VALUE upper bound across the whole events table.
    total_model = execute_scalar(
        c, db_type,
        "SELECT COALESCE(SUM(attributed_revenue),0) FROM " + _fqn(schema, MODEL),
    )
    # Per-session summed event value (numeric cast), a safe upper bound.
    if db_type == "duckdb":
        cast_expr = "TRY_CAST(event_value AS DOUBLE)"
    else:
        cast_expr = "TRY_CAST(event_value AS DOUBLE)"
    upper = execute_scalar(
        c, db_type,
        "WITH per_session AS ("
        " SELECT session_id, SUM(" + cast_expr + ") AS rev"
        " FROM " + _fqn(SRC_SCHEMA, STG_EVENTS) +
        " WHERE session_id IS NOT NULL GROUP BY session_id"
        ") SELECT COALESCE(SUM(rev),0) FROM per_session",
    )
    require(
        total_model is not None and upper is not None
        and float(total_model) <= float(upper) + 0.01,
        "[KG-ADHERENCE] KG: 'sum revenue per session first, then aggregate ... to "
        "avoid inflation from the confirmed 1:~3.1 session->event-row fanout'. "
        "Total attributed_revenue (" + str(total_model) + ") must not exceed the "
        "per-session summed EVENT_VALUE bound (" + str(upper) + ").",
    )


def test_revenue_zero_when_purchase_absent(conn):
    c, db_type = conn
    schema = _out_schema(db_type)
    # KG structural risk: EVENT_NAME='purchase' has 0 matches under any casing.
    # A spec-faithful (literal 'purchase') solution yields attributed_revenue = 0
    # for every row. If the agent used a substitute event token, this test flags
    # it so the choice is deliberate & justified rather than accidental.
    purchase_events = execute_scalar(
        c, db_type,
        "SELECT COUNT(*) FROM " + _fqn(SRC_SCHEMA, STG_EVENTS) +
        " WHERE event_name = 'purchase'",
    )
    total_model = execute_scalar(
        c, db_type,
        "SELECT COALESCE(SUM(attributed_revenue),0) FROM " + _fqn(schema, MODEL),
    )
    if purchase_events == 0:
        require(
            float(total_model) == 0.0,
            "[KG-ADHERENCE] KG structural risk: literal EVENT_NAME='purchase' -> "
            "0 rows, so under the spec's purchase rule attributed_revenue is 0 for "
            "every output row. Got total=" + str(total_model) + " — if non-zero, "
            "the agent substituted a different event token (must be justified).",
        )
    else:
        require(True, "")


# ── Metric consistency: conversions must reflect converted sessions ─────────
def test_conversions_consistent_with_truthy_bound(conn):
    c, db_type = conn
    schema = _out_schema(db_type)
    # KG: converted numerator depends on is_converted interpretation (47 literal
    # or 160 truthy). Total conversions in the mart must not exceed the truthy
    # upper bound of 160 (within the 90-day window it is <= 160).
    total_conv = execute_scalar(
        c, db_type,
        "SELECT COALESCE(SUM(conversions),0) FROM " + _fqn(schema, MODEL),
    )
    require(
        total_conv is not None and int(total_conv) <= 160,
        "[KG-ADHERENCE] KG is_converted ambiguity: truthy set ('1','Y','true') = "
        "160 sessions is the maximum eligible. SUM(conversions)=" +
        str(total_conv) + " must not exceed 160.",
    )
\end{promptbox}

\section{Math \& System Benchmark Optimization}
\label{sec:app-math}

\subsection{Detailed Methodology}
\label{sec:app-math-method}

All math and ADRS runs share a 100-iteration budget and a matched backbone (\texttt{claude-sonnet-5} or \texttt{openai-gpt-5.6-terra}).
AdaEvolve, OpenEvolve, and GEPA mutate through a stateless completion; \cocoevolve{} uses the agentic loop of Appendix~\ref{sec:app-loop}, skipping the InfoGain branch because these tasks have no instance pool.
The AdaEvolve row in Table~\ref{tab:optimization_results} reruns AdaEvolve's control loop with the same backbone model as \cocoevolve{}. Matching the backbone removes model choice as an explanation for the observed gap between these search setups.
We report the best objective reached within the shared budget.

\paragraph{Task instances and metrics.}
Table~\ref{tab:app-math-tasks} records the instance size, the scalar reported in the main tables, and the Human/SOTA source.
Higher is better except Cloudcast, which reports total broadcast cost.

\begin{table}[htbp]
\centering
\adjustbox{max width=\textwidth}{%
\begin{tabular}{llll}
\toprule
Task & Instance & Reported metric & Human/SOTA \\
\midrule
Circle Packing (Square) & $N{=}26$, unit square & sum of radii & 2.63598 \\
Circle Packing (Rect) & $N{=}21$, perimeter 4 & sum of radii & 2.36583 \\
Heilbronn (Triangle) & $N{=}11$ & min triangle area / host area & 0.0365 \\
Heilbronn (Convex) & $N{=}13$ & min triangle area / hull area & 0.0309 \\
MinMax Distance & $d{=}3$, $N{=}14$ & $(\min/\max)^2$ & 0.24004 \\
Signal Processing & causal window filter & \texttt{overall\_score} composite & 0.718 \\
\midrule
Cloudcast ($\downarrow$) & multi-cloud broadcast & total cost & 626.2 \\
EPLB & MoE expert placement & balancedness/speed mean & 0.1265 \\
Prism & GPU model placement & $1/\mathrm{avg\_kvpr}$ + success & 21.89 \\
LLM-SQL & column reorder & $0.95\cdot$hit-rate ${+}0.05\cdot$runtime & 0.692 \\
TXN & scheduling & combined throughput-style objective & 2{,}725 \\
\bottomrule
\end{tabular}}
\caption{Math and ADRS evaluator settings. Math Human/SOTA is the better of the published human reference and the best previously published automated result, from AlphaEvolve~\citep{novikov2025alphaevolve} and AdaEvolve~\citep{cemri2025adaevolve}. Systems Human/SOTA values are the single AdaEvolve references.}
\label{tab:app-math-tasks}
\end{table}

\subsection{Higher-precision math and ADRS tables}
\label{sec:app-math-full-precision}

The main-text table rounds values for readability. Tables~\ref{tab:math_results_full} and~\ref{tab:systems_results_full} report values at greater precision and retain rankings computed from the underlying values.

\begin{table}[htbp]
    \centering
    \adjustbox{max width=\textwidth}{%
    \begin{tabular}{l cc cc cc cc cc cc}
    \toprule
     & \multicolumn{2}{c}{Circle Packing (Square)} & \multicolumn{2}{c}{Circle Packing (Rect)} & \multicolumn{2}{c}{Heilbronn (Triangle)} & \multicolumn{2}{c}{Heilbronn (Convex)} & \multicolumn{2}{c}{MinMax Distance} & \multicolumn{2}{c}{Signal Processing} \\
    \cmidrule(lr){2-3} \cmidrule(lr){4-5} \cmidrule(lr){6-7} \cmidrule(lr){8-9} \cmidrule(lr){10-11} \cmidrule(lr){12-13}
    Strategy & Avg & Best & Avg & Best & Avg & Best & Avg & Best & Avg & Best & Avg & Best \\
    \midrule
    Human/SOTA    & -- & 2.63598 & -- & 2.36583 & -- & 0.0365 & -- & 0.0309 & -- & 0.24004 & -- & 0.718 \\
    \midrule
    \multicolumn{13}{l}{\textit{Backbone: Sonnet-5}} \\
    OpenEvolve       & \textbf{$2.63598 \pm 0.00000$} & \textbf{2.63598} & \textbf{$2.36557 \pm 0.00045$} & \textbf{2.36583} & $0.03197 \pm 0.00317$ & \underline{0.0346} & $0.02775 \pm 0.00216$ & 0.02900 & \underline{$0.23780 \pm 0.00309$} & \underline{0.23991} & $0.6499 \pm 0.0396$ & 0.6727 \\
    GEPA             & $2.60218 \pm 0.05856$ & \textbf{2.63598} & \underline{$2.36530 \pm 0.00091$} & \textbf{2.36583} & \textbf{$0.03588 \pm 0.00080$} & \textbf{0.0365} & \underline{$0.03018 \pm 0.00132$} & \textbf{0.03094} & \textbf{$0.24005 \pm 0.00000$} & \textbf{0.24005} & $0.6031 \pm 0.0481$ & 0.6309 \\
    AdaEvolve     & $2.63329 \pm 0.00258$ & \textbf{2.63598} & $2.36423 \pm 0.00250$ & \textbf{2.36583} & \underline{$0.03538 \pm 0.00196$} & \textbf{0.0365} & $0.02913 \pm 0.00148$ & \underline{0.03076} & \textbf{$0.24005 \pm 0.00000$} & \textbf{0.24005} & \underline{$0.7470 \pm 0.0247$} & \underline{0.7656} \\
    \cocoevolve{} & \underline{$2.63542 \pm 0.00098$} & \textbf{2.63598} & $2.36285 \pm 0.00516$ & \textbf{2.36583} & $0.03529 \pm 0.00185$ & \textbf{0.0365} & \textbf{$0.03066 \pm 0.00036$} & \textbf{0.03094} & \textbf{$0.24005 \pm 0.00000$} & \textbf{0.24005} & \textbf{$0.7656 \pm 0.0243$} & \textbf{0.7841}\\
    \midrule
    \multicolumn{13}{l}{\textit{Backbone: GPT-5.6-terra}} \\
    OpenEvolve       & \textbf{$2.63598 \pm 0.00000$} & \textbf{2.63598} & \underline{$2.36478 \pm 0.00091$} & \underline{2.36583} & $0.03038 \pm 0.00018$ & 0.0305 & $0.02483 \pm 0.00226$ & 0.02709 & $0.23974 \pm 0.00016$ & 0.23983 & $0.5653 \pm 0.0380$ & 0.6075 \\
    GEPA             & \underline{$2.63435 \pm 0.00283$} & \textbf{2.63598} & $2.36407 \pm 0.00186$ & \underline{2.36583} & \textbf{$0.03507 \pm 0.00024$} & \underline{0.03521} & \textbf{$0.02607 \pm 0.00301$} & \textbf{0.02781} & \textbf{$0.24005 \pm 0.00000$} & \textbf{0.24005} & $0.6327 \pm 0.0491$ & 0.6610 \\
    AdaEvolve        & $2.63315 \pm 0.00200$ & \underline{2.63431} & \textbf{$2.36584 \pm 0.00001$} & \textbf{2.36584} & \underline{$0.03457 \pm 0.00059$} & 0.0352 & $0.02353 \pm 0.00104$ & 0.02413 & \underline{$0.23982 \pm 0.00019$} & \underline{0.24002} & \textbf{$0.7921 \pm 0.0197$} & \textbf{0.8133} \\
    \cocoevolve{}    & $2.63375 \pm 0.00255$ & \textbf{2.63598} & $2.36378 \pm 0.00354$ & \underline{2.36583} & $0.03408 \pm 0.00214$ & \textbf{0.0365} & \underline{$0.02544 \pm 0.00205$} & \underline{0.02712} & \textbf{$0.24005 \pm 0.00000$} & \textbf{0.24005} & \underline{$0.7563 \pm 0.0608$} & \underline{0.80478} \\
    \bottomrule
    \end{tabular}}
    \caption{Mathematical optimization benchmarks.
    We report \textbf{Mean} $\pm$ \textbf{Std} and \textbf{Best} objective values.
    Higher is better for all metrics.
    ($N$ values: Square=$26$, Rect=$21$, Triangle=$11$, Convex=$13$; MinMax $d{=}3$, $N{=}14$).
    See Table~\ref{tab:app-math-tasks}.
    Human/SOTA is the better of the published Human reference and the best previously published automated result on each problem, drawn from AlphaEvolve~\cite{novikov2025alphaevolve} and AdaEvolve~\cite{cemri2025adaevolve}.
    Values at greater precision corresponding to the main-paper Table~\ref{tab:optimization_results}. Bold/underline follow the same within-backbone ranking on the underlying values.
    }
    \label{tab:math_results_full}
\end{table}

\begin{table}[htbp]
    \centering
    \adjustbox{max width=\textwidth}{%
    \begin{tabular}{l cc cc cc cc cc}
    \toprule
     & \multicolumn{2}{c}{Cloudcast ($\downarrow$)} & \multicolumn{2}{c}{EPLB} & \multicolumn{2}{c}{Prism} & \multicolumn{2}{c}{LLM-SQL} & \multicolumn{2}{c}{TXN} \\
    \cmidrule(lr){2-3} \cmidrule(lr){4-5} \cmidrule(lr){6-7} \cmidrule(lr){8-9} \cmidrule(lr){10-11}
    Strategy & Avg & Best & Avg & Best & Avg & Best & Avg & Best & Avg & Best \\
    \midrule
    Human/SOTA & -- & 626.2 & -- & 0.1265 & -- & 21.89 & -- & 0.692 & -- & 2{,}725 \\
    \midrule
    \multicolumn{11}{l}{\textit{Backbone: Sonnet-5}} \\
    OpenEvolve       & $660.693089 \pm 0.007070$ & 660.689007 & $0.1273 \pm 0.0001$ & 0.1273 & $26.236 \pm 0.032$ & \underline{26.256} & $0.7132 \pm 0.0115$ & 0.7263 & $3{,}945 \pm 123$ & 4{,}016 \\
    GEPA             & $621.090169 \pm 5.213452$ & \underline{618.080181} & $0.1332 \pm 0.0103$ & \underline{0.1451} & \underline{$26.256 \pm 0.000$} & \underline{26.256} & $0.5235 \pm 0.3733$ & \textbf{0.7522} & $3{,}874 \pm 218$ & 4{,}000 \\
    AdaEvolve     & \underline{$618.096950 \pm 0.008222$} & 618.087600 & \underline{$0.1446 \pm 0.0006$} & {0.1450} & \textbf{$26.391 \pm 0.233$} & \textbf{26.660} & \underline{$0.7235 \pm 0.0161$} & \underline{0.7328} & \underline{$4{,}115 \pm 17$} & \underline{4{,}132} \\
    \cocoevolve{} & \textbf{$618.063457 \pm 0.006616$} & \textbf{618.059637} & \textbf{$0.1454 \pm 0.0002$} & \textbf{0.1455} & \underline{$26.256 \pm 0.000$} & \underline{26.256} & \textbf{$0.7280 \pm 0.0058$} & 0.7316 & \textbf{$4{,}292 \pm 55$} & \textbf{4{,}348} \\
    \midrule
    \multicolumn{11}{l}{\textit{Backbone: GPT-5.6-terra}} \\
    OpenEvolve       & $618.096205 \pm 0.005931$ & 618.092781 & $0.1272 \pm 0.0000$ & 0.1272 & $26.076 \pm 0.136$ & 26.152 & \underline{$0.7242 \pm 0.0103$} & \underline{0.7301} & $4{,}107 \pm 134$ & 4{,}184 \\
    GEPA             & \textbf{$618.080973 \pm 0.019122$} & \textbf{618.069933} & $0.1272 \pm 0.0000$ & 0.1272 & \underline{$26.259 \pm 0.005$} & \underline{26.264} & $0.7193 \pm 0.0076$ & 0.7237 & \textbf{$4{,}268 \pm 74$} & \textbf{4{,}348} \\
    AdaEvolve     & $618.085178 \pm 0.013202$ & \textbf{618.069933} & \underline{$0.1415 \pm 0.0060$} & \textbf{0.1454} & $25.674 \pm 0.080$ & 25.734 & $0.7168 \pm 0.0065$ & 0.7236 & $4{,}144 \pm 20$ & 4{,}167 \\
    \cocoevolve{}    & \underline{$618.081749 \pm 0.010282$} & \underline{618.071097} & \textbf{$0.14472 \pm 0.00035$} & \underline{0.1451} & \textbf{$26.338 \pm 0.141$} & \textbf{26.501} & \textbf{$0.7283 \pm 0.0020$} & \textbf{0.7302} & \underline{$4{,}250 \pm 81$} & \underline{4{,}329} \\
    \bottomrule
    \end{tabular}}
    \caption{ADRS systems benchmarks.
    We report \textbf{Mean} $\pm$ \textbf{Std} and \textbf{Best} objective values.
    Higher is better, except for \textbf{Cloudcast} where lower ($\downarrow$) is better.
    Instance sizes and metrics are in Table~\ref{tab:app-math-tasks}.
    Human/SOTA values are from~\cite{cemri2025adaevolve}, which reports only a single reference value.
    Values at greater precision corresponding to the main-paper Table~\ref{tab:optimization_results}. Bold/underline follow the same within-backbone ranking on the underlying values.}
    \label{tab:systems_results_full}
\end{table}

Signal Processing's \texttt{overall\_score} is a weighted blend of a composite $J(\theta)$ (slope changes, lag, error, false reversals), smoothness, correlation, noise reduction, and success rate.

\paragraph{Signal Processing strategy convergence.}
Surprisingly, AdaEvolve and \cocoevolve{} independently converge to essentially the same champion strategy: both fit the complete signal with an L1 data term and an L1 second-difference penalty, expressed as the same sparse linear program, and return the aligned tail of the resulting piecewise-linear trend. Their material difference is only how the regularization weight is chosen. \cocoevolve{} scales a fixed coefficient by a robust noise estimate, whereas AdaEvolve uses a clipped noise-to-motion rule. In the reported exchange experiment, swapping the regularization rules also swaps the measured performance. The two champions therefore share the same broad filtering approach, and the regularization rule appears to explain their score difference in this comparison.

\subsection{Per-task analysis}
\label{sec:app-math-programs}
\label{sec:app-math-leaders}

Across both backbones, \cocoevolve{} is best or tied best in 13 of 24 math cells and 12 of 20 ADRS cells when Avg and Best are counted separately. It leads 25 of 44 cells overall, compared with 18 for GEPA, 13 for AdaEvolve, and 6 for OpenEvolve.

On Sonnet-5, \cocoevolve{} has the highest Heilbronn Convex mean and ties GEPA on Best. In ADRS, it leads both Avg and Best on Cloudcast, EPLB, and TXN, and it leads LLM-SQL on Avg. AdaEvolve leads Prism, while GEPA leads LLM-SQL on Best.

On GPT-5.6-terra, the leaders are distributed across all four methods. \cocoevolve{} leads Prism and LLM-SQL and has the highest Heilbronn Triangle Best. AdaEvolve leads Signal Processing and Circle Packing (Rect). GEPA leads Heilbronn Convex and TXN and has the best Cloudcast Avg. OpenEvolve has the best Circle Packing (Square) mean.
Circle Packing at both instances and MinMax Distance produce several ties at or near the reference values, so their means rather than their Bests are the informative column.

The traces below examine two patterns in the table. Heilbronn Convex tests whether stalled trials recover near a known reference score; TXN shows a structural rewrite with room for a larger gain. Circle Packing provides a saturated case where final best scores are tied, and Cloudcast shows several methods finishing close together. These traces explain selected runs, not every aggregate result.
InfoGain is inactive here: these tasks have no instance pool.

\paragraph{Figures.}
Heilbronn Convex layouts and best-so-far: Figures~\ref{fig:heilbronn-layouts}--\ref{fig:heilbronn-search} (Appendix~\ref{sec:app-math-heilbronn}).
Circle Packing (Square) layouts and best-so-far: Figures~\ref{fig:circle-packing-layouts}--\ref{fig:circle-packing-search} (Appendix~\ref{sec:app-math-circle}).

\subsubsection{Heilbronn Convex \texorpdfstring{($N{=}13$)}{(N=13)}}
\label{sec:app-math-heilbronn}

\begin{figure}[ht]
    \centering
    \includegraphics[width=\textwidth]{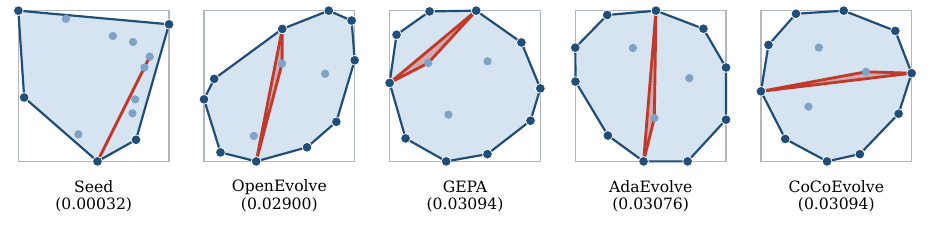}
    \caption{Heilbronn Convex ($N{=}13$) on Sonnet-5. Point sets in the unit square: hull vertices (navy), interior points (lighter), hull outlined, smallest triangle in red. Seed is unstructured. OpenEvolve reaches a weaker $13$-gon ($0.02900$); GEPA, AdaEvolve, and \cocoevolve{} all reach structured $13$-gons at their Best-cell scores.}
    \label{fig:heilbronn-layouts}
\end{figure}

\begin{figure}[ht]
    \centering
    \includegraphics[width=\textwidth]{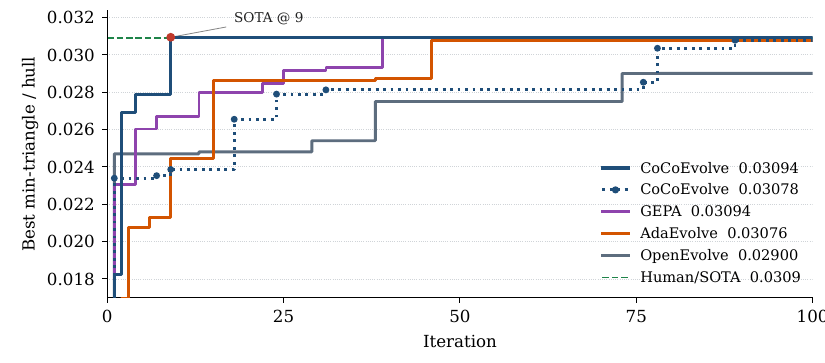}
    \caption{Heilbronn Convex best-so-far on Sonnet-5, cropped just below the first rewrite. Solid traces are each method's Best-cell run; Human/SOTA is the dashed green line. The dotted navy trace is a second retained \cocoevolve{} trial: it stalls near $0.0281$ and recovers to $0.03078$. Solid \cocoevolve{} reaches Human/SOTA at iteration $9$. The claim is Avg, not the tied Best.}
    \label{fig:heilbronn-search}
\end{figure}

\paragraph{Setup.}
The objective is the ratio of the smallest triangle area among $N{=}13$ points to the area of their convex hull; Human/SOTA is $0.0309$.
The seed returns $13$ uniform random points ($0.00032$).
All four methods run the matched $100$-iteration Sonnet-5 protocol of Appendix~\ref{sec:app-math-method}; \cocoevolve{}'s Avg in Table~\ref{tab:math_results} is three retained trials.
GEPA ties \cocoevolve{} on Best, so the analysis below is not a unique-champion story: it is why Avg stays high when two of those three trials start badly.
Figure~\ref{fig:heilbronn-layouts} shows each method's Best-cell geometry.
Figure~\ref{fig:heilbronn-search} overlays one Best-cell trace per method and a second retained \cocoevolve{} trial (dotted) that recovers from a stall; both \cocoevolve{} traces enter the published Avg.

\paragraph{Search.}
The three retained \cocoevolve{} finals are $0.03094$, $0.03078$, and $0.03025$ (Table~\ref{tab:math_results}: $0.03066 \pm 0.00036$).
Two of the three spend most of the budget near $0.028$ and still finish at the ceiling, which is what keeps Avg above GEPA and the standard deviation a quarter of GEPA's.

The solid trace is the trial that does not need a recovery.
Iterations $1$--$2$ replace the random seed with from-scratch solvers ($0.0182$, then $0.0269$).
The reflector then keeps a constrained-NLP family (\texttt{scipy.optimize.minimize}, SLSQP) and a hillclimb of that family lands Human/SOTA at iteration $9$, with a scratch self-check before each commit.
GEPA reaches the same $0.03094$ at iteration $39$.
That is sample efficiency on one trial, not the claim for the column.

The dotted trial is the CMA-ES recovery and contributes to the three-run Avg.
Its first rewrite scores $0.0234$ and stalls near $0.0281$ from iteration $31$ while the mutator retries SciPy global solvers.
At iteration $75$, the reflector marks that family as exhausted and proposes a NumPy CMA-ES.
A scratch prototype reaches $0.03035$ at iteration $78$.
At iteration $89$, the mutator returns to a dual-annealing program from iteration $19$, which scored $0.02648$, rather than using the current best CMA-ES child.
It queries the ranked archive and adds CMA-ES with an exact-bilinear SLP polish, producing the $0.03078$ champion.

The third retained trial (not plotted) is the same mechanism with a different family.
It crawls from $0.0244$ to $0.0275$ over $73$ iterations of scipy global solvers, then implements Tutte barycentric embedding (\texttt{scipy.sparse.linalg.spsolve}) at iteration $86$ and jumps to $0.03025$.

Either recovery is load-bearing for the Avg win.
If the CMA-ES trial had remained at $0.0281$ and the Tutte trial at its pre-jump $0.0275$, the three-run mean would be $0.0289$, below GEPA.
If only one of the two recovered, the mean would still sit near $0.0297$.
Heilbronn therefore shows how historical synthesis improves reliability near a ceiling.
The ledger triggers a new family after a stall, and the archive query transfers that family to a viable island.
This mechanism explains the high Avg rather than a uniquely high Best.

\subsubsection{Circle Packing (Square)}
\label{sec:app-math-circle}

\begin{figure}[ht]
    \centering
    \includegraphics[width=\textwidth]{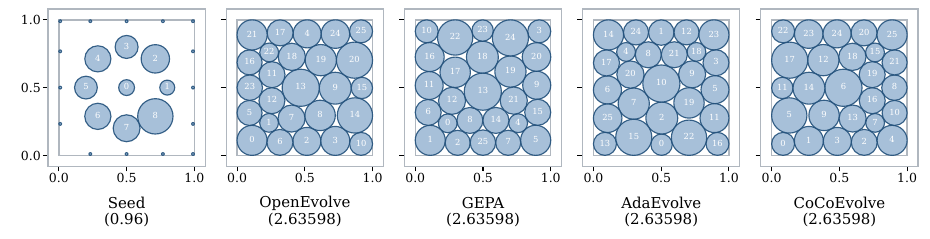}
    \caption{Circle Packing (Square, $N{=}26$) on Sonnet-5. Circles in the unit square, labeled $0$--$25$. The seed is a two-ring heuristic ($0.96$); OpenEvolve, GEPA, AdaEvolve, and \cocoevolve{} are size-biased ceiling packings at Human/SOTA ($2.63598$).}
    \label{fig:circle-packing-layouts}
\end{figure}

\begin{figure}[ht]
    \centering
    \includegraphics[width=\textwidth]{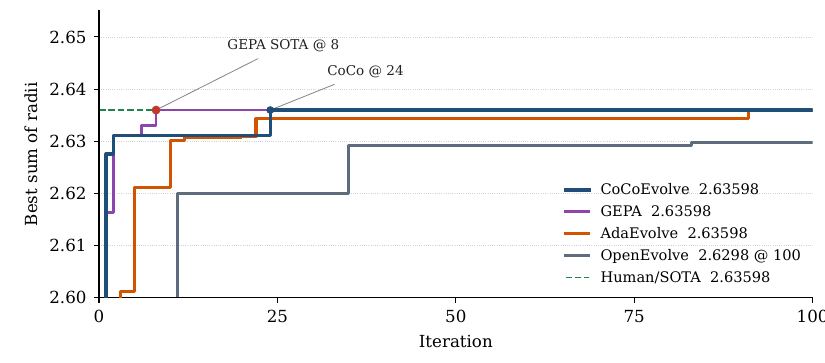}
    \caption{Circle Packing (Square, $N{=}26$) best-so-far on Sonnet-5, cropped near the ceiling. Solid traces are each method's Best-cell run; Human/SOTA is the dashed green line. GEPA first reaches $2.63598$ at iteration $8$, \cocoevolve{} at iteration $24$, AdaEvolve at iteration $91$. OpenEvolve is at $2.6298$ at iteration $100$ and only hits the ceiling at iteration $111$, outside the shared window.}
    \label{fig:circle-packing-search}
\end{figure}

\paragraph{Setup.}
The objective is the sum of radii of $N{=}26$ non-overlapping circles in the unit square; Human/SOTA is $2.63598$.
The seed is a two-ring heuristic ($0.96$).
This instance saturates: every method in Table~\ref{tab:math_results} reaches Human/SOTA in at least one Sonnet-5 trial, so Best cannot rank them.
The traces below are therefore a budget-use comparison, not a final-cell comparison.
Figure~\ref{fig:circle-packing-layouts} shows one ceiling packing per method.
Figure~\ref{fig:circle-packing-search} overlays the Best-cell best-so-far of each method, cropped near the ceiling, under the same $100$-iteration window.

\paragraph{Search.}
This is a saturation control, not a \cocoevolve{} win.
Once every method can hit the ceiling, the remaining signal is how the shared budget is spent getting there.
On the traces in Figure~\ref{fig:circle-packing-search}, GEPA is first (iteration $8$), \cocoevolve{} is second (iteration $24$), AdaEvolve reaches the ceiling at iteration $91$, and OpenEvolve is still at $2.6298$ at iteration $100$ (its Best of $2.63598$ is from iteration $111$, outside the shared window).
We do not read first-to-ceiling as a method ranking: GEPA winning that race on a saturated packing instance is compatible with Heilbronn, where the same protocol separates the methods on Avg rather than Best.

\subsubsection{Cloudcast}
\label{sec:app-math-cloudcast}

\paragraph{Setup.}
The objective is the simulator's true union-of-edges cost of a multi-cloud broadcast tree (lower is better); Human/SOTA is $626.2$.
The seed enumerates Azure single- and two-hub plans ranked by a node-count/bottleneck proxy, with an arborescence fallback, and scores $974$.
The analysis below is the \cocoevolve{} Best-cell Sonnet-5 run under the same $100$-iteration protocol.
OpenEvolve remains far from the others; GEPA, AdaEvolve, and \cocoevolve{} all finish in a tight $618$ cluster (Table~\ref{tab:systems_results}), so this subsection is about the construction that reached that cluster, not about ranking inside it.

\paragraph{Search.}
The champion no longer searches only for hubs. It evaluates six structurally distinct tree families: terminal MSA, relay hub, capacitated min-cost flow, Dreyfus--Wagner DP, resistance ranking, and TSP relay. A proxy replays the simulator's cost model and selects the cheapest candidate.

The first two iterations reduce cost from $974$ to $636$ by reading the cost graph and instance configuration and by testing scratch candidates against the live evaluator. Later cross-lineage grafts add Dreyfus--Wagner ranking and a TSP-relay construction, which produce the remaining improvement. The trace documents a substantial rewrite of the seed's search procedure. Because GEPA, AdaEvolve, and \cocoevolve{} finish close together, Cloudcast gives little evidence for ranking their final scores. TXN provides a clearer example of a large score change after a structural rewrite.

\subsubsection{TXN scheduling}
\label{sec:app-math-txn}

\paragraph{Setup.}
The objective is a combined throughput-style score of a lock-aware transaction schedule (higher is better); Human/SOTA is $2{,}725$.
The seed is tabu search over swap and or-opt-1 from Copeland / MDS / assignment-problem / greedy starts, and scores $2{,}725$.
The reported runs show substantial headroom above that seed and reference value. Both the mean and best scores can therefore distinguish methods here.
The analysis below is the \cocoevolve{} Best-cell Sonnet-5 run that produced $4{,}348$, under the same $100$-iteration protocol.

\paragraph{Search.}
The champion replaces the seed's tabu loop with a three-stage pipeline. An assignment-problem relaxation and a single-path greedy construction from the top Copeland candidate each build a starting schedule, and the cheaper one is kept; simulated annealing then applies double-bridge reheating, and first-improvement local search performs the final cleanup.

Copeland ranking and an EDA over pairwise order probabilities first move the score above the seed. The EDA also queries the ranked archive instead of editing only the parent. The final program uses assignment/greedy construction followed by simulated annealing, not a retuned tabu tenure.

TXN is therefore the far-from-ceiling counterpart to Heilbronn. In Heilbronn, the ledger improves reliability by rescuing stalled trials. In TXN, the same keep-or-switch loop replaces the seed metaheuristic and improves both Avg and Best beyond the other methods.

\section{Prompt Optimization}
\label{sec:app-prompt-opt}

\subsection{Detailed Methodology}
\label{sec:app-prompt-opt-method}

We evaluate three tasks from GEPA's benchmark suite.
All three compare an unmodified seed, GEPA, and \cocoevolve{} from the same initial program; each \cocoevolve{} run uses 100 iterations.

\paragraph{HotpotQA.}
The artifact is a four-stage retrieve-summarize-answer pipeline over a BM25 index of Wikipedia abstracts, with prompts for first-hop summarization (\texttt{summarize1}), second-hop query generation (\texttt{create\_query\_hop2}), second-hop summarization (\texttt{summarize2}), and final answering (\texttt{final\_answer}).
Each mutation changes one prompt while retrieval remains fixed, and GEPA's round-robin selector mutates the same four components in the same cycle.
We use GEPA's official 150/300/300 train, validation, and held-out-test split. The GEPA search budgets were chosen to be broadly comparable to the 100-iteration \cocoevolve{} runs, although the two methods account for evaluations differently.
The frozen Sonnet-5 backbone serves as both mutator and student.

GEPA uses the same \texttt{Benchmark} base class for AIME, HotpotQA, and IFBench. The artifact runner passes individual query instances in separate training and validation sets for all three tasks. GEPA samples three training examples by default for each reflective step. A proposed edit must strictly improve its parent's score on that minibatch. Each accepted candidate then triggers a full validation-set evaluation, whose per-example scores inform the Pareto selector's choice of future parents.

\paragraph{AIME.}
The evolved artifact is a single chain-of-thought system prompt that a frozen student model reads with each competition math problem to produce an integer answer. GEPA's fixed-seed split draws 45 training and 45 validation problems from \texttt{AI-MO/aimo-validation-aime}; its disjoint test set contains 30 problems from \texttt{MathArena/aime\_2025}. AIME uses \texttt{openai-gpt-5.6-luna} as both mutator and student. An initial Sonnet-5 run gave the seed a perfect training score, leaving no signal for optimization.

\paragraph{IFBench.}
The artifact has two prompt-driven stages, \texttt{generate\_response} and \texttt{ensure\_correct\_response}. The evaluator scores the fraction of each query's instruction constraints that the response satisfies. It supports 83 checker types: 25 from IFEval and 58 IFBench-native types, including word and letter counts, formatting, and keyword use. Eight leniency variants account for markdown emphasis and optional leading or trailing lines. Neither stage sees the constraint list. GEPA's fixed split uses the first 300 rows of its training file for validation and samples 150 training rows from the next 300 with a fixed seed. We use its bundled 294-row test file unchanged; it has six fewer rows than the public IFBench release. IFBench also uses \texttt{openai-gpt-5.6-luna} as both mutator and student.

The IFBench run uses the default \texttt{FullEvaluationPolicy} and 100 concurrent workers.

\paragraph{Repeated evaluation.}
With \texttt{openai-gpt-5.6-luna}, we evaluated the same evolved AIME prompt twice on the 45 training problems: once during optimization and once afterward. The second evaluation got five fewer answers correct, even though the prompt had not changed. The student's nonzero decoding temperature can make its answers vary between evaluations. Each optimizer ran once and produced one final candidate. We evaluated that unchanged candidate three times on each AIME and IFBench split and report the mean and standard deviation in Table~\ref{tab:prompt_opt}. These repeats measure evaluation variability, not search variability. On AIME test, five baseline questions, eight GEPA questions, and six \cocoevolve{} questions changed correctness across repeats, even though the baseline's aggregate test score was $25/30$ every time.

\subsection{An efficiency variant: sim-graph targeting and cascaded evaluation on HotpotQA}
\label{sec:app-prompt-opt-simgraph}

The HotpotQA run of Appendix~\ref{sec:app-prompt-opt-method} scores its \emph{entire} $150$-question training pool and $300$-question validation pool on every one of its $100$ iterations, unconditionally -- $450$ real model-backed questions per iteration regardless of how informative any single one is. DABStep's own evaluator (Appendix~\ref{sec:app-tools}, Appendix~\ref{sec:app-infogain}) instead scores only a bounded, informativeness-targeted subset per iteration and extrapolates the rest from the parent's own history. We test a HotpotQA variant that combines similarity-based target selection, P0/P1/P2 cascaded evaluation, and mutator self-testing. The comparison measures this combined configuration. It cannot attribute any score or time difference to one of the three changes alone.

\paragraph{Setup.} We keep the seed, 100-iteration budget, and backbone from Appendix~\ref{sec:app-prompt-opt-method}. We add three mechanisms.

\emph{Target selection.} A similarity graph and InfoGain selector operate over the 150-question training pool (Appendix~\ref{sec:app-infogain}). Each mutation receives one selected focus question instead of the full undifferentiated pool.

\emph{Cascaded evaluation.} The P0/P1/P2 cascade mirrors the DABStep evaluator (Appendix~\ref{sec:app-agent-method}). P0 evaluates only the selected target and requires a score of $0.6$. If P0 passes, P1 evaluates a stratified sample of 20 regression questions, or about $13\%$ of the pool. One third of this sample is selected by similarity-weighted risk to the target, and two thirds are sampled uniformly from the parent's tested history. P1 requires at least $0.9\times$ the parent's score on the same sample. If P1 passes, P2 explores similarity-ranked frontier tiers of 10 questions each. One random scout represents each tier, and the next tier opens only if the current tier scores at least $6/10$. The training cascade evaluates at most $\lfloor 150/3\rfloor{=}50$ questions per iteration, excluding the one-time seed bootstrap, mutator self-tests, and separately sampled validation passes. When the parent is already at ceiling on the target, a live pretest can bypass P0 and P1 and proceed directly to P2.

The child's score combines its updated per-question record with the parent's inherited record and divides the total by 150. We do not use DABStep's additive formula because that formula assumes sparse initial coverage. HotpotQA has dense coverage from the seed's full-pool bootstrap, so the additive formula would double-count observations.

\emph{Mutator self-test.} The mutator receives \texttt{test\_draft}, with a budget of three calls per turn. It can score an in-progress edit directly on the selected focus question. The mutator prompt also instructs the model to verify its draft before finishing. Both features reproduce DABStep's self-verification affordance (Appendix~\ref{sec:app-tools}).

Validation is sampled independently with probability $0.1$ per iteration. When validation is skipped, the child inherits the parent's validation score and per-question record unchanged.

\paragraph{Results.}

\begin{table}[htbp]
    \centering
    \begin{tabular}{l ccc}
    \toprule
    Evaluator & Train & Val & Test \\
    \midrule
    Full pool, every iteration (Appendix~\ref{sec:app-prompt-opt-method}) & 0.800 & 0.783 & 0.793 \\
    Sim-graph target + P0/P1/P2 cascade & 0.800 & 0.790 & 0.767 \\
    \bottomrule
    \end{tabular}
    \caption{HotpotQA, Sonnet-5, \cocoevolve{}: the published full-pool-every-iteration search versus the sim-graph-targeted/cascaded-evaluation variant of this appendix. Both rows are a single $100$-iteration search, freshly re-scored against the identical train/val/test pools.}
    \label{tab:hotpotqa_simgraph_cascade}
\end{table}

Across 100 iterations, the cascaded evaluator has a median wall-clock time of $121$s per iteration. A full 450-question pass at the same concurrency takes roughly $900$--$1100$s. The seed's one-time full-pool bootstrap takes $1115$s, which is consistent with that estimate.

In the first 50 logged iterations, seven ($14\%$) took $400$--$1100$s; one was the seed bootstrap, and the others included a sampled pass over all 300 validation questions. Extrapolating that mix to 100 iterations gives an estimated $5.5$--$6$ hours of evaluator time, versus an estimated $25$--$30$ hours for full-pool evaluation at every iteration. The implied reduction is about $4$--$5\times$. These are estimates from the observed timing mix, not measured total run times.

\paragraph{Analysis.} The combined variant matches the full-pool run's observed train accuracy ($0.800$) and has $0.007$ higher validation accuracy, but its test accuracy is $0.026$ lower, about eight questions out of 300. The timing estimate suggests lower evaluator cost. With one search and one final re-score per variant, the result does not establish equal held-out quality or isolate which added mechanism produced the difference. It also does not replace or improve the main HotpotQA result in Table~\ref{tab:prompt_opt}.

\clearpage
\section{Complete Base and Final Programs}
\label{sec:app-program-listings}

The following unabridged listings pair the seed and final program from one run each of Heilbronn Convex, Cloudcast, and TXN.
The Cloudcast and TXN listings contain only the evolved function; module-level imports (e.g., \texttt{networkx as nx}, \texttt{random}) and harness classes such as \texttt{BroadCastTopology} are supplied by the surrounding benchmark file.

\subsection{Heilbronn Convex}
\label{sec:app-program-listings-heilbronn}

\championlisting{Heilbronn Convex: base program}{listings/heilbronn-seed.py}
\championlisting{Heilbronn Convex: final evolved program}{listings/heilbronn-best.py}

\subsection{Cloudcast}
\label{sec:app-program-listings-cloudcast}

\championlisting{Cloudcast: base program}{listings/cloudcast-seed.py}
\championlisting{Cloudcast: final evolved program}{listings/cloudcast-best.py}

\subsection{TXN}
\label{sec:app-program-listings-txn}

\championlisting{TXN: base program}{listings/txn-seed.py}
\championlisting{TXN: final evolved program}{listings/txn-best.py}

\end{document}